\documentclass[sort&compress,preprint]{elsarticle}
\usepackage[utf8]{inputenc}

\usepackage{soul}
\soulregister\cite7
\soulregister\citep7
\soulregister\citet7
\soulregister\ref7
\soulregister\eqnref7
\soulregister\eqref7

\usepackage{graphicx}
\usepackage{amsmath,amssymb}
\usepackage[version=4]{mhchem}
\usepackage{lscape}  
\usepackage{siunitx}
\usepackage{rotating}
\usepackage{longtable,tabularx}
\usepackage[]{caption}
\usepackage{algorithm}
\usepackage{booktabs}
\usepackage{algpseudocode}
\usepackage{adjustbox} 
\usepackage[usenames, dvipsnames]{color}
\usepackage{tikz} 
\usepackage{colortbl}
\usepackage[margin=3.0cm]{geometry}
\usepackage{subfigure}
\usepackage[bookmarks=true,bookmarksnumbered=true,colorlinks=true,linkcolor=lblue,citecolor=lblue,urlcolor=lblue]{hyperref}
\hypersetup{
    colorlinks=false,
    pdfborder={0 0 0},
}
\newcommand{\degree}{\ensuremath{^\circ\,}}
\usepackage{natbib}
\usepackage{dsfont}
\usepackage{upgreek}
\usepackage{amsmath}
\usepackage{multirow}
\DeclareMathOperator{\E}{\mathbb{E}}

\DeclareMathOperator*{\argmin}{arg\,min}

\algrenewcommand\textproc{}
\let\oldReturn\Return 
\renewcommand{\Return}{\State\oldReturn}

\definecolor{myblue}{HTML}{4C72B0}
\definecolor{myred}{HTML}{C54E52}
\definecolor{mygreen}{HTML}{56A968}
\definecolor{lightgray}{HTML}{F1F2F6}
\definecolor{black1}{HTML}{000000}
\definecolor{black2}{HTML}{808080}
\definecolor{black3}{HTML}{cccccc}

\definecolor{rev1}{HTML}{F3F298} 
\definecolor{rev2}{HTML}{B2E0AE} 
\definecolor{other}{HTML}{C8C7FF} 

\journal{Progress in Aerospace Sciences}

\graphicspath{{./Figure/}}

\title{Machine Learning in Aerodynamic Shape Optimization}

\author[nus]{Jichao Li}
\author[um]{Xiaosong Du}
\author[um]{Joaquim R. R. A. Martins\corref{cor1}}
\address[nus]{National University of Singapore, Department of Mechanical Engineering}
\address[um]{University of Michigan, Department of Aerospace Engineering}
\cortext[cor1]{Corresponding author}

\makeatletter
\AtBeginDocument{\@ifpackageloaded{natbib}{\ifNAT@numbers\if@filesw\immediate\write\@auxout{\string\global\string\NAT@numberstrue}\fi\fi}{}}
\makeatother

\begin{document}

\begin{frontmatter}

\begin{abstract}

{Machine learning (ML) has been increasingly used to aid aerodynamic shape optimization (ASO), thanks to the availability of aerodynamic data and continued developments in deep learning.}
{We review the applications of ML in ASO to date and provide a perspective on the state-of-the-art and future directions.}
We first introduce conventional ASO and current challenges.
Next, we introduce ML fundamentals and detail ML algorithms that have been successful in ASO.
Then, we review ML applications to ASO addressing three aspects: compact geometric design space, fast aerodynamic analysis, and efficient optimization architecture.
In addition to providing a comprehensive summary of the research, we comment on the practicality and effectiveness of the developed methods.
We show how cutting-edge ML approaches can benefit ASO and address challenging demands, such as interactive design optimization.
Practical large-scale design optimizations remain a challenge because of the high cost of ML training.
{Further research on coupling ML model construction with prior experience and knowledge, such as physics-informed ML, is recommended to solve large-scale ASO problems.}


\end{abstract}


\end{frontmatter}

\setcounter{tocdepth}{3}
\tableofcontents

\section*{Nomenclature}

\subsection*{Acronyms}

{\renewcommand\arraystretch{1.0}
\noindent\begin{longtable*}{@{}l @{\quad=\quad} p{10cm}}
AD  & Algorithmic differentiation \\
ANN & Artificial neural network \\
ASM & Active subspace method \\
ASO & Aerodynamic shape optimization \\
BGD & Batch gradient descent \\
CFD & Computational fluid dynamics \\
CNN & Convolutional neural networks \\
CRM & Common Research Model, an aircraft model developed by NASA \\
DBN & Deep belief networks \\
DMD & Dynamic mode decomposition \\
DNN & Deep neural networks \\
DRL & Deep reinforcement learning \\
EGO & Efficient global optimization \\
FFD & Free-form deformation \\
GAN & Generative adversarial networks \\
GEK & Gradient-enhanced kriging \\
GMM & Gaussian mixture model \\
GTM & Generative topographic mapping \\
Isomap & Isometric (feature) mapping \\
KNN & $k$-nearest neighbors \\
LSTM& Long-short term memory \\
MCMC& Markov chain Monte Carlo \\
ME  & Mixture of experts \\
MGD & Mini-batch gradient descent \\
ML  & Machine learning \\
PCA & Principal component analysis \\
PINN & Physics-informed neural network \\
POD & Proper orthogonal decomposition \\
RBL & restricted Boltzmann machine \\
RL  & Reinforcement learning \\
RNN & Recurrent neural networks \\
SGD & Stochastic gradient descent \\
SVD & Singular value decomposition \\
SVM & Support vector machine \\
VAE & Variational autoencoder \\
XDSM & Extended design structure matrix \\
\end{longtable*}}

\subsection*{Symbols}

 {\renewcommand\arraystretch{1.0}
 \noindent\begin{longtable*}{@{}l @{\quad=\quad} p{14cm}}
 $\alpha$  & Angle of attack \\
 \(C_D\) & Drag coefficient \\
 \(C_L\) & Lift coefficient \\
 \(C_M\) & Moment coefficient \\
 \(M\) & Mach number \\
 \(Re\) & Reynolds number \\
 \end{longtable*}}

\section{Introduction}
\label{sec:intro}

Aerodynamic shape optimization (ASO)  {is an approach that is now available for aerodynamic designers to explore the design of lifting surfaces and other devices where lift and drag are important.}
 {Especially when} coupled with computational fluid dynamics (CFD), ASO is positioned to be an essential procedure in modern aircraft design and other design applications of CFD.
Aided by powerful high-performance computing (HPC) resources, CFD-based ASO  {advances} design of wings~\cite{Lyu2015b,LeDoux2015a,Osusky2015a,Kenway2016b}, tails~\cite{Chen2016b,Singh2016,Hao2018,Sanchez-Carmona2021,Muralikrishna2014}, engine nacelles~\cite{Song2007a,Gray2020a,Li2013850,Albert2013,Sasaki2011}, and other components~\cite{AbbasBayoumi2011,Stalewski2012,Welstead2012,Hashimoto2015,Batrakov2018,Liao2021a}.
 {In particular, ASO considerably} reduces the aircraft development's cycle time and improves the design's performance.

An important breakthrough in ASO is gradient-based design optimization with aerodynamic derivatives computed by the adjoint method~\cite{Jameson1988,Kenway2019a}.
Gradient information enables efficient and effective searching  {within} high-dimensional design space.
As reported by \citet{Lyu2014f}, gradient-free methods (such as the genetic algorithm~\cite{Goldberg:1989:GAS} and particle swarm algorithm~\cite{Kennedy1995}) tend to have quadratic or even cubic growth of function evaluations with respect to the increase of design-variable dimensionality.
In contrast, gradient-based methods tend to follow a more linear trend~\cite{Martins2021}.
The adjoint method accurately computes aerodynamic derivatives at a cost independent of the number of design variables~\cite{Kenway2019a}.
With such advances, gradient-based optimization has been prevailing in practical engineering applications~\cite{Gray2019a}.
Nevertheless, because of the iterative and costly simulation-based evaluations within optimization steps, ASO still cannot effectively satisfy some practical demands, such as fast interactive design optimization.

 {On the other hand,} machine learning (ML) has emerged as a tool to solve physical problems by learning from data~\cite{Bishop2006,Tieleman2012}.
 {Trained ML models are computationally efficient because they typically} work without the need  {of solving physical} governing equations.
 {Provided} with enough data,  {well trained} ML models could be accurate and general for predictions and descriptions of the underlying physics.
Traditional ML approaches (such as kriging~\cite{Krige1951,Jeong2005,Keane2011,Koziel2016b}), have been successfully used in various engineering fields~\cite{Queipo2005a,Zhao2013,Hu2016b,Noack2019AKA}.
 {In the meantime,} the development of deep learning~\cite{LeCun2015doi,Zhang2018d,Emmert-Streib2020,Alzubaidi2021} enables large-scale practical tasks in broad areas, such as computer vision~\cite{Wu2017b,Voulodimos2018,Mahony2019DeepLV}, medical image analysis~\cite{Shen2017MIA,Panchal2019,MAIER201986,Liu2021MIA,Puttagunta2021}, computational mechanics~\cite{Oishi2017,Brunton2020,Bahri2020,Kunin2021neural}, and aerospace engineering~\cite{Armes2013,Rengasamy2018,Podorozhniak2019,Brunton2021a,DANGUT2021127}.

The key to utilizing ML is data.
 {Fortunately, there are some aerodynamic data available to train ML.
Despite a limited amount, historical and current designs that have proved to perform well in practical applications contain valuable knowledge for ML to learn from.}
Furthermore, new data can be generated as needed through CFD simulations.
ML has the potential to speed up ASO by leveraging these aerodynamic data.
 {A popular} approach is to construct  {data-driven} surrogate models (also known as metamodels) to replace the costly simulations~\cite{Queipo2005a}.
 {In contrast, another rising surrogate modeling branch} is to train neural networks  {respecting} given  {physical} laws~\cite{vonRueden2021}, which is known as physics-informed neural networks (PINNs)~\cite{Raissi2019} ( {see review papers on PINNs}~\cite{Karniadakis2021a,Cai2021,Viana2021}).

 {Nevertheless, for design purposes, the training data must include samples that adequately represent the design space.
More specifically, it is preferable to have enough samples that are well distributed in all dimensions (design variables) to fill the design space.}
 {This leads to high demand for training data and makes it intractable to apply ML to large-scale high-dimensional ASO problems.
A successful application of ML in practical ASO problems is usually based on an in-depth analysis of the specific difficulty and a combination with conventional methods such as adjoint.}

Recent developments in ML (especially in deep learning) have improved the scope and effectiveness of ASO.
 {There is a need for a comprehensive review of these developments that makes connections between the various approaches and puts them into context.
The present review addresses this need; we summarize the ML approaches and their applications to ASO, assess their effectiveness, and comment on the prospects.
This review should stimulate further development in this field.
}

The remainder of this paper is organized as follows.
First, we review the state-of-the-art ASO and the existing challenges in Sec.~\ref{sec:cfd-aso}.
Then, we introduce commonly-used ML techniques and algorithms that have been successfully introduced into ASO in Sec.~\ref{sec:ml}.
Readers who are already familiar with ML approaches can skip this section.
In Sec.~\ref{sec:aso}, we present a comprehensive review of the ML applications in ASO with an emphasis on the geometric design space, aerodynamic evaluation, and optimization architecture.
 {Then we end this paper with} conclusions and outlooks in Sec.~\ref{sec:conclusion}.

\section{Aerodynamic Shape Optimization}
\label{sec:cfd-aso}

In this section, we present a general ASO process and discuss the existing challenges that lead to the need for ML approaches.
  {Aerodynamic design problems with significant configuration changes~\cite{Hwang2012c}, such as changing the wing span and sweep~\cite{Brooks2018a,Bons2020d}, are multidisciplinary optimization problems that involve other disciplines, such as structural design and stability and control~\cite{Martins2013,Gray2019a}.
In this review, we only address aerodynamic design for a fixed configuration, focusing on local sectional shape design.
}

\subsection{General Process}

Typical ASO problems have a well-defined objective function, constraints, and design variables.
The objective function can be the aerodynamic performance to be minimized (such as the drag coefficient, $C_D$) or maximized (such as the lift-to-drag ratio) at one or multiple flight conditions (usually defined by the Mach number and Reynolds number).
There are two main types of constraints in ASO: geometric constraints (such as thickness, area, and volume) and aerodynamic constraints (such as the lift coefficient, $C_L$, and the moment coefficient, $C_M$) at certain flight conditions.
The design variables are mainly defined by a shape parameterization method and can also include the angle of attack at each flight condition to satisfy the lift constraint.

There are various approaches to parameterizing the aerodynamic shape. 
For airfoil parameterization, common methods include the NACA airfoil definition, PARSEC~\cite{Sobieczky1999a}, Hicks--Henne bump functions~\cite{Hicks1978}, class shape transformation (CST)~\cite{Kulfan2008},  free-form deformation (FFD)~\cite{Sederberg1986,Hsu1992}, and B\'ezier curves.
Many of these parametrizations have shown to be special cases of B-spline curves~\cite{Rajnarayan2018a}.
Using the NACA airfoil definition and the PARSEC method consistently produces reasonable airfoil shapes, but these approaches do not have much geometric freedom because they involve a limited number of design variables.
Parameterization methods such as CST and FFD can provide more geometric freedom by increasing the number of design variables.
From the point of view of aviation airlines, obtaining the optimal aerodynamic performance is of high priority.
Thus, a large number of design variables have been a necessity in ASO efforts.
Researchers have found that tens of design variables are required for two-dimensional airfoils~\cite{He2019c} and hundreds of shape design variables are required for three-dimensional wing problems~\cite{Lyu2015b}.

The high computational cost of aerodynamic analysis and the high dimensionality of the geometric design space are two compounding challenges in ASO.
Gradient-based optimization algorithms are the most suitable for addressing problems with these two characteristics because they are efficient in searching high-dimensional design spaces.
The derivatives of the aerodynamic functions of interest with respect to all design variables are required when using gradient-based optimization.
In the past decades, the ASO community focused on how to compute these derivatives efficiently and accurately~\cite{Jameson1988,Jameson2003,Peter2010a,Kenway2019a,Martins2022a}.
With the development of the adjoint method, especially the Jacobian-free implementation using source code transformation algorithmic differentiation~\cite{Kenway2019a}, CFD-based ASO via an adjoint-enabled gradient-based optimization algorithm has become a popular choice in aircraft design~\cite{Lyu2015b,Osusky2015a,Economon2016a,Yildirim2019a,Gray2020a}.
We explain the general process of this method using the open-source MACH-Aero framework~\footnote{\url{https://github.com/mdolab/MACH-Aero}}~\cite{Martins2022a} as an example, but the overall process and steps are similar in other CFD-based ASO frameworks.

\begin{figure}[h]
\centering
\includegraphics[width=\linewidth]{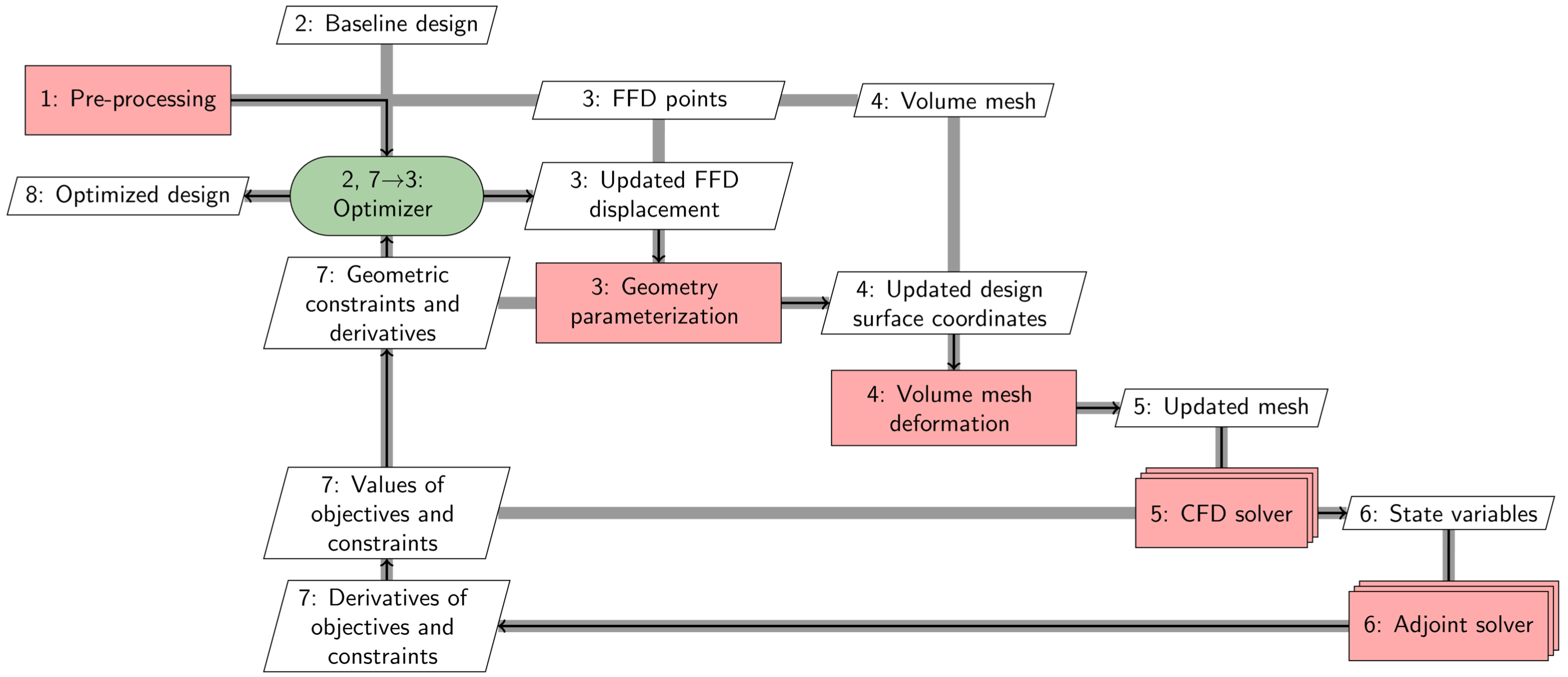}
\caption{The workflow for the MACH-Aero framework~\cite{Martins2022a} is representative of the workflow of CFD-based ASO. The data dependencies and process are shown using XDSM~\cite{Lambe2012a}.
The diagonal nodes are process components, and the off-diagonal nodes are the data transferred between components.
The thick gray lines represent the data flow, while the black lines represent the process flow.
}
\label{fg_MACH}
\end{figure}

The ASO workflow of MACH-Aero is shown in Fig.~\ref{fg_MACH} using an extended design structure matrix (XDSM) representation~\cite{Lambe2012a}.
The diagonal components represent the main procedures in the CFD-based optimization.
In the pre-processing procedure, the CFD mesh and the initial geometric parameterization (such as the FFD control box) of the baseline shape are established.
Then, the objective function, constraints, and bounds of design variables are defined in the optimizer via a Python interface.
The optimal aerodynamic shape is iteratively solved.
In each iteration, the gradient-based optimizer determines how to update the design variables;
based on the updated geometric design variables, the aerodynamic shape is deformed, and the geometric constraints and derivatives are evaluated;
the volume mesh is deformed to perform CFD analysis of the new aerodynamic shape, and the adjoint solver computes the aerodynamic derivatives;
based on the objective and constraint function values and their derivatives, the optimizer determines the search direction and step length, which yields the design variables for the next iteration.

In MACH-Aero, all the components are wrapped using a Python interface for modularity and ease of use.
The gradient-based optimizer (such as SLSQP~\cite{Kraft1988} and SNOPT~\cite{Gill2005a}) is provided through the pyOptSparse interface~\cite{Wu2020a}.
The geometric parametrization can use either FFD~\cite{Kenway2010b} or OpenVSP~\cite{Hahn2010a} implemented in pyGeo~\footnote{\url{https://github.com/mdolab/pygeo}}.
The volume mesh deformation is based on the inverse distance method provided by IDWarp~\footnote{\url{https://github.com/mdolab/idwarp}}~\cite{Secco2021a}.
ADflow~\footnote{\url{https://github.com/mdolab/adflow}}~\cite{Mader2020a} and DAFoam~\footnote{\url{https://github.com/mdolab/dafoam}}~\cite{He2018b} are available to perform aerodynamic evaluations, and both include an efficient adjoint solver~\cite{Kenway2019a,He2020b} to compute the aerodynamic derivatives.

There are other frameworks for CFD-based ASO, and they have a similar workflow.
For example, in SU2~\cite{Economon2016a}~\footnote{\url{https://su2code.github.io/}}, ASO follows a similar workflow, and different components are wrapped by a Python script.
The SLSQP optimizer of the SciPy package is the default optimizer in SU2.
SU2\_GEO provides the FFD parameterization and evaluates geometric constraints (both values and derivatives).
SU2\_CFD performs both direct and adjoint computations, and the aerodynamic derivatives are further computed in SU2\_DOT.
SU2\_DEF deforms the CFD mesh after the design variables are updated.
 {Similar CFD-based optimization frameworks have been developed by other research groups, such as M\"{u}ller~\cite{Mueller2005,Xu2014a,Mykhaskiv2017a,Xu2017b}, Nadarajah~\cite{bisson2015adjoint,Shi-Dong2017a,Poirier2016,Khayatzadeh2012a,Castonguay2007a}, and Zingg~\cite{Buckley2010,Buckley2013,Lee2015a,Rashad2016b,Reist2020a}}.

\subsection{Existing Challenges}
\label{sec:cfd-asoChallenges}

Enhanced by the efficient adjoint method, high-fidelity CFD-based ASO has been applied in a wide range of ASO problems, such as airfoil design~\cite{Nemec2004,Buckley2010,He2019c}, wing design~\cite{Jameson1998,Dwight2009,Leung2012,Lyu2015b,Shi-Dong2017a,Xu2017b}, and wing-body-tail configuration design~\cite{Chen2016b,Meheut2016,Shitrit2021}.
Nevertheless, there are still some challenges calling for advanced solutions.

First, most aerodynamic shape parameterization methods are inefficient and lead to high-dimensional design space having many regions with abnormal shapes, which brings unnecessary difficulties to ASO.
When using conventional parameterization methods, a large number of shape design variables are required to ensure convergence to the real optimal design.
As the number of variables increases, the design space includes many abnormal shapes, such as wavy airfoil surfaces, which are usually evaluated in intermediate optimization steps.
With such abnormal aerodynamic shapes, it is time-consuming or even impossible for CFD simulations to converge, potentially leading to optimization failure.
Although efforts such as the approximate Newton--Krylov algorithm~\cite{Yildirim2019b} have been made to improve convergence capability for a wide range of shapes, intermediate designs with abnormal aerodynamic shapes are still undesirable in ASO~\cite{He2019c,Li2021d}.
An ideal solution is to develop a parameterization using fewer design variables that excludes abnormal shapes from the geometric design space.

Second, multiobjective CFD-based design optimization is typically too expensive.
Multiobjective optimization is of great interest in aerodynamic design because there are usually multiple metrics of interest~\cite{Nemec2004,Leifsson2016}.
However, there is no efficient optimization algorithm to solve high-dimensional multiobjective design problems.
A commonly used approach in ASO is to convert the design problem into a series of single-objective optimization problems and solve them one by one, which is time-consuming~\cite[Ch. 9]{Martins2021}.
This approach typically finds only a few points in the Pareto frontier with no guarantee that they are uniformly distributed.
Most multiobjective applications have only considered low-dimensional design problems, such as airfoil shape design~\cite{Nemec2004doi} and wing platform design~\cite{Obayashi2005}, with a few exceptions~\cite{Kenway2014c,Bons2020b}.

Third, CFD-based optimization has mostly been deterministic, ignoring aleatory uncertainties in operating conditions or geometry changes due to wear and tear and manufacturing inaccuracies~\cite{Schillings2011}.
Ignoring these uncertainties may in practice lead to unexpected performance loss.
To improve the robustness of objective metrics and reliability on design constraints, ASO should be performed with reasonable consideration of the uncertainties.
This can be done by increasing the number of design points~\cite{Chai2018,Li2020e} or using stochastic metrics~\cite{Ulker2021}, both of which lead to a significant increase in the computational cost.
Therefore, most studies in robust design~\cite{Panzeri2018a} or reliability-based design~\cite{Papadimitriou2018} are merely on two-dimensional airfoil shape design~\cite{Huyse2002,Wu2017a} or three-dimensional configuration design considering the uncertainty of several operating parameters such as the Mach number and lift coefficient~\cite{Liem2017a,BahamondeJacome2017,Liem2015a,Hwang2019a,Suprayitno2020}, which is still far from satisfying the industrial demand.

Fourth, there is still a lack of techniques to utilize different aerodynamic data and models together effectively.
There are already a series of typical CFD models of different fidelity existing for ASO by solving the Euler equations and Reynolds-averaged Navier--Stokes (RANS) equations.
Lower-fidelity CFD models are also available through coarsening high-fidelity CFD meshes~\cite{Peherstorfer2018a,Han2020}.
Besides, with the advances in experimental aerodynamics equipment, an increasing amount of experimental data is available.
Ideally, we would use experimental data complemented with multi-fidelity numerical simulations to guide design optimizations.

Fifth, there are discontinuous ASO problems that are difficult for the current gradient-based frameworks to handle.
For example, laminar-turbulent transition dominates the aerodynamic performance in low-Reynolds-number aerodynamic shape optimization.
To capture the flow transition, the simulation model includes functions that do not have continuous derivatives, which introduces difficulties to the adjoint implementation~\cite{Shi2020a,Halila2020d,Halila2022}.
This issue in evaluating aerodynamic derivatives causes optimization difficulties.
The high dimensionality of the design space further exacerbates this issue.
Finally, discontinuous aerodynamic objective functions are unsuitable for gradient-based optimization.

Lastly, interactive design optimization where aerodynamic evaluations run near instantaneously is a desired capability that enables designers to rapidly evaluate the influence of different variables, constraints, and design requirements.
However, the high computational cost of high-fidelity simulations makes the interaction cycle too slow for it to be considered interactive.

ML techniques are useful in addressing these demands or challenges, especially in developing more compact geometric parameterization, faster aerodynamic evaluations, and more efficient optimization architectures.
These approaches and corresponding applications will be presented and reviewed in the remainder of this paper.

\section{Machine Learning Methods}
\label{sec:ml}

ML approaches enable computers to learn from data.
\citet{Mitchell1997} defined ML as a computer program ``learning from experience E with respect to some task T and some performance measure P, if its performance on T, as measured by P, improves with experience E."
ML has emerged as a cutting-edge tool in modern life due to the promising advantages over traditional programming techniques~\cite{Gron2017}.
ML algorithms often simplify code and perform better on problems whose solutions require long lists of rules.
In addition, ML can adapt to updated data sets after establishing detecting rules or policies.
Moreover, ML can discover complex implicit data patterns of complex problems, which lead users to richer insights.

ML algorithms are classified into broad categories via the following metrics:
algorithms simply comparing new data with training data or constructing a model of training data patterns to make predictions;
algorithms with or without human supervision during model training;
algorithms learning through batch training or incrementally on the fly.
Most ML approaches use constructed models for predictions, while instance-based algorithms, such as $k$-nearest neighbors (KNN), are also popular.
In this section, we will focus on ML approaches that have achieved success in ASO.
Specifically, we will introduce the four major categories: supervised learning, unsupervised learning, semi-supervised learning, and reinforcement learning.
We will detail neural network fundamentals in the remainder of this section and present a few recent artificial neural network (ANN)~\cite{LeCun2015} architectures, which have advanced the state-of-the-art in various fields besides the aerospace industry.
 {This section provides a convenient self-contained summary of the ML approaches mentioned in Sec.~\ref{sec:aso}.
Readers who are already familiar with these approaches may skip this section.}

\subsection{Supervised Learning}
\label{sec:mlSL}


Supervised learning models are trained by data sets containing inputs and the corresponding model observations (also called ``labels'' or ``targets'')~\cite{Singh2016ARO,Nasteski2017}.
Classification and regression are the most common supervised learning tasks.
Classification refers to constructing models to separate the input data into discrete categories, such as face detection and handwriting recognition.
In contrast, regression methods directly model the mapping relationship between inputs and continuous model observations.
Regression fits a wide range of real-world problems, such as predicting stock price and aircraft performance.
In this section, we focus on the supervised learning methods, which have been successfully introduced to ASO.

\subsubsection{$k$-Nearest Neighbors}
\label{sec:mlKNN}

\begin{figure}[h]
\centering
\includegraphics[width=\linewidth]{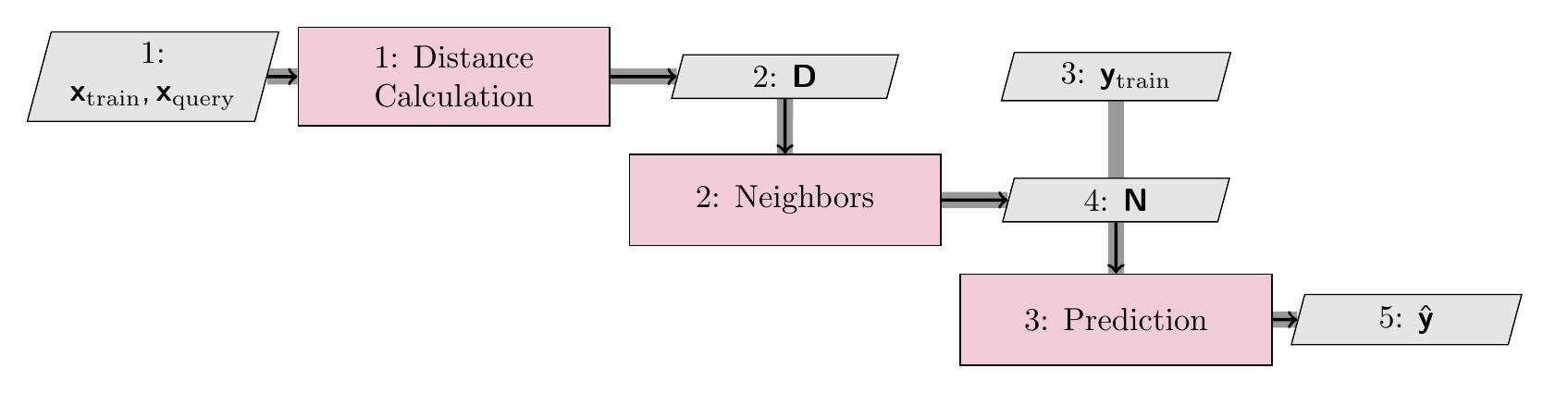}
\caption{KNN consists of three main steps:
(1) calculate the distance ($\mathbf{D}$) between query data ($\mathbf{x}_\text{query}$) and training data ($\mathbf{x}_\text{train}$);
(2) select $k$ nearest neighbors ($\mathbf{N}$) to $\mathbf{x}_\text{query}$ based on the distance;
(3) predict labels ($\hat{\mathbf{y}}$) through voting of the $\mathbf{N}$ neighbor labels ($\mathbf{y}_\text{train}$).
The diagonal nodes are process components, and the off-diagonal nodes are the data transferred between components.
The thick gray lines represent the data flow, while the black lines represent the process flow.
}
\label{fig:knn}
\end{figure}

\begin{figure}[h]
\centering
\includegraphics[width=0.55\linewidth]{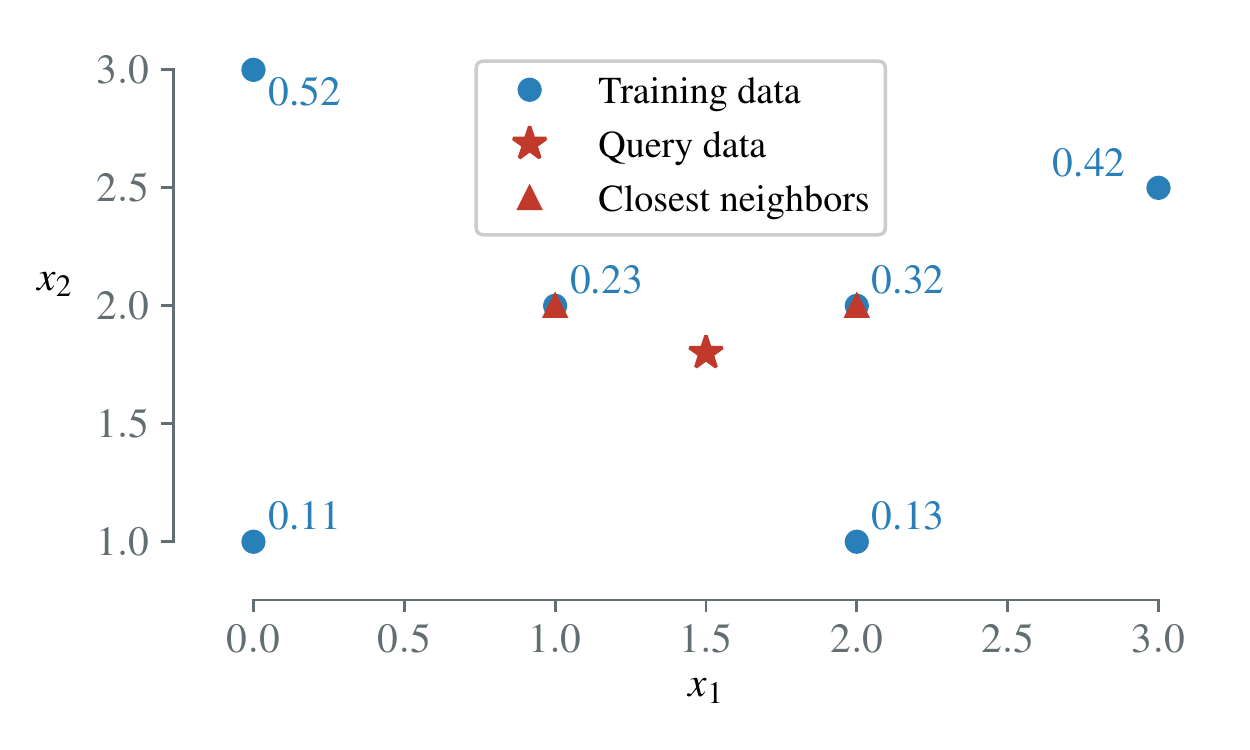}
\vspace{-0.5cm}
\caption{KNN regression example using two closest nearest neighbors based on Euclidean distance: the predicted label equals is $(0.23 + 0.32) / 2 = 0.275$.
}
\label{fig:knn_example}
\end{figure}

KNN~\cite{Cunningham2020,Taunk2019} classifies categories or predicts labels based on the distance between an untried sample point and its $k$ nearest training data neighbors (Fig.~\ref{fig:knn}).
Distance calculation methods~\cite{Walters-Williams2010} include Euclidean, Manhattan, Minkowski, and Hamming algorithms.
The most commonly used Euclidean calculation is\begin{equation}
    \delta = \sqrt{ \sum^{m}_{i=1}(x_{1,i} - x_{2,i})^2 },
\end{equation}
where $\delta$ is the distance, $x_1$ and $x_2$ are two arbitrary data neighbors, and $m$ is the input dimension.
Classification KNN analyzes the categories of the neighbors and assigns the category for the query data based on a majority vote, while regression KNN makes predictions based on the mean values of the neighbors (Fig.~\ref{fig:knn_example}).
Instead of assigning uniform weights, the prediction step can also use biased weights based on the distance such that the closer neighbors contribute more.
Therefore, KNN is an instance-based ML model since we do not need to construct and train a model.

We summarize the advantages and disadvantages as follows.
On the one hand, KNN is simple to implement, flexible to classification and regression problems, and does well with sufficient representative data.
On the other hand, a significant challenge of KNN is how to determine the number of neighbors, $k$.
Smaller $k$ values can be noisy and provide unstable decision boundaries, while larger values lead to smoother decision boundaries but may not represent the output space well.
One simple suggested solution is to use $k = \sqrt{n}$, where $n$ is the number of training data points.
However, the value is still not guaranteed to be optimal.
The simplicity and overall good performance of KNN attract attention from the aerospace industry.
For example, \citet{Wang2019_knn} collected sufficient flight data for rotary-wing drones and managed to realize wind speed estimation using KNN.
They completed a parametric study for the optimal $k$ value produced a 3.3\% average relative error compared with experimental results.

\subsubsection{Support Vector Machines}
\label{sec:mlSVM}

\begin{figure}[h]
\centering
\includegraphics[width=\linewidth]{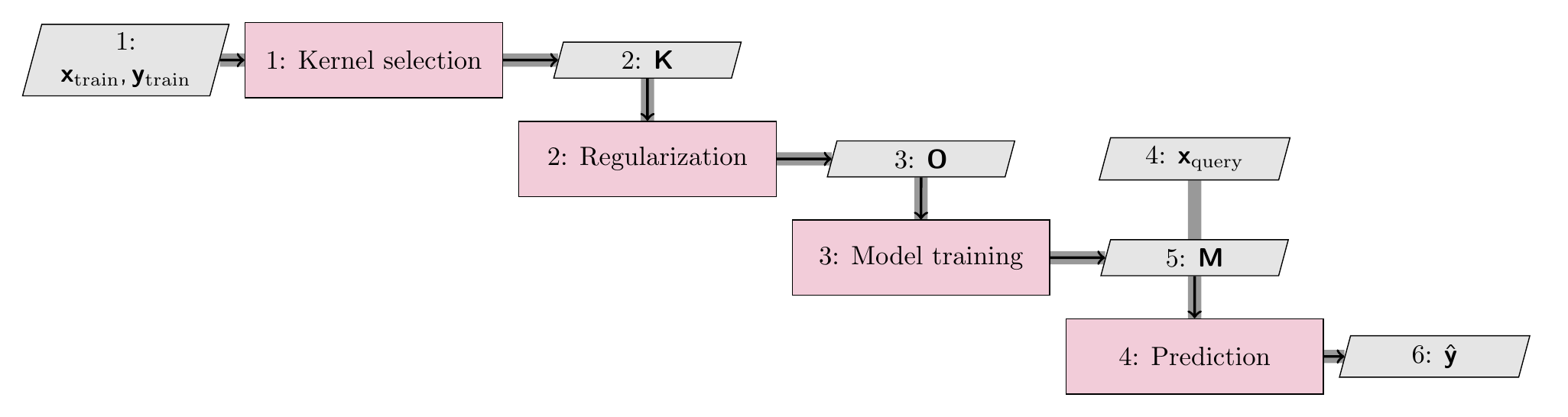}
\caption{SVM first requires users to select a kernel function ($\mathbf{K}$) to transform training data set ($\mathbf{x}_\text{train}$, $\mathbf{y}_\text{train}$) for efficient computation in high-dimensional implicit feature space.
Then users need to select regularization terms for the objective function ($\mathbf{O}$) to determine the penalty on misclassified samples.
The training process aims at the SVM model ($\mathbf{M}$) with optimal hyperplanes, and then prediction tasks ($\hat{\mathbf{y}}$) on query samples ($\mathbf{x}_\text{query}$) can be performed.
}
\label{fig:svm}
\end{figure}

\begin{figure}[h]
\centering
\includegraphics[width=0.4\linewidth]{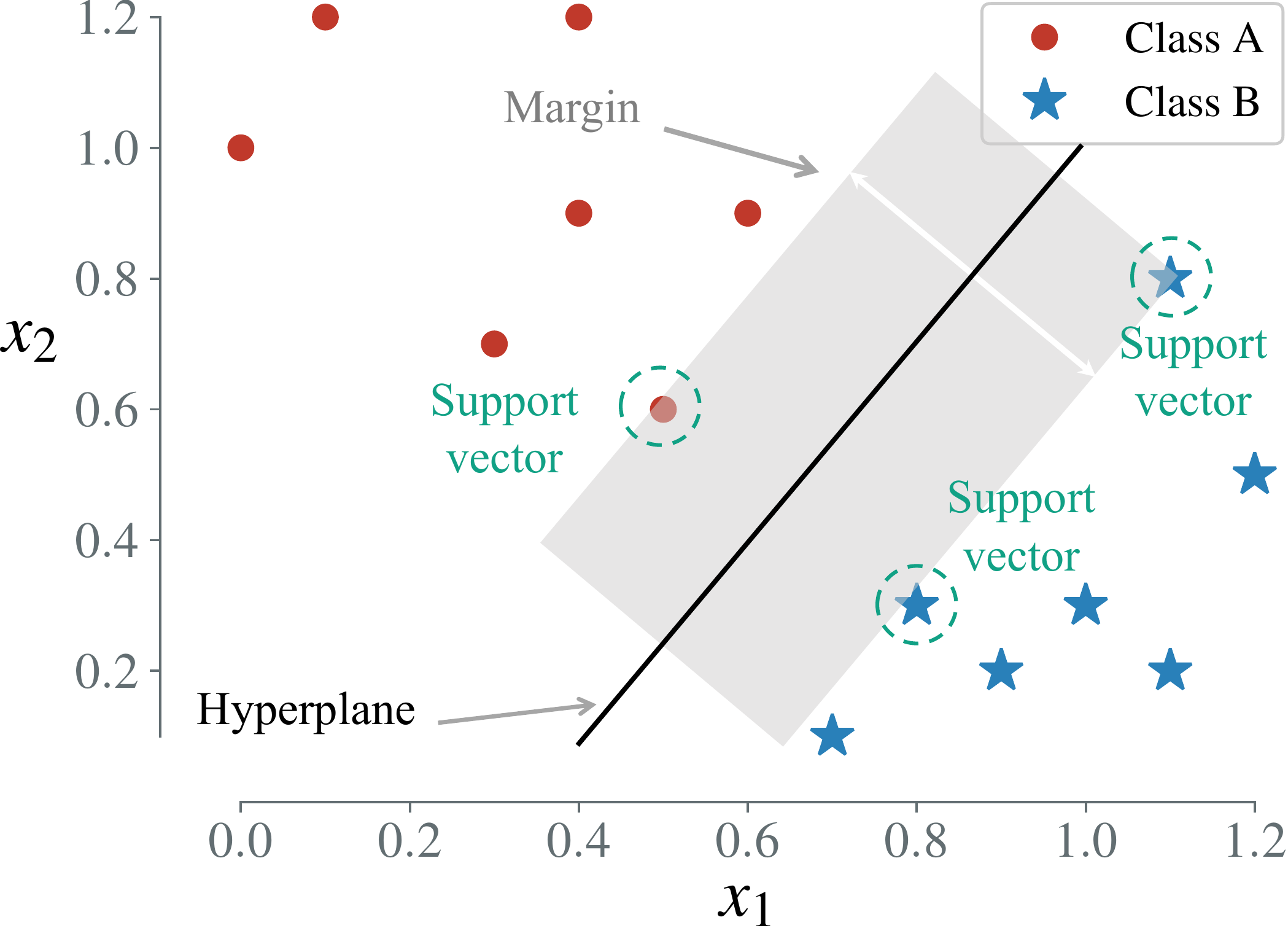}
\caption{SVM classification example, where the optimal hyperplane (a line for two-class classification) has the largest margin.
The data samples that fall on the margin boundaries are support vectors.
}
\label{fig:svm_example}
\end{figure}

Support vector machine (SVM)~\cite{Hearst1998,Guenther2016} handles both classification and regression tasks by constructing hyperplanes in the model-response space.
An optimal hyperplanes-based separation has the largest distance to the nearest training data points of any class because the larger the margin, the lower the generalization error of the classification (Fig.~\ref{fig:svm}).
The training data points that fall on the margin boundaries are called the support vectors (Fig.~\ref{fig:svm_example}).

SVM in classification problems is called a support vector classifier (SVC)~\cite{Tong2000,Cervantes2020}.
Given $n$ training data $X \in \mathds{R}^{n \times m}$ and corresponding two-class categories $Y \in [-1, 1]$, the goal is to determine the unknown parameters $w \in \mathds{R}^m$ and $b \in \mathds{R}$ and predict the correct sign of $w^\intercal \phi(x) + b$ for the most samples.
Therefore, the SVC primal problem is:
\begin{equation}
\begin{aligned}
    & \min \limits_{w, b, \zeta} \frac{1}{2}w^\intercal w + C \sum \limits_{i=1}^n \zeta_i , \\
    & \text{subject to} \\
    & y_i \cdot (w^\intercal\phi(x_i)+b) \geq 1 - \zeta_i, \\
    & \zeta_i \geq 0, i = 1, \ldots, n.
\end{aligned}
\end{equation}
Under such formulation, the margin is maximized by minimizing $w^\intercal w$ and the penalty $C \sum \limits_{i=1}^n \zeta_i$.
Specifically, the positive-value $\zeta_i$ adds a penalty to the above objective function if the $i\text{th}$ sample is within the hyperplane boundaries or misclassified, and the constant $C$ controls the strength of the penalty.

Lagrangian duality principle~\cite{Rong2020,Hager1976} simplifies the SVC primal problem as
\begin{equation}
\begin{aligned}
    & \min \limits_{\alpha} \frac{1}{2} \alpha^\intercal Q \alpha - e^\intercal \alpha, \\
    & \text{subject to} \\
    & y^\intercal \alpha = 0, \\
    & 0 \leq \alpha_i \leq C, i=1, \ldots, n,
\end{aligned}
\end{equation}
where $e$ is the vector of all ones, $Q$ is an $n$ by $n$ positive semi-definite matrix, $Q_{i,j} \equiv y_i y_j K(x_i, x_j)$, $K(x_i, x_j) = \phi(x_i)^\intercal \phi(x_j)$ is the kernel, and $\alpha$ is a dual coefficient vector upper-bounded by $C$.
The dual problem is a quadratic function subject to linear constraints, which quadratic programming algorithms can solve efficiently.
Once we construct the SVC by solving the above optimization problem, the predicted classification on a given sample $x^\prime$ becomes:
\begin{equation}
    y^\prime = \sum \limits_{i \in SV} y_i \alpha_i K(x_i, x^\prime ) + b,
\end{equation}
where $SV$ is the support vector set.
We only need to sum over the support vectors because the dual coefficients $\alpha$ are zeros for the other training data.

Similarly, SVM used for regression problems is referred to as support vector regression (SVR)~\cite{Stoean2006,Jap2015}.
Given $n$ training predictors $x_i \in \mathds{R}^m$ and corresponding continuous targets $y_i \in \mathds{R}^p$, the SVR primal optimization problem becomes:
\begin{equation}
\begin{aligned}
    & \min \limits_{w, b, \zeta, \zeta^*} \frac{1}{2}w^\intercal w + C \sum \limits_{i=1}^n (\zeta_i + \zeta_i^*), \\
    & \text{subject to} \\
    & y_i - w^\intercal\phi(x_i) - b \leq \epsilon + \zeta_i, \\
    & w^\intercal\phi(x_i) + b \leq \epsilon + \zeta_i^*, \\
    & \zeta_i, \zeta_i^* \geq 0, i = 1, \ldots, n.
\end{aligned}
\end{equation}
The objective function is penalized when the predictions are at least $\epsilon$ away from actual targets.
The penalty is determined by $\zeta_i$ or $\zeta_i^*$ depending the predictions are above or below the $\epsilon$ tube.
A similar Lagrangian duality simplification can be used to obtain the following dual optimization problem:
\begin{equation}
\begin{aligned}
    & \min \limits_{\alpha, \alpha^*} \frac{1}{2} (\alpha - \alpha^*)^\intercal Q (\alpha - \alpha^*) + \epsilon e^\intercal (\alpha + \alpha^*) - y^\intercal(\alpha - \alpha^*), \\
    & \text{subject to} \\
    & e^\intercal (\alpha - \alpha^*) = 0, \\
    & 0 \leq \alpha_i, \alpha_i^* \leq C, i=1, \ldots, n,
\end{aligned}
\end{equation}
The SVR prediction at $x^\prime$ is

\begin{equation}
    y^\prime = \sum \limits_{i \in SV} (\alpha_i - \alpha_i^*) K(x_i, x^\prime) + b.
\end{equation}

As a commonly used ML approach, SVM mainly has the following advantages and disadvantages~\cite{Auria2008}.
One of the most appealing aspects of SVM is its versatility, which fits SVM well into a wide range of real-world applications~\cite{Attewell2015data,Byun2002applications}.
SVM is robust and efficient to the observations that are far away from the hyperplane since SVM only considers support vectors (data samples that fall on hyperplane margin boundaries).
SVM successfully completes classifications with many classes even with few training samples in the data est.
SVM adapts to nonlinear decision/classification boundaries through various kernel functions and produces solutions even when the data are not linearly separable.
SVM provides a unique solution instead of local minima found by ANN.
SVM may have better classification performance for imbalanced data since they only rely on the support vectors~\cite{Attewell2015data}.
On the other hand, SVM can be computationally intensive and costly in memory especially when the data set is large.
The selection of an SVM kernel is tricky since a random choice may have negative effects on the performance~\cite{Horvath2003neural}.

We see a range of SVM applications in support of aerodynamic analysis and optimization.
For example, \citet{Perez2016} used SVM as a surrogate model to estimate objective function (lift-drag ratio), in combination with an evolutionary algorithm for ASO problems.
SVM surrogate managed to address 14 input parameters for a two-dimensional case and 36 inputs for a three-dimensional case.
These successes make SVM a competitive surrogate candidate for ASO applications.

\subsubsection{Decision Tree and Random Forest}
\label{sec:mlDTRF}

\begin{figure}[h]
\centering
\includegraphics[width=\linewidth]{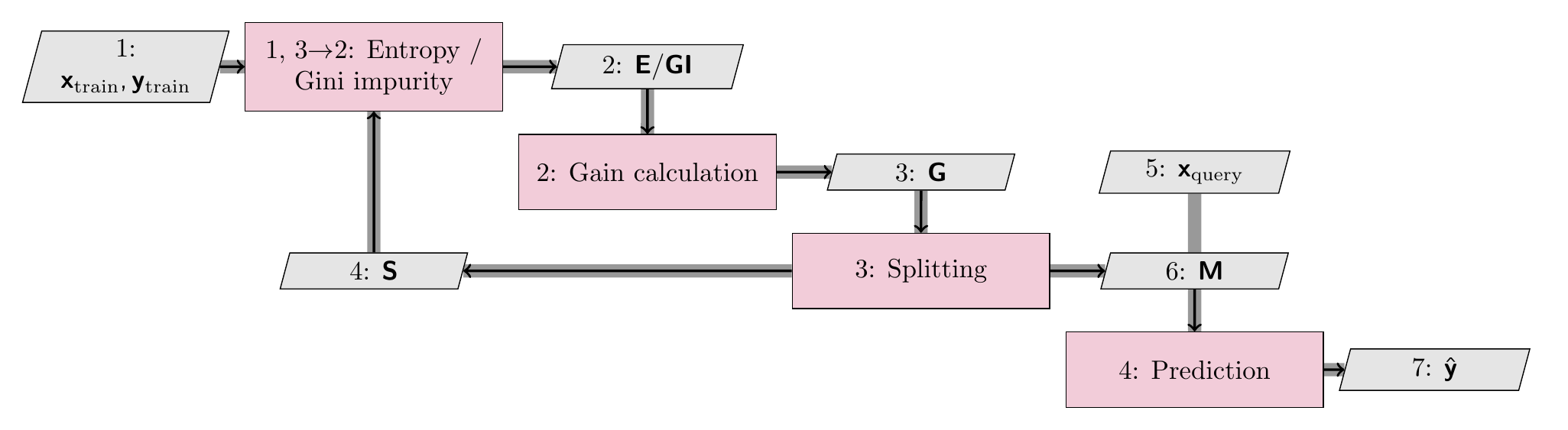}
\caption{Decision trees compute the entropy ($\mathbf{E}$) or Gini impurity ($GI$) before the split and after the tentative splits to get the information gain or Gini gain ($\mathbf{G}$).
We draw the decision tree's split ($\mathbf{S}$) that has the highest $\mathbf{G}$ value and iterate this splitting process until all attributes are set as tree nodes.
Then, we can use the decision tree model ($\mathbf{M}$) for classification or regression tasks ($\hat{\mathbf{y}}$) on query data samples ($\mathbf{x}_\text{query}$).
}
\label{fig:decisionTree}
\end{figure}

\begin{figure}[h]
\centering
\includegraphics[width=\linewidth]{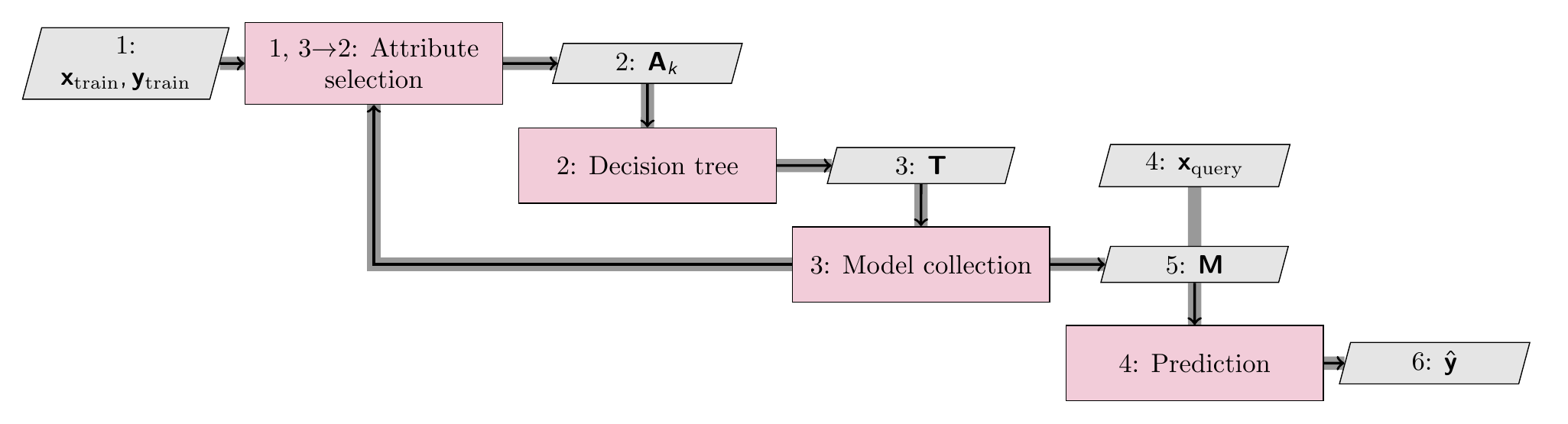}
\caption{Random forests randomly select $k$ attributes ($\mathbf{A}_k$) out of the total attributes and construct a decision tree ($\mathbf{T}$) as introduced in Fig.~\ref{fig:decisionTree}.
We collect the constructed decision tree and iterate until we have the pre-defined number of trees.
In the end, the random forest model ($\mathbf{M}$) can predict labels ($\hat{\mathbf{y}}$) on any given query data ($\mathbf{x}_\text{query}$).
}
\label{fig:randomForest}
\end{figure}

A decision tree~\cite{Rokach2014,Safavian1991} is a supervised learning method that works for both classification and regression tasks (Fig.~\ref{fig:decisionTree}).
A random forest~\cite{Breiman2001,Fawagreh2014} assembles a series of decision trees that contribute to the classification/regression problems from broader aspects (Fig.~\ref{fig:randomForest}).
A decision tree follows the principle of partitioning data by iteratively asking questions.
These questions are crucial for a decision tree because the more informative the questions, the better the model's predictive performance.
Gini impurity~\cite{Gini1921} and entropy~\cite{Shannon1948} are two main ways to quantify a split question's quality.

Gini impurity is a measurement of the likelihood that the model incorrectly labels a new sample if the label decision is randomly made according to the distribution of existing labels in the data set~\cite{zimmermann2019asymmetric}.
Following the definition, the Gini impurity is mathematically formulated as
\begin{equation}
    GI = \sum \limits_{i=1}^{N_C} P(i) (1 - P(i)),
\end{equation}
where $N_C$ is the total number of classes and $P(i)$ is the probability of the $i$th class in the current data set.
Thus, if there is only one class in the current data set, the Gini impurity is zero.
In contrast, the more equally distributed categories the data set has, the higher the Gini impurity is.
The split quality can be determined through weighting impurity of each branch by
\begin{equation}
    GI_\text{split} = \sum \limits_{i=1}^{N_B} w_i GI_i,
\end{equation}
where $N_B$ is the number of branches after split, $GI_i$ is the Gini impurity in the $i$th branch, and $w_i$ equals the number of training data in $i$th branch divided by the total amount.
Thus, the Gini gain is defined as $GI - GI_\text{split}$ and higher Gini gain represents better split.

Entropy is a measurement of randomness or impurity in a data set~\cite{zimmermann2019asymmetric}.
In general, the more randomness the data has, the higher the entropy is.
The entropy is mathematically defined as
\begin{equation}
    E(x) = - \sum \limits_{k=1}^{N_F} \Big(P(x=k) \log_2 \big(P(x=k)\big) \Big),
\end{equation}
where $N_F$ is the total number of features, $P$ is the probability of a target feature.
Information gain is the metric based on entropy to quantify a split quality; it is the difference between the entropy before the split and the entropy weighted by the number of training data samples in each branch after the split.

Gini impurity favors larger partitions and is easy to implement, whereas entropy favors smaller partitions with distinct values.
These two methods make decision trees powerful and easy to implement and are suitable for a mixture of data types (\emph{e.g.}, continuous, categorical).

One decision tree, however, is prone to overfitting especially when the tree is particularly deep, so we normally assemble multiple decision trees for better performance, which leads to random forests.
A random forest combines a series of decision trees that work as parallel estimators.
In classification problems, the random forest method makes the final prediction based on the majority vote of the results from each decision tree.
In regression problems, the final prediction is the mean value of the results from each tree.

Random forests have better performance and lower overfitting risk than a single decision tree.
The success of random forests mainly results from using uncorrelated decision trees by bootstrapping and feature randomness.
Bootstrapping involves repeatedly drawing sample data with replacement from a data source to avoid overfitting and improve the model stability in the ML field.
Feature randomness means randomly selecting features for each decision tree within a random forest.
Random forests extend decision trees and show outstanding performance but do not scale well with large-scale input dimensions.
\citet{Darari2019} constructed a random forest surrogate to support design space exploration and extracted design parameter importance for a better understanding of the design space.
\citet{Dube2020} compared a range of ML approaches, including kriging, decision trees, linear regression, random forests, and ANN for automotive drag predictions and concluded that ANN gave the best performance.

\subsubsection{Traditional Surrogate Models}
\label{sec:mlSurro}


Surrogate models, also known as metamodels, are a special type of supervised ML algorithms applied in engineering fields~\cite{Leifsson2011surrogate,Leifsson2013surrogate,Peherstorfer2018a}.
In this section, we refer to traditional surrogate models as the algorithms that have been introduced to ASO in early works before deep learning attracted more attention.
In particular, surrogate models accurately approximate simulation-based model output with simple algebraic operations, such as polynomial expansions and correlation-based prediction.
Trained surrogate models are used in lieu of computationally expensive simulation models when rapid reactions are required. 
We now introduce the basic theory of a commonly used surrogate modeling approach in ASO: kriging (also known as Gaussian process regression)~\cite{Krige1951,Forrester2007kriging,han2012hierarchical,Bouhlel2016a,Zhou2020}.

\begin{figure}[h]
\centering
\includegraphics[width=\linewidth]{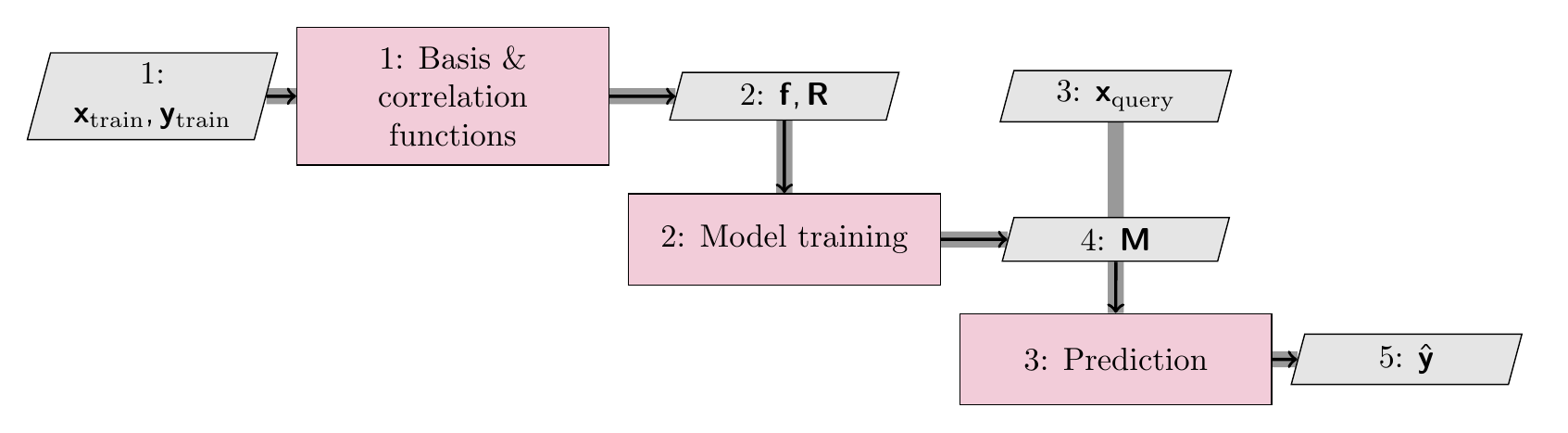}
\caption{Kriging model requires users to select basis ($\mathbf{f}$) and correlation functions ($\mathbf{R}$) for the regression tasks.
The most commonly used model training approach for kriging is the maximum likelihood estimation, which typically assumes Gaussian distributed residuals and selects the optimal hyper-parameters most consistent with the observed data.
}
\label{fig:kriging}
\end{figure}

Kriging models any finite collection of model responses as a multivariate normal distribution distribution~\cite{Beers2004,Oliver1990}.
Thus, kriging treats model observations as points sampled from a Gaussian process (Fig.~\ref{fig:kriging}), which is a key assumption of kriging.
Kriging model can be expressed as
\begin{equation}
    Y^{KR}(\boldsymbol{x}) = \boldsymbol{\beta} \boldsymbol{f}(\boldsymbol{x}) + \sigma^2 Z,
\label{eqn: Kriging}
\end{equation}
where the first term $\boldsymbol{\beta} \boldsymbol{f}(\boldsymbol{x})$ is the trend function (mean of the Gaussian process), $\boldsymbol{f}$ is a basis function vector, $\boldsymbol{\beta}$ are the unknown coefficients to be determined, the $\sigma^2$ is the variance of the Gaussian process, and $Z$ is a zero mean, unit variance, stationary Gaussian process.

\begin{figure}[h]
\centering
\includegraphics[width=0.6\linewidth]{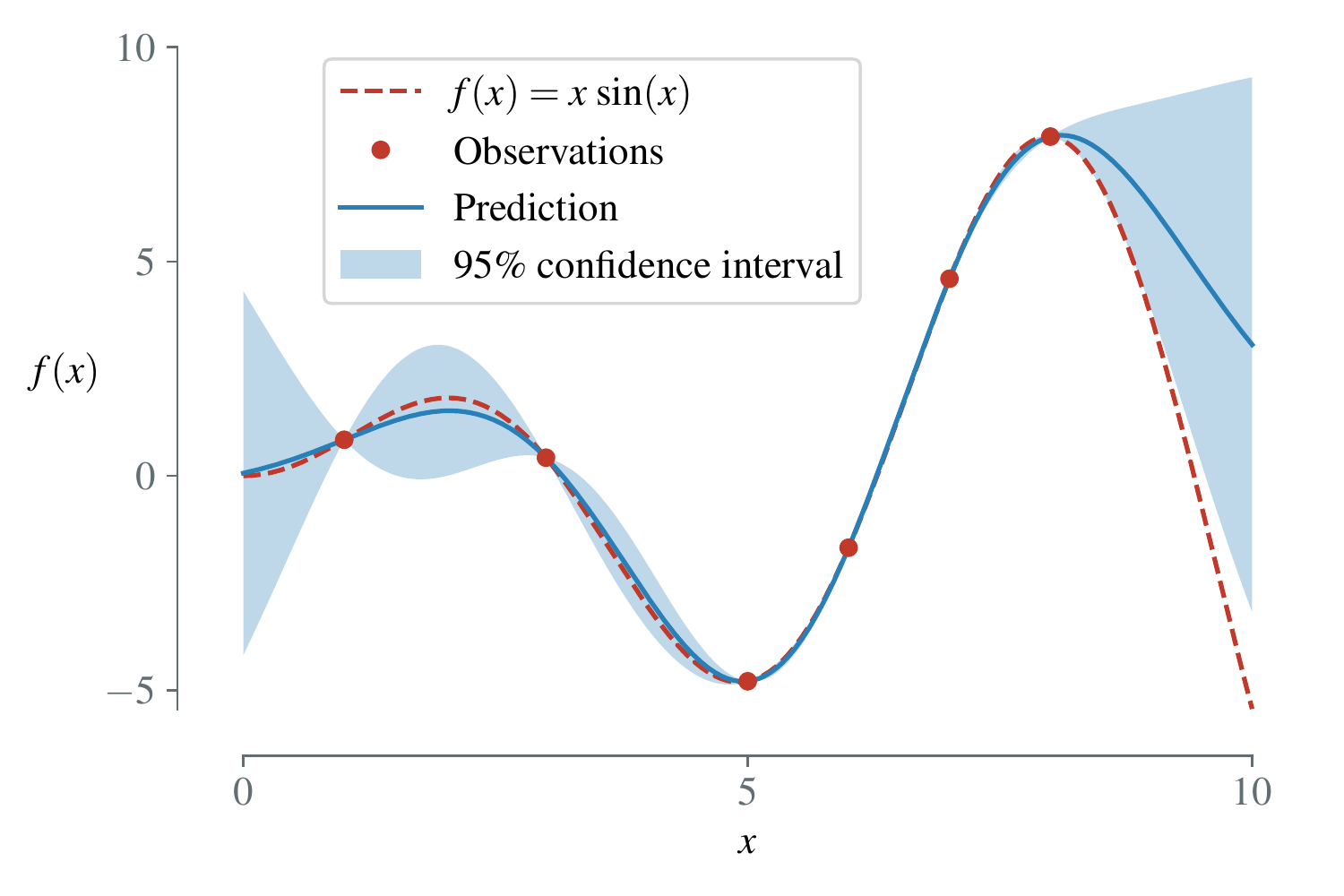}
\caption{Regression example using a kriging surrogate model~\cite{Pedregosa2011}.
True function $f(x) = x \sin(x)$, red dots are training samples and observations, blue curve is the mean predicted values along $x$ axis while shaded region is 95\% predictive confidence interval.
The prediction at training observations has a tight confidence interval, while the unobserved locations have a larger interval.
}
\label{fig:kriging_example}
\end{figure}

Kriging makes predictions based on the Gaussian process assumption that the prediction and training model responses have a joint Gaussian distribution defined by:
\begin{equation}
    \left\{
    \begin{tabular}{@{}l@{}}
        $\hat{y}(x')$ \\
        $\boldsymbol{y}$
    \end{tabular}
    \right\}
    \sim N_{N+1}
    \left(
    \left\{
    \begin{tabular}{@{}l@{}}
        $\boldsymbol{f}^\intercal(\boldsymbol{x})\boldsymbol{\beta}$ \\
        $\boldsymbol{F} \boldsymbol{\beta}$
    \end{tabular}
    \right\},
    \sigma^2
    \left\{
    \begin{tabular}{@{}l@{}}
        1 \ \ \ \ \ $\boldsymbol{r}^\intercal(\boldsymbol{x})$ \\
        $\boldsymbol{r}(\boldsymbol{x})$ \ \ $\boldsymbol{R}$
    \end{tabular}
    \right\}
    \right),
\end{equation}
where $\boldsymbol{x}'$ is the point to be predicted on, $\boldsymbol{F}$ is a matrix of basis functions, $F_{i,j}=f_j(\boldsymbol{x}_i), i = 1, \ldots, N, j = 1, \ldots, P$, $\boldsymbol{r}(\boldsymbol{x})$ is a correlation vector between the $\boldsymbol{x}'$ and training data, $\boldsymbol{R}$ is the correlation matrix of training data with $R_{ij} = R(\boldsymbol{x}_i, \boldsymbol{x}_j; \boldsymbol{\theta})$. Thus, the mean and variance of the prediction at $\boldsymbol{x}'$ are
\begin{equation}
    \mu(\boldsymbol{x}') = \boldsymbol{f}(\boldsymbol{x}')^\intercal\boldsymbol{\beta} + \boldsymbol{r} (\boldsymbol{x})^\intercal \boldsymbol{R}^{-1} (\boldsymbol{y} - \boldsymbol{F \beta}),
\end{equation}
\begin{equation}
    \sigma^2(\boldsymbol{x'}) = \sigma^2 (1 - \boldsymbol{r}^\intercal(\boldsymbol{x}) \boldsymbol{R}^{-1} \boldsymbol{r}(\boldsymbol{x}) + \boldsymbol{u}^\intercal(\boldsymbol{x}) (\boldsymbol{F}^\intercal \boldsymbol{R}^{-1} \boldsymbol{F})^{-1} \boldsymbol{u}(\boldsymbol{x})),
\end{equation}
where
\begin{equation}
    \boldsymbol{\beta} = (\boldsymbol{F}^\intercal \boldsymbol{R}^{-1} \boldsymbol{F})^{-1} \boldsymbol{F}^\intercal \boldsymbol{R}^{-1} \boldsymbol{y},
\end{equation}
\begin{equation}
    \boldsymbol{u}(\boldsymbol{x}) = \boldsymbol{F}^\intercal \boldsymbol{R}^{-1} \boldsymbol{r} (\boldsymbol{x}) - \boldsymbol{f}(\boldsymbol{x}),
\end{equation}
and Gaussian correlation function is defined as
\begin{equation}
    R(\boldsymbol{x}, \boldsymbol{x}^\prime) = \exp \left( - \sum_{i=1}^P \left( \left( \boldsymbol{x}_i - \boldsymbol{x}_i^\prime \right) / \theta_i \right)^2 \right).
\end{equation}
Thus, the maximum likelihood estimation on $\boldsymbol{\theta}$ is solved by
\begin{equation}
    \hat{\boldsymbol{\theta}} = \argmin_{\boldsymbol{\theta}} \left( 1/2 \log(\det(\boldsymbol{R})) + N/2 \log(2\pi\sigma^2) + N/2 \right).
\label{eqn:mle}
\end{equation}
The standard deviation-based kriging predictive confidence interval (Fig.~\ref{fig:kriging_example}) is an important property that enables adaptive sampling strategies and efficient global optimization~\cite{Jones1998a,Bartoli2019b}.

\begin{figure}[h]
\centering
\includegraphics[width=\linewidth]{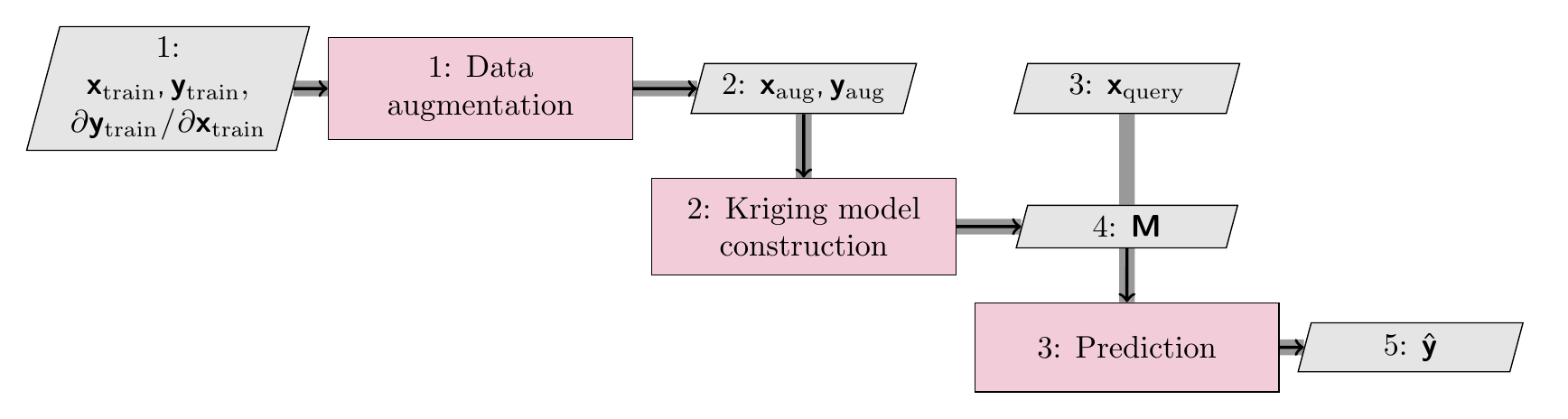}
\caption{Indirect GEK requires data augmentation to get new samples around existing data via the first-order Taylor approximation.
Augmented data ($\mathbf{x}_\text{aug}, \mathbf{y}_\text{aug}$) combining existing and newly generated data are used for Kriging modeling and prediction as introduced in Fig.~\ref{fig:kriging}.
}
\label{fig:indirectGEK}
\end{figure}

\begin{figure}[h]
\centering
\includegraphics[width=\linewidth]{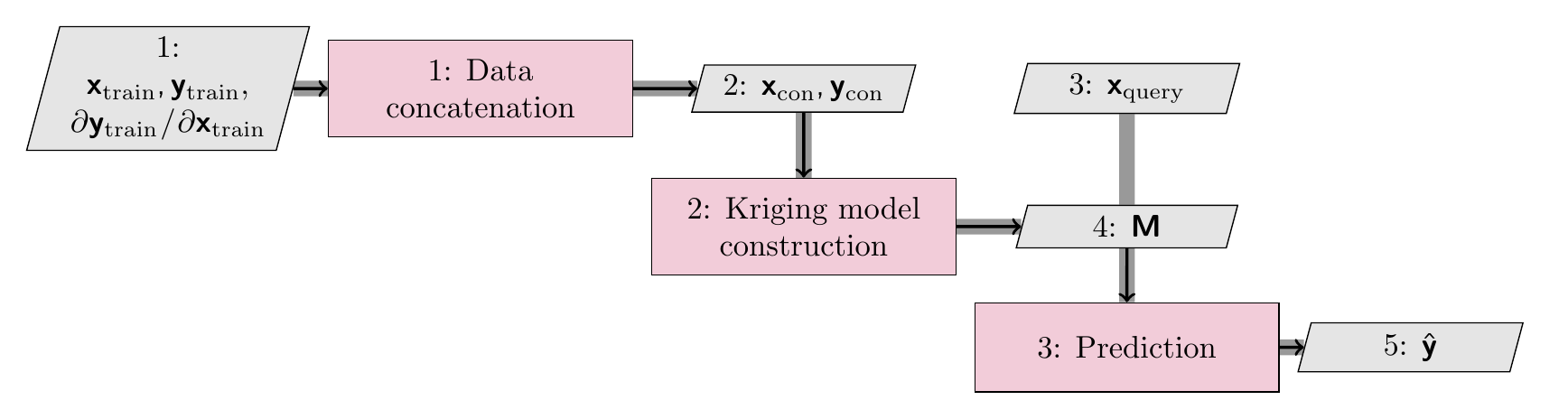}
\caption{Direct GEK directly concatenates existing data and corresponding gradients.
Concatenated data ($\mathbf{x}_\text{con}, \mathbf{y}_\text{con}$) are used for Kriging modeling and prediction as introduced in Fig.~\ref{fig:kriging}.
}
\label{fig:directGEK}
\end{figure}

Gradient-enhanced kriging (GEK) surrogate improves the predictive performance of kriging by incorporating gradient information when available~\cite{Ulaganathan2015,Bouhlel2019a}.
There are two ways of constructing GEK models, indirect GEK (Fig.~\ref{fig:indirectGEK}) and direct GEK (Fig.~\ref{fig:directGEK}).

Indirect GEK generates new points around training data through the first-order Taylor approximation
\begin{equation}
    y(\boldsymbol{x}_i + \Delta x_j \boldsymbol{e}_j) = y(\boldsymbol{x}_i) + \frac{\partial y(\boldsymbol{x}_i)}{\partial x_j} \Delta x_j,
\end{equation}
where $\delta x_j$ is the step added in the $j$th direction, and $\boldsymbol{e}_j$ is the $j$th row of a $P \times P$ identity matrix.
Indirect GEK does not require a modification of a kriging model.
However, the size of the correlation matrix increases rapidly from $N \times N$ to $(N+P) \times (N+P)$, which makes high-dimensional problems challenging for GEK models.
In contrast, direct GEK incorporates derivatives into the model observations to formulate
\begin{equation}
    \boldsymbol{y} = \left[ y(\boldsymbol{x}_1), \cdots, y(\boldsymbol{x}_N), \frac{\partial y(\boldsymbol{x}_1)}{\partial \boldsymbol{x}_1}, \cdots, \frac{\partial y(\boldsymbol{x}_1)}{\partial \boldsymbol{x}_P}, \cdots, \frac{\partial y(\boldsymbol{x}_N)}{\partial \boldsymbol{x}_P} \right]^\intercal.
\end{equation}
Thus, the correlation matrix includes four main blocks, the correlation among the training data, the gradients, the gradients and training data, and between training data and gradients.
The size of the correlation matrix increases quadratically from $N \times N$ to $(N+P) \times (N+P)$.
Therefore, direct GEK has a similar issue as indirect GEK.
\citet{Bouhlel2019a} applied the partial-least squares method to drastically reduce the number of hyperparameters while maintaining high-level accuracy.

\begin{figure}[h]
\centering
\includegraphics[width=\linewidth]{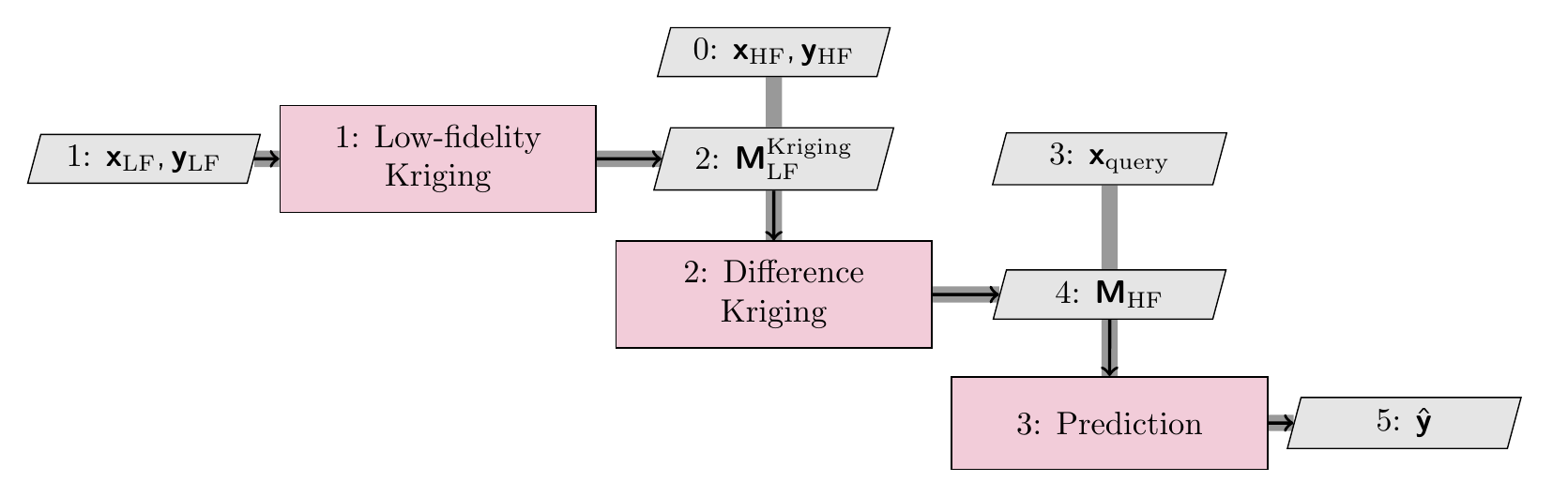}
\caption{Cokriging consists of constructing two kriging models as introduced in Fig.~\ref{fig:kriging}.
The first kriging model ($\mathbf{M}^\text{Kriging}_\text{LF}$) is formulated on LF data ($\mathbf{x}_\text{LF}, \mathbf{y}_\text{LF}$) while the second kriging model is set up on the difference between HF observations ($\mathbf{y}_\text{HF}$) and scaled LF observations (multiply by a parameter $\rho$) at $\mathbf{x}_\text{HF}$.
If $\mathbf{y}_\text{LF}$ is not available at $\mathbf{x}_\text{HF}$, $\mathbf{M}^\text{Kriging}_\text{LF}$ can be used for quick estimations.
}
\label{fig:coKriging}
\end{figure}

Another important branch of kriging is cokriging, which is a multivariate kriging and capable of fusing information from models of different accuracy fidelity levels~\cite{Forrester2007kriging}.
A cokriging model consists of  {a low-fidelity (LF) model-based kriging surrogate multiplied by a scaling factor and another kriging surrogate modeling the difference between the high-fidelity (HF) and LF models as shown in Fig.~\ref{fig:coKriging}, and can be written as}
\begin{equation}
    Y^{CK}(\boldsymbol{x}) = \rho Y^{KR}_\text{LF}(\boldsymbol{x}) + Y^{KR}_\text{Diff}(\boldsymbol{x}),
\label{eqn: Cokriging}
\end{equation}
where $\rho$ is the scaling factor to be determined via maximum likelihood optimization, $Y^{KR}_\text{LF}$ is the kriging surrogate of LF model, and $Y^{KR}_\text{Diff}$ is the kriging surrogate corresponding to the difference.
The cokriging predictor has a generalized format of kriging by incorporating multi-fidelity information,
\begin{equation}
    \mu(\boldsymbol{x}') = \boldsymbol{f}_\text{CK}(\boldsymbol{x}')^\intercal\boldsymbol{\beta} + \boldsymbol{r}_\text{CK} (\boldsymbol{x})^\intercal \boldsymbol{R}_\text{CK}^{-1} (\boldsymbol{y}_\text{CK} - \boldsymbol{F}_\text{CK} \boldsymbol{\beta}_\text{CK}),
\end{equation}
where the trend function $\boldsymbol{f}_\text{CK}^\intercal = [\rho \boldsymbol{f}_\text{LF}^\intercal, \boldsymbol{f}_\text{HF}^\intercal]$, and
\begin{equation}
    \boldsymbol{F}_\text{CK} =
    \begin{bmatrix}
    \boldsymbol{f}_\text{LF}(\boldsymbol{x}_1)^\intercal & \cdots & \boldsymbol{f}_\text{LF}(\boldsymbol{x}_{N_\text{LF}})^\intercal &
    \rho \boldsymbol{f}_\text{HF}(\boldsymbol{x}_1)^\intercal & \cdots & \rho \boldsymbol{f}_\text{HF}(\boldsymbol{x}_{N_\text{HF}})^\intercal \\
    0 & 0 & 0 &
    \boldsymbol{f}_\text{HF}(\boldsymbol{x}_1)^\intercal & \cdots & \boldsymbol{f}_\text{HF}(\boldsymbol{x}_{N_\text{HF}})^\intercal
    \end{bmatrix}^\intercal.
\end{equation}
The correlation between the untried sample point ($\boldsymbol{x}'$) and training data is
\begin{equation}
    \boldsymbol{r}^\intercal = \bigg[\rho \sigma^2_\text{LF} R_\text{LF}\big( \boldsymbol{x}', \boldsymbol{X}_\text{LF}; \boldsymbol{\theta}_\text{CK} \big), \rho^2 \sigma^2_\text{LF} R_\text{LF}\big( \boldsymbol{x}', \boldsymbol{X}_\text{HF}; \boldsymbol{\theta}_\text{CK} \big) + \sigma^2_\text{HF} R_\text{HF}\big( \boldsymbol{x}', \boldsymbol{X}_\text{HF}; \boldsymbol{\theta}_\text{CK} \big) \bigg],
\end{equation}
and the covariance matrix among HF and LF training data is
\begin{equation}
    \boldsymbol{R}_\text{CK} =
    \begin{bmatrix}
    \sigma^2_\text{LF} R_\text{LF}\big( \boldsymbol{X}_\text{LF}, \boldsymbol{X}_\text{LF}; \boldsymbol{\theta}_\text{CK} \big) &
    \rho \sigma^2_\text{LF} R_\text{LF}\big( \boldsymbol{X}_\text{LF}, \boldsymbol{X}_\text{HF}; \boldsymbol{\theta}_\text{CK} \big) \\
    \rho \sigma^2_\text{LF} R_\text{LF}\big( \boldsymbol{X}_\text{HF}, \boldsymbol{X}_\text{LF}; \boldsymbol{\theta}_\text{CK} \big) &
    \rho^2 \sigma^2_\text{LF} R_\text{LF}\big( \boldsymbol{X}_\text{HF}, \boldsymbol{X}_\text{HF}; \boldsymbol{\theta}_\text{CK} \big) + \sigma^2_\text{HF} R_\text{HF}\big( \boldsymbol{X}_\text{HF}, \boldsymbol{X}_\text{HF}; \boldsymbol{\theta}_\text{CK} \big)
    \end{bmatrix},
\end{equation}
and $\boldsymbol{\theta}$ is estimated via the maximum likelihood method (Eq.~\eqref{eqn:mle}).

There are other surrogate modeling methods that have been successfully introduced to engineering areas.
For example, the orthogonal bases-based polynomial chaos expansion (PCE)~\cite{Wiener1938,Xiu2010} provides analytical mean and standard deviation of model observations, as well as the Sobol' indices for sensitivity analysis~\cite{Sudret2008,Crestaux2009}.
 {The generalized PCE can be written as}
\begin{equation}
\label{eqn: pce}
Y(\boldsymbol{x}) = \sum_{i=1}^\infty \alpha_i \Phi_i(\boldsymbol{x}),
\end{equation}
 {where $\boldsymbol{x} \in \mathbb{R}^{n}$ is an $m$-dimensional random input vector, $Y(\boldsymbol{X})$ is a computational model of $\boldsymbol{X}$, $i$ is the index of $i\text{th}$ polynomial term, $\Phi_i$ is multivariate polynomial basis, and $\alpha_i$ is the corresponding basis function coefficient.
In practice, the following truncated-form PCE is used:}
\begin{equation}
\label{eqn: truncated_pce}
Y(\boldsymbol{x}) \approx Y^{PC}(\boldsymbol{x}) = \sum_{i=1}^P \alpha_i \Phi_i(\boldsymbol{x}),
\end{equation}
 {where $Y^{PC}(\boldsymbol{x})$ is the approximate truncated PCE model, and the total reference number of required sample points follows the factorial formula of ${(p+m)!}/{p!m!}$,
where $p$ is the required order of the PCE. 
\citet{Blatman2009} proposed to use the hyperbolic truncation technique to reduce the interaction terms following the sparsity-of-effect principle~\cite{Baker2016}.} 
 {Thus, the summation of the truncated PCE predictions and associated residuals to match high-fidelity observations at training points ($\boldsymbol{X}$) is as follows:}
\begin{equation}
\label{eqn: pce_residual}
Y(\boldsymbol{X}) \approx Y^{PC}(\boldsymbol{X}) {+ \epsilon_{PC}} \equiv \boldsymbol{\alpha}^\intercal\boldsymbol{\Phi}(\boldsymbol{X}) + \epsilon_{PC},
\end{equation}
 {where $\epsilon_{PC}$ is the residual between $Y(\boldsymbol{X})$ and $Y^{PC}(\boldsymbol{X})$. 
This residual can be minimized using the least-squares method as}
\begin{equation}
\label{eqn: least_squares}
\hat{\boldsymbol{\alpha}} = \text{arg }\underset{\boldsymbol{\alpha}}{\text{min}} \  {\E}[\boldsymbol{\alpha}^\intercal\boldsymbol{\Phi}(\boldsymbol{X}) - Y(\boldsymbol{X})].
\end{equation}
 {One common technique is to add an $L_1$ regularization term to favor low-rank solutions as follows~\cite{Udell2016}:}
\begin{equation}
\label{eqn: penalty}
\hat{\boldsymbol{\alpha}} = \text{arg }\underset{\boldsymbol{\alpha}}{\text{min}} \  {\E}[\boldsymbol{\alpha}^\intercal\boldsymbol{\Phi}(\boldsymbol{X}) - Y(\boldsymbol{X})] + \lambda ||\boldsymbol{\alpha}||_1,
\end{equation}
 {where $\lambda$ is a penalty factor, and $||\boldsymbol{\alpha}||_1$ is the $L_1$ norm of the coefficients of the PCE.}
 {\citet{Sudret2008} introduced least angle regression (LARS) to determine PCE coefficients and further reduce the basis terms via early stop.}

\begin{figure}[h]
\centering
\includegraphics[width=\linewidth]{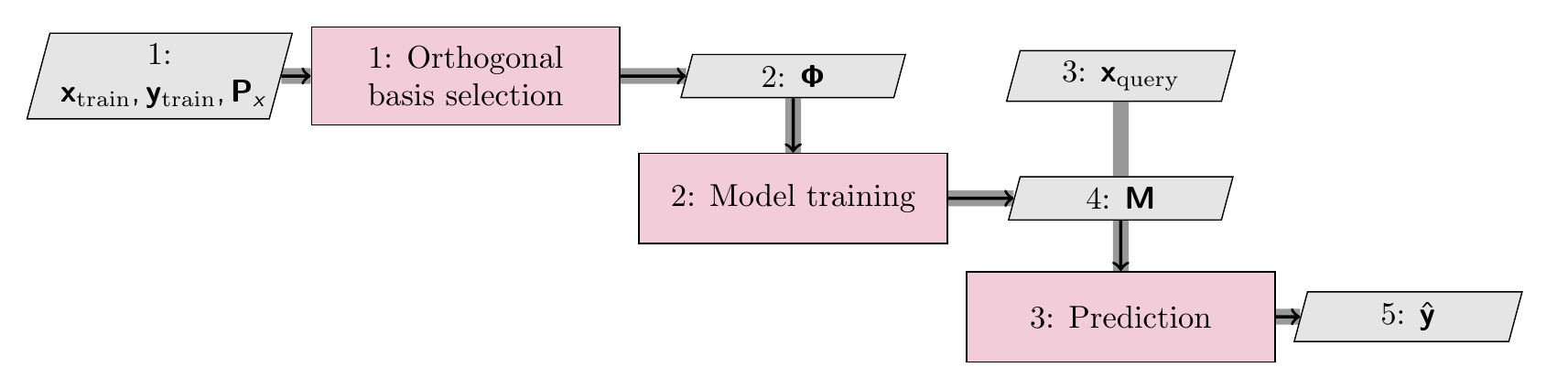}
\caption{PCE requires the users to provide a pre-defined or assumed probabilistic distribution ($\mathbf{P}_x$) on the input parameters.
PCE sets up orthogonal bases corresponding with the probabilistic distributions and sparsely selects lower-order bases ($\mathbf{\Phi}$) using hyperbolic basis truncation (also known as $q$-norm)~\cite{Sudret2008}.
Ordinary least squares (OLS) is the most commonly used training algorithm.
LARS scheme leads to state-of-the-art PCE model training as introduced by \citet{Sudret2008} because LARS further reduces the basis terms via early stop.
Eventually, we can use PCE model ($\mathbf{M}$) for prediction ($\hat{\boldsymbol{y}}$) on query data ($\boldsymbol{x}_\text{query}$).
}
\label{fig:PCE}
\end{figure}

\begin{figure}[h]
\centering
\includegraphics[width=\linewidth]{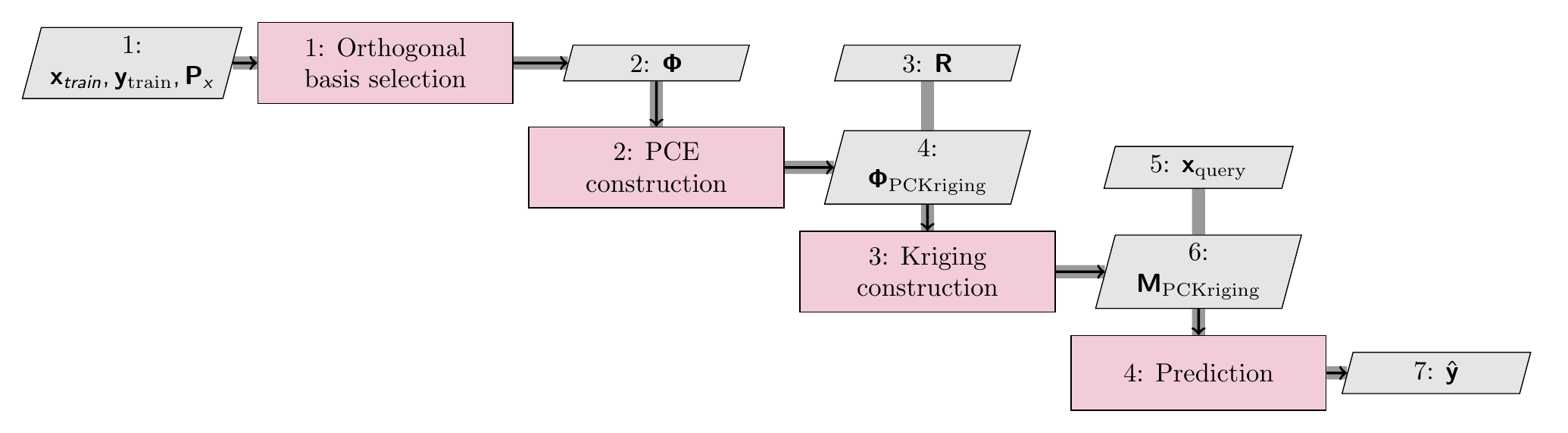}
\caption{PC-Kriging modeling combines the construction process of PCE and kriging.
We first establish a PCE model as shown in Fig.~\ref{fig:PCE}.
Then we only use the selected bases ($\mathbf{\Phi}_\text{PCKriging}$) by hyperbolic truncation and LARS as the basis function of kriging.
The rest follows the same process as training a kriging model (Fig.~\ref{fig:kriging}) by setting the correlation function ($\mathbf{R}$).
}
\label{fig:PCKriging}
\end{figure}

PCE is a regression-based method, and the orthogonal bases are selected corresponding to the probability distribution of input parameters (Fig.~\ref{fig:PCE}) so that PCE fits well in analysis and design under uncertainty~\cite{Du2019a}.
 {On the other hand, kriging models are interpolation-based surrogates that guarantee predictions to exactly match the observations at the training samples.}
Researchers developed the PCE-based kriging (PC-Kriging)~\cite{Schoebi2015} and PCE-based Cokriging (PC-Cokriging)~\cite{Du2020pcco} models to combine  {the regression-based} PCE with  {the interpolation-based} kriging and cokriging models, respectively.
In particular, PC-Kriging uses PCE as a trend function to capture the general shape and kriging approximation to interpolate through training data (Fig.~\ref{fig:PCKriging}).
 {Mathematically, PC-Kriging extends original kriging} (Eq.~\eqref{eqn: Kriging}) as
\begin{equation}
    Y^{PCK}(\boldsymbol{X}) = \boldsymbol{\alpha}^\intercal \boldsymbol{\Phi}(\boldsymbol{X}) + \sigma^2Z(\boldsymbol{X}),
\end{equation}
 {where $Y^{PCK}$ is the approximation using PC-Kriging.
The first right-hand-side term is the truncated-form PCE, which is used as the trend function within the universal Kriging formula, and in the second term $\sigma$ and $Z(\boldsymbol{X})$ denote the constant standard deviation and the zero mean and unit variance stationary Gaussian process (Eq.~\eqref{eqn: Kriging}).}

\begin{figure}[h]
\centering
\includegraphics[width=\linewidth]{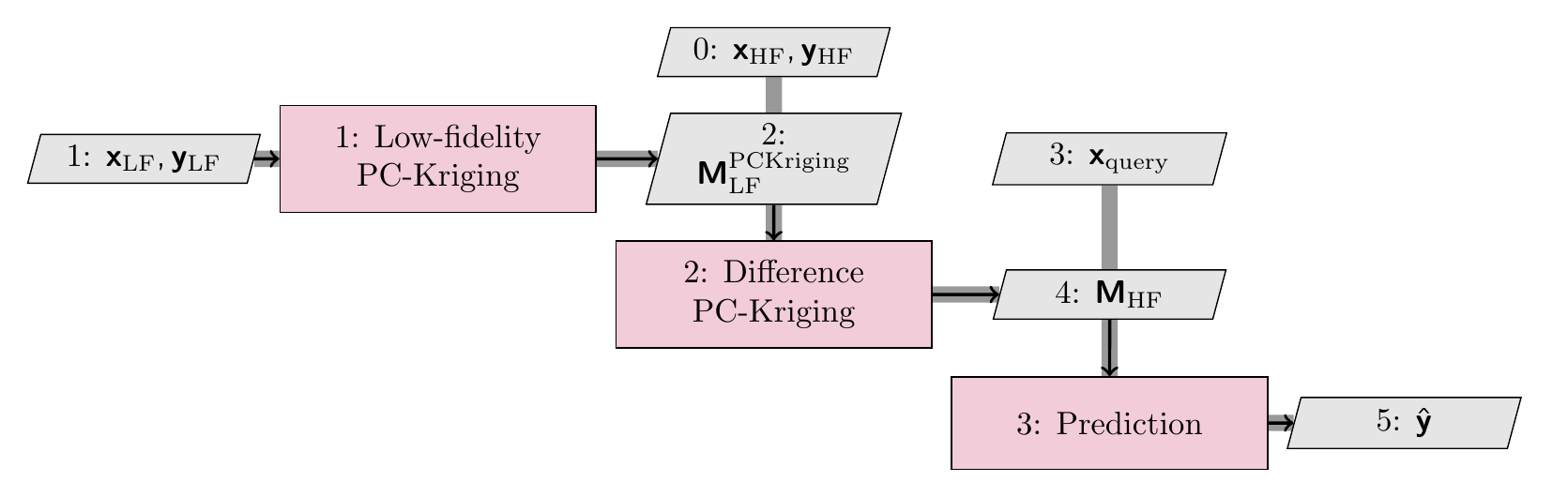}
\caption{PC-Cokriging is similar to cokriging (Fig.~\ref{fig:coKriging}) except that we need to formulate low-fidelity PC-Kriging and difference PC-Kriging models (Fig.~\ref{fig:PCKriging}) rather than kriging models. }
\label{fig:PCCoKriging}
\end{figure}

\begin{figure}[h]
\centering
\includegraphics[width=0.6\linewidth]{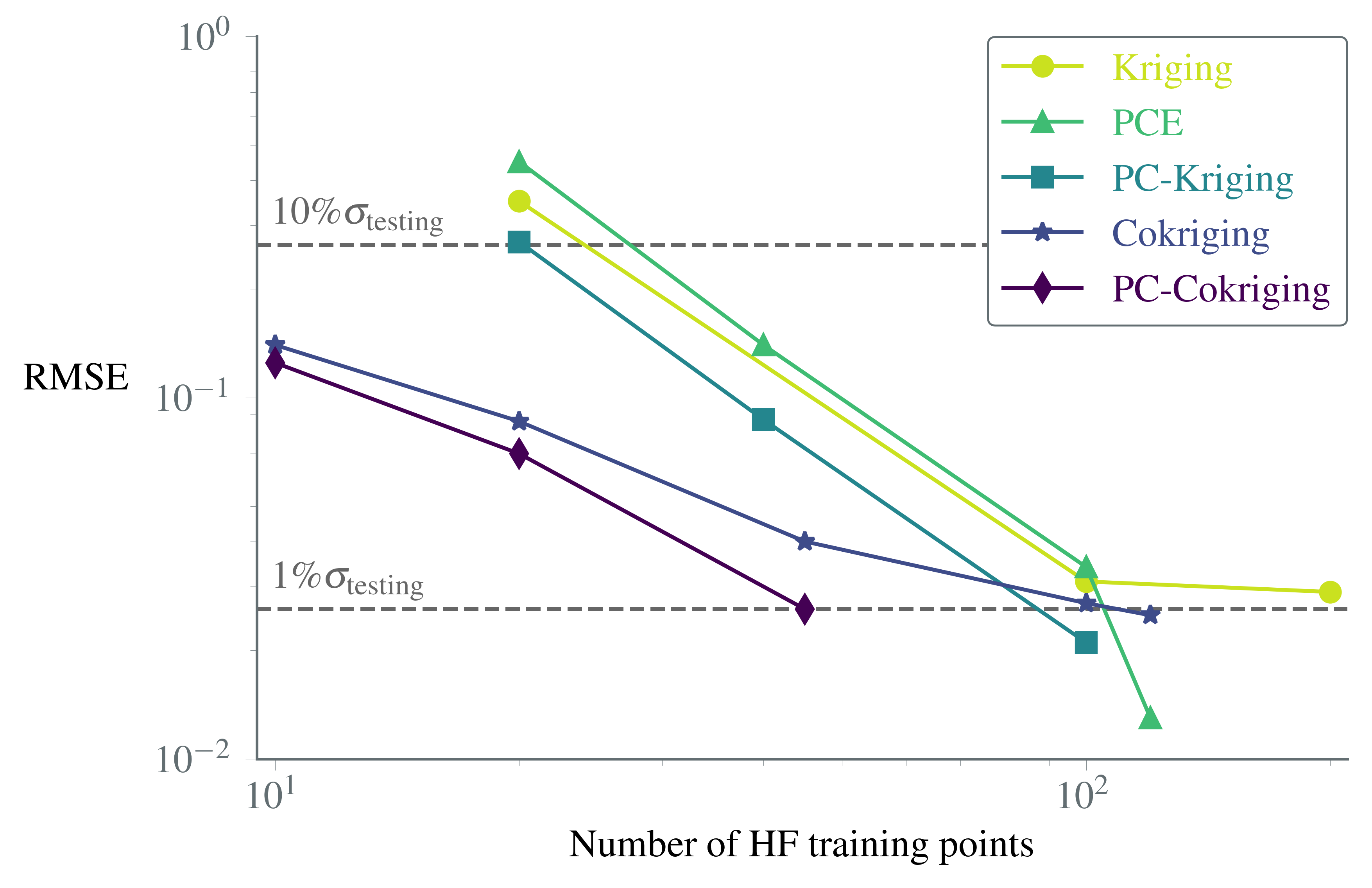}
\caption{\citet{Du2020pcco} compared a range of kriging and PCE related algorithms on Currin function~\cite{Currin1988ABA} and its low-fidelity version~\cite{Xiong2013LF}.
Results show that to reduce the root mean squared error (RMSE) to 1\% of the standard deviation of the testing labels, PC-Cokriging required around 50 samples while PC-Kriging and LARS-based PCE models required around 100 samples.
Cokriging took more than 100 samples meaning that the low-fidelity model was not highly correlated with the Currin function, but PC-Cokriging was still capable of achieving promising predictive performance.
}
\label{fig:RMSE_Currin}
\end{figure}

PC-Cokriging generalizes PC-Kriging to the multi-fidelity modeling architecture (Fig.~\ref{fig:PCCoKriging}) following the same principle as cokriging extending kriging (Eq.~\eqref{eqn: Cokriging}).
 {Mathematically, PC-Cokriging has the generalized formula}
\begin{equation}
\label{eqn: pccokriging}
Y^{PCCK}(\boldsymbol{X}) = \rho Y^{PCK}_{\text{LF}}(\boldsymbol{X}) + Y^{PCK}_{\text{Diff}}(\boldsymbol{X}),
\end{equation}
 {where $Y^{PCK}_{\text{LF}}$ is the PC-Kriging approximation of the low-fidelity model data, and $Y^{PCK}_{\text{Diff}}$ is the PC-Kriging approximation of the difference of the output of the high- and low-fidelity model data.
The $Y^{PCK}_{\text{LF}}$ and} $Y^{PCK}_{\text{Diff}}$ terms are constructed using the PC-Kriging method.
\citet{Du2020pcco} developed PC-Cokriging model and demonstrated the outstanding performance on a series of analytical examples (Fig.~\ref{fig:RMSE_Currin}) and engineering cases.
Moreover, other multi-fidelity surrogate methods~\cite{Peherstorfer2018a,eldred,feldstein2020}, such as manifold mapping, managed to accurately predict scalar and vector quantities for multilevel aerodynamic design optimizations~\cite{Hemker2007,DU2019Inverse}.

\subsection{Unsupervised Learning}
\label{sec:mlUL}


In contrast to supervised learning, which reads labeled training data, unsupervised learning works with unlabeled training data.
Unsupervised learning algorithms analyze the data pattern and automatically set up learning rules~\cite{Ghahramani2004,Usama2019}.
Unsupervised learning has prevailed in various important problems, such as anomaly detection~\cite{Thudumu2020comprehensive,Pang2021} and novelty detection~\cite{Pimentel2014,Berg2019unsupervised}, clustering~\cite{Min2018,Rodriguez2019}, and dimensionality reduction\cite{Reddy2020,Howley2006}.
We focus on clustering and dimensionality reduction in the rest of this section because they have more applications in ASO.

Clustering is the task of  {automatically} separating  {unlabeled} data into different groups.
Data in each group share similar characteristics and has highly dissimilar characteristics compared with data in different groups~\cite{Min2018,Rodriguez2019}.
An important real-world clustering application is fake news identification~\cite{Parikh2018,Yazdi2020,Oliveira2021}.
Clustering approaches have also achieved success to detect spam emails that annoy individual users and waste network bandwidth~\cite{Basavaraju2010,Shah2018,Martino2020}.
Other practical clustering applications include market and customer segmentation, which splits the target market into smaller categories and segments customers into groups of similar characteristics~\cite{Kashwan2013,Kansal2018,Janardhanan2020}.
Clustering techniques  {can help} divide the design space in ASO problems.
Such design-space division enables the construction of separate supervised regression models, each of which handles a subgroup of similar-feature data.
We will elaborate on the $k$-means method and Gaussian mixture model (GMM) in this section.

Dimensionality reduction refers to the technique of transforming high-dimensional data into low-dimensional  {representation} while retaining the original data's key properties~\cite{Reddy2020}.
For regression problems containing highly correlated high-dimensional inputs, dimensionality reduction can alleviate the ``curse of dimensionality'' issue by reducing the input dimension.
Both linear reduction techniques and nonlinear reduction approaches are popular in ASO and we introduce both types in the rest of this section.

\subsubsection{$k$-means}
\label{sec:mlKM}

\begin{figure}[h]
\centering
\includegraphics[width=\linewidth]{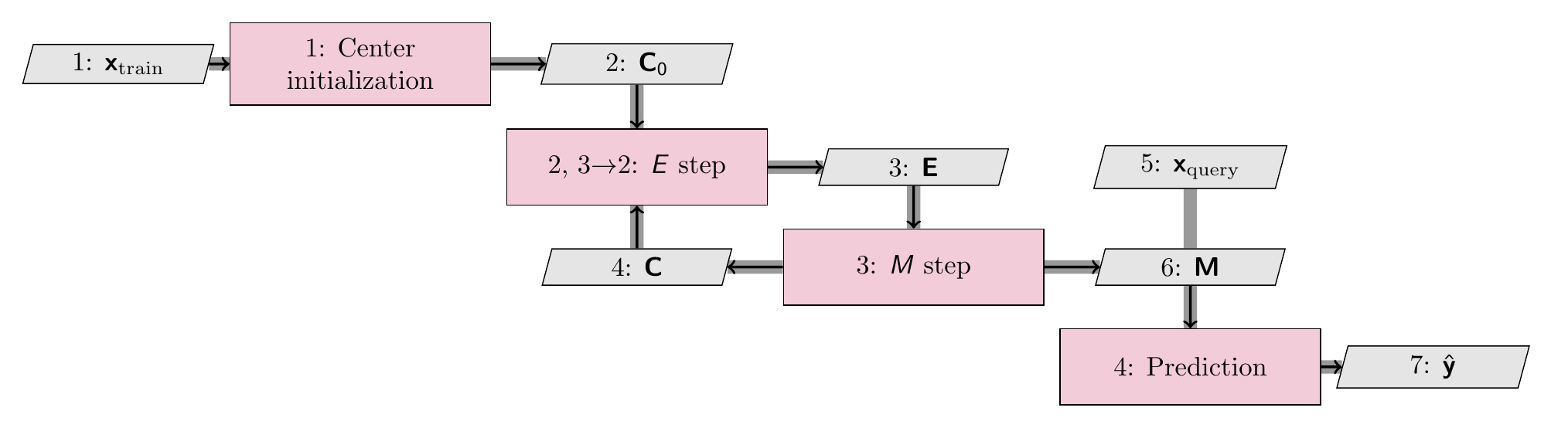}
\caption{$k$-means starts with initial guess on cluster centers ($\mathbf{C}_0$) then loops between $E$ and $M$ steps to keep assigning data, averaging data locations ($\mathbf{E}$) and updating cluster centers ($\mathbf{C}$) until there is no change on cluster centers.
}
\label{fig:KM}
\end{figure}

The $k$-means algorithm searches for a user-defined number of clusters within a multidimensional unlabeled data set~\cite{Arthur2007,Na2010}.
The clustering process follows two assumptions: the ``cluster center'' is the arithmetic mean of all the data within the cluster; each data sample is closer to its cluster center.
The cluster centers greatly affect the clustering results, so they need to be well-placed.
One way of achieving ``optimal'' locations of cluster centers is through the expectation-maximization (EM) steps (Fig.~\ref{fig:KM}).
The $E$ step is so named because this step involves calculating the expectation  {of which cluster each data point belongs to.}
 {The $i\text{th}$ cluster, can be expressed as}
\begin{equation}
    C_i = \{ x_q: ||x_q - c_i||_2 \leq ||x_q - c_j||_2 \ \forall j, 1 \leq j \leq k \},
\end{equation}
 {where $x_q$ is the query data point, and $c_i$ is the cluster center.}
The $M$ step is associated with maximizing some fitness function that defines the cluster center locations.
In the simplest case, maximizing the fitness function is accomplished by taking the mean of the data in each cluster. 
``Convergence'' means there is no change in cluster centers.

\begin{figure}[h]
\begin{center}
    \subfigure[] {\includegraphics[   width=.49\textwidth]{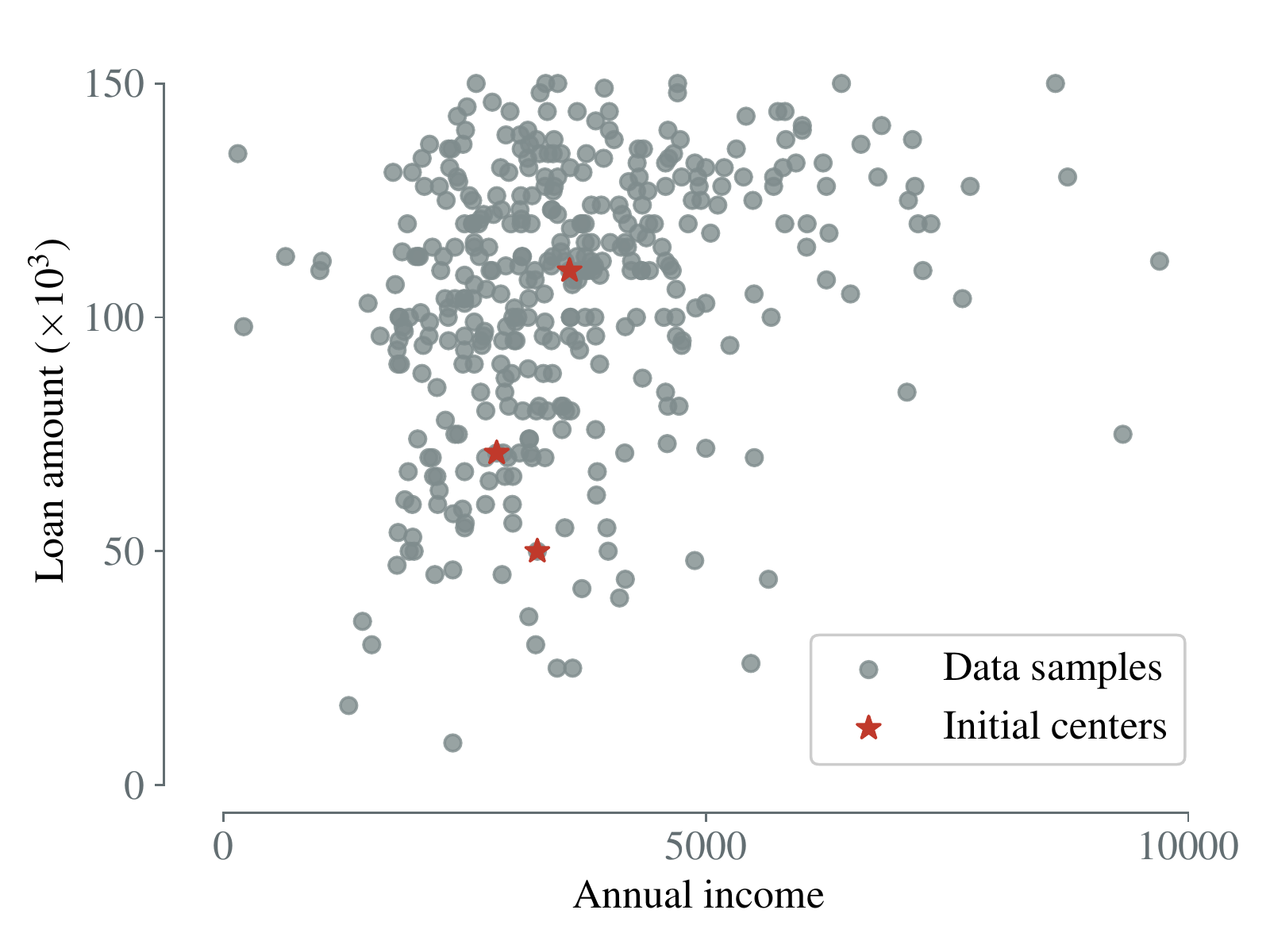}\label{}}
    \hspace{0.3mm}
    \subfigure[] {\includegraphics[
    width=.49\textwidth]{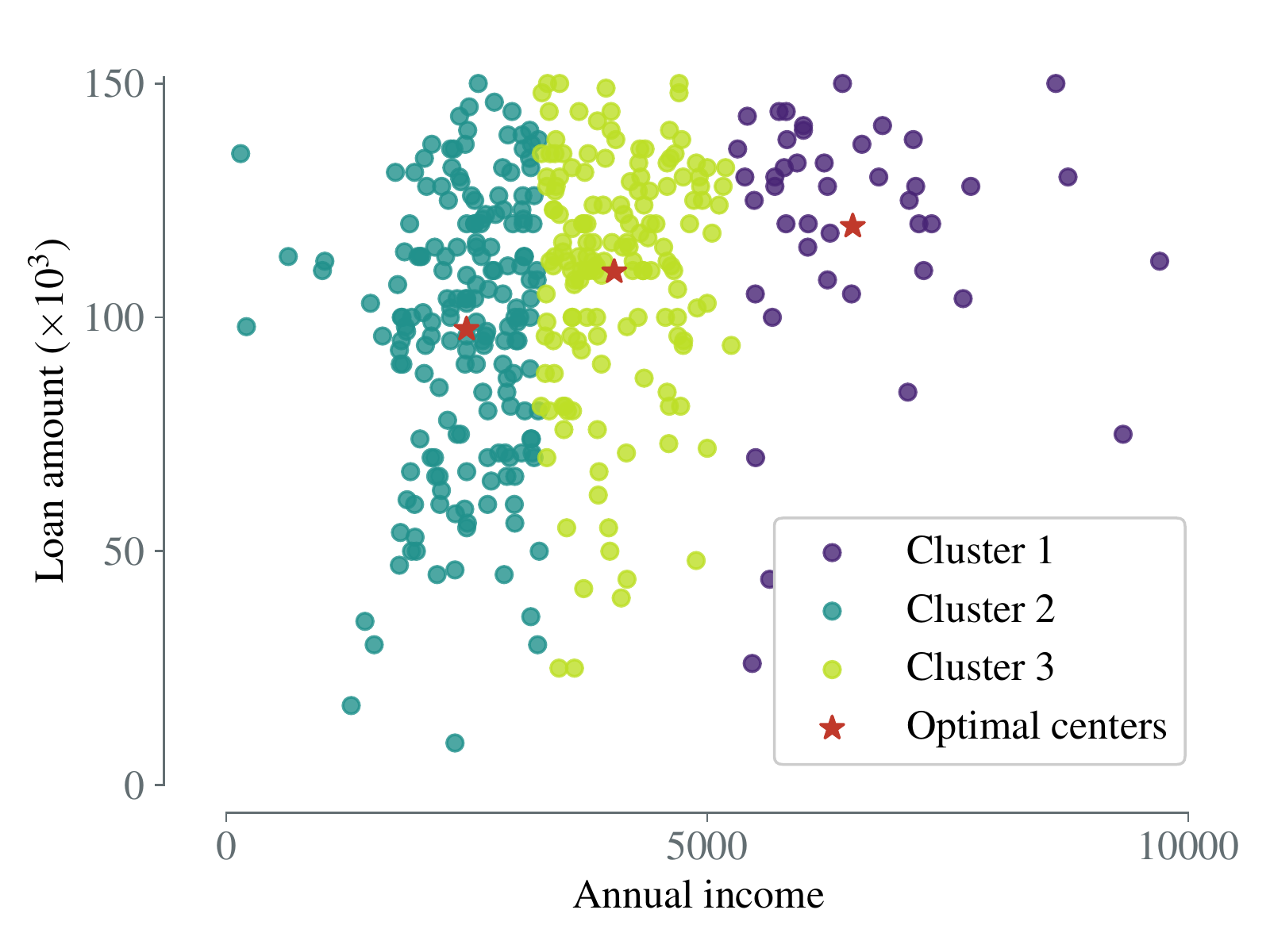}\label{}}
\end{center}
\vspace{-0.5cm}
\caption{An example of wholesale customers data set reveals that $k$-means method is robust even provided with initial centers that are close to each other:
(a) random initial guess on cluster centers and data samples;
(b) optimal cluster centers and clustered data.
In the meantime, we can see $k$-means has linear cluster boundaries.}
\label{fig:KM_example}
\end{figure}

$k$-means algorithm is fast, robust, and straightforward to understand.
It works well if data samples are distinct and well-separated from each other (Fig.~\ref{fig:KM_example}).
However, the globally optimal results may not be achieved even using the EM approach.
In addition, the cluster number needs to be selected beforehand, leading to a difficult decision~\cite{Hamerly2003}.
Moreover, the $k$-means algorithm is limited to linear cluster boundaries, while most clustering problems have complex-geometry boundaries.

$k$-means algorithm is seldom used alone in aerodynamic applications.
Instead, $k$-means method typically addresses clustering preprocessing to raise the performance of other supervised learning approaches.
\citet{Sanwale2018} completed aerodynamic parameter estimation involving surrogate modeling on aerodynamic force and moment coefficients.
Specifically, they applied a radial basis function (RBF) neural network model using $k$-means clustering algorithm to find the centers of RBF, and achieved success on the real-time aircraft flight data of the Advanced Technologies Testing Aircraft System project.

\subsubsection{Gaussian Mixture Model}
\label{sec:mlGMM}
\begin{figure}[h]
\centering
\includegraphics[width=\linewidth]{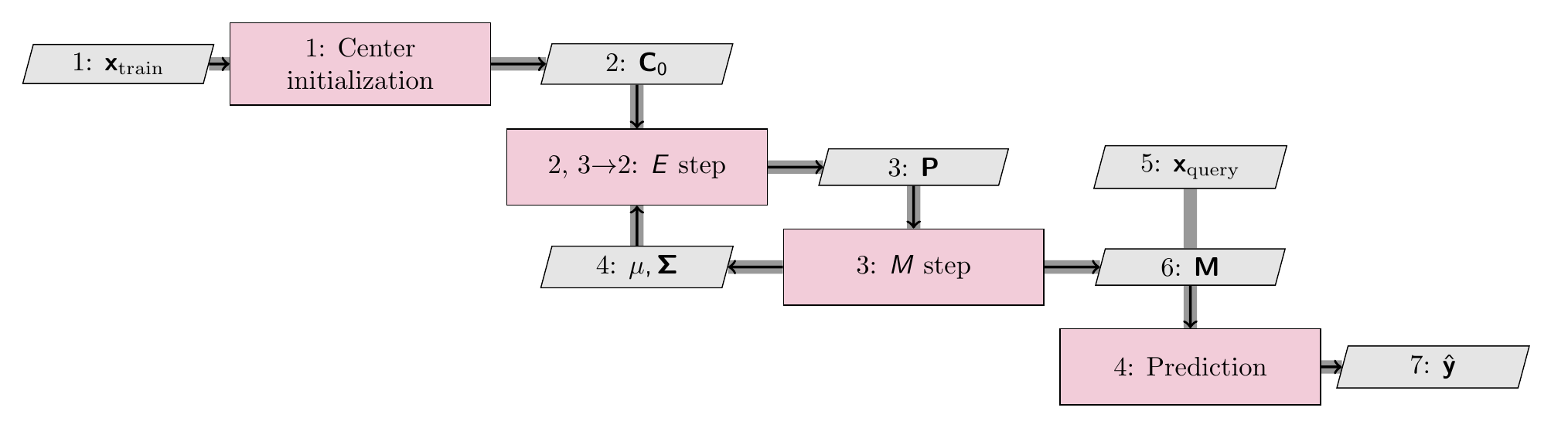}
\caption{GMM starts with initial guess on cluster centers ($\mathbf{C}_0$) then loops between $E$ and $M$ steps.
Specifically, $E$ steps calculates the probabilities ($\mathbf{P}$) of data samples belonging to each cluster while $M$ step updates the statistics (mean $\mathbf{\mu}$ and covariance $\mathbf{\Sigma}$) of each cluster.
}
\label{fig:GMM}
\end{figure}

GMM assumes that a mixture of multiple Gaussian distributions with unknown parameters generates all the data points, where each Gaussian distribution is associated with one cluster~\cite{Reynolds2009,Mohamed2009}.
GMM generalizes $k$-means clustering by considering not only mean values but also the covariance structure of data features.
Given a data set with $d$ features, GMM has $k$ multivariate Gaussian distributions where $k$ is the number of clusters and each distribution has a certain mean and covariance matrix:
\begin{equation}
    f(\boldsymbol{x} | \boldsymbol{\mu}, \Upsigma) = \frac{1}{\sqrt{2 \pi |\Upsigma|}} \exp \big[ - \frac{1}{2} (\boldsymbol{x}-\boldsymbol{\mu})^\intercal \Upsigma^{-1}(\boldsymbol{x}-\boldsymbol{\mu}) \big],
\end{equation}
where $\boldsymbol{\mu}$ is the mean vector of length $d$ and $\Upsigma$ is a $d \times d$ covariance matrix. GMM also uses the EM algorithm~\cite{Reynolds2009, Naim2012convergence} to solve for $\boldsymbol{\mu}$ and $\Upsigma$ (Fig.~\ref{fig:GMM}):
\begin{enumerate}
    \item Define $k$ centers, one for each cluster. The better choice is to place them far away from each other as much as possible.

    \item EM iterations until convergence.
    \begin{enumerate}
        \item $E$ step: calculate the probability of $x_i$ belonging to the cluster $c_1$, \ldots, $c_k$ by:
        \begin{equation}
            p_{i,c} = \frac{q_c N(x_i; \boldsymbol{\mu}_c, \Upsigma_c)}{\sum_{c' \in C} q_{c'}N(x_i; \boldsymbol{\mu}_{c'}, \Upsigma_{c'})} ,
        \end{equation}
        which is the probability of $x_i$ belonging to $c$ divided by the sum of probability $x_i$ belonging to $c_1$, \ldots, $c_k$. The probability is high if the point is assigned to the right cluster and low otherwise.

        \item $M$ step: update the $q$, $\boldsymbol{\mu}$ and $\Upsigma$ in the following manner:
        \begin{equation}
            q = \frac{N_c}{N},
        \end{equation}
        \begin{equation}
            \boldsymbol{\mu}_c = \frac{1}{N_c} \sum_{i=1}^N p_{i,c}\boldsymbol{x}_i,
        \end{equation}
        \begin{equation}
            \Upsigma_c = \frac{1}{N_c} \sum_{i=1}^N p_{i,c} (\boldsymbol{x}_i - \boldsymbol{\mu}_c)^\intercal(\boldsymbol{x}_i - \boldsymbol{\mu}_c),
        \end{equation}
        where $N_c$ is the number of data samples assigned to the cluster, $N$ is the total number of data samples.
    \end{enumerate}
\end{enumerate}

\begin{figure}[h]
\begin{center}
    \subfigure[] {\includegraphics[   width=.49\textwidth]{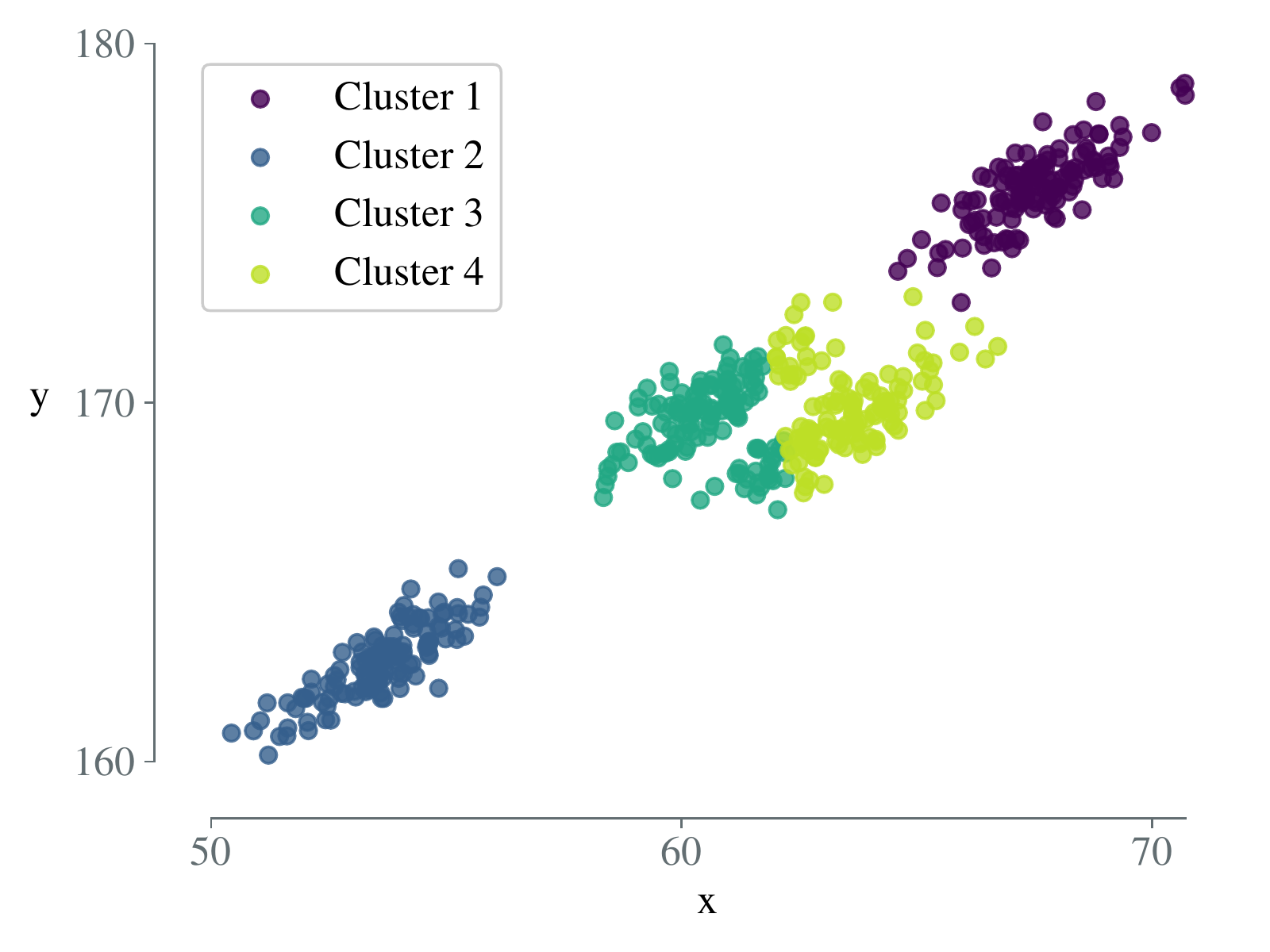}\label{}}
    \hspace{0.3mm}
    \subfigure[] {\includegraphics[
    width=.49\textwidth]{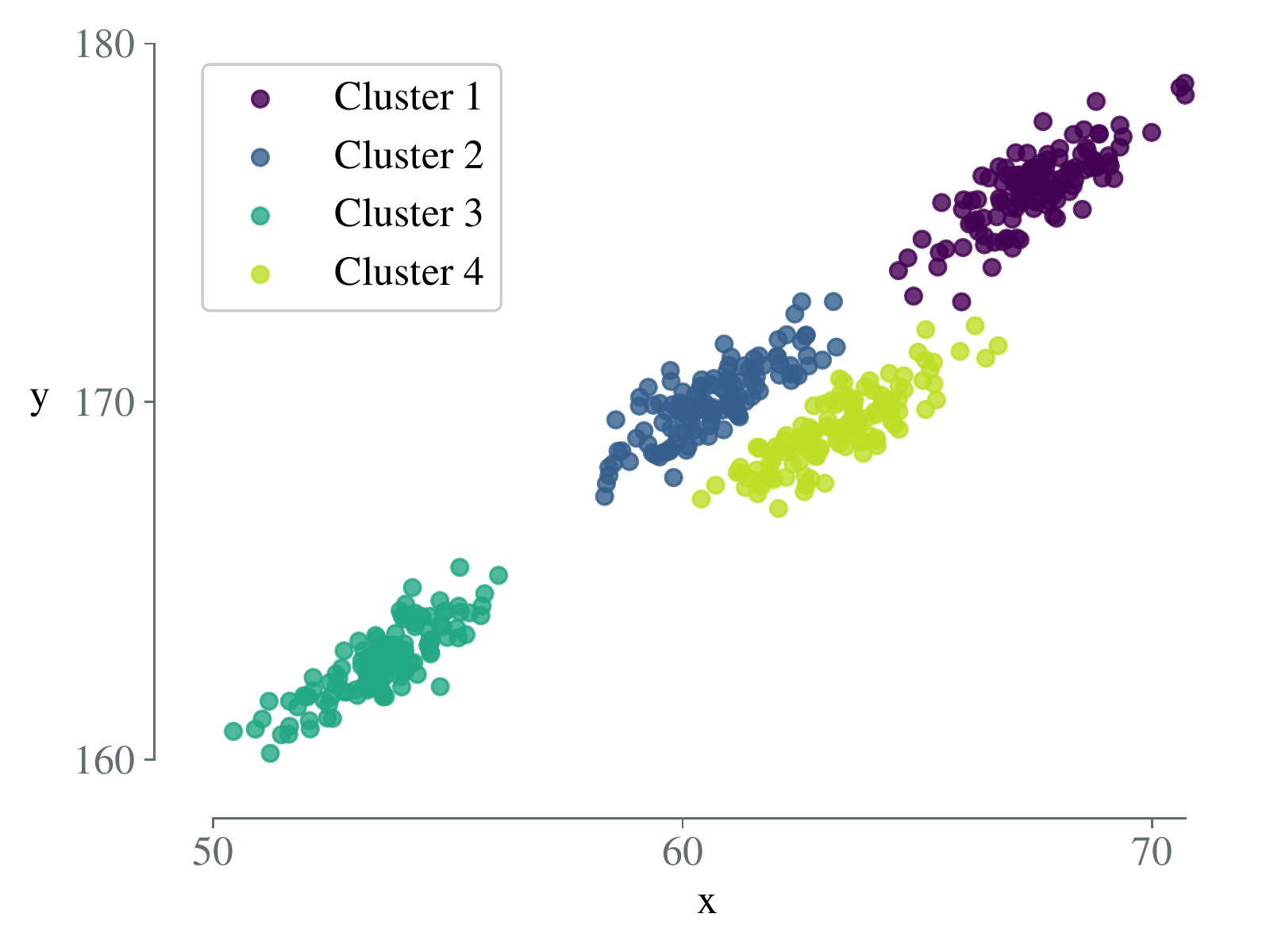}\label{}}
\end{center}
\vspace{-0.5cm}
\caption{The $k$-means method clusters the data in a circular manner while GMM manages to capture the elliptical cluster shapes:
(a) distance-based KM method formulates circular clusters;
(b) distribution-based GMM captures the elliptical cluster shapes.
}
\label{fig:GMM_example}
\end{figure}

Compared with the $k$-means algorithm, GMM maximizes only the likelihood, it will not bias the means towards zero, or bias the cluster sizes to have specific structures that might or might not apply~\cite{Pedregosa2011} (Fig.~\ref{fig:GMM_example}).
However, the structure becomes problematic when the data estimating the covariance is insufficient.
In aerodynamic applications, GMM is also preferred over $k$-means method.
\citet{Liem2015} completed aerodynamic data clustering using GMM, then proposed a mixture of experts approach to combine GEK models {based on the divide-and-conquer principle}.
The proposed approach was successfully demonstrated on conventional and unconventional aircraft configurations, where the surrogate model was used to represent the aircraft performance.

\subsubsection{Principal Component Analysis}
\label{sec:mlPCA}

\begin{figure}[h]
\centering
\includegraphics[width=\linewidth]{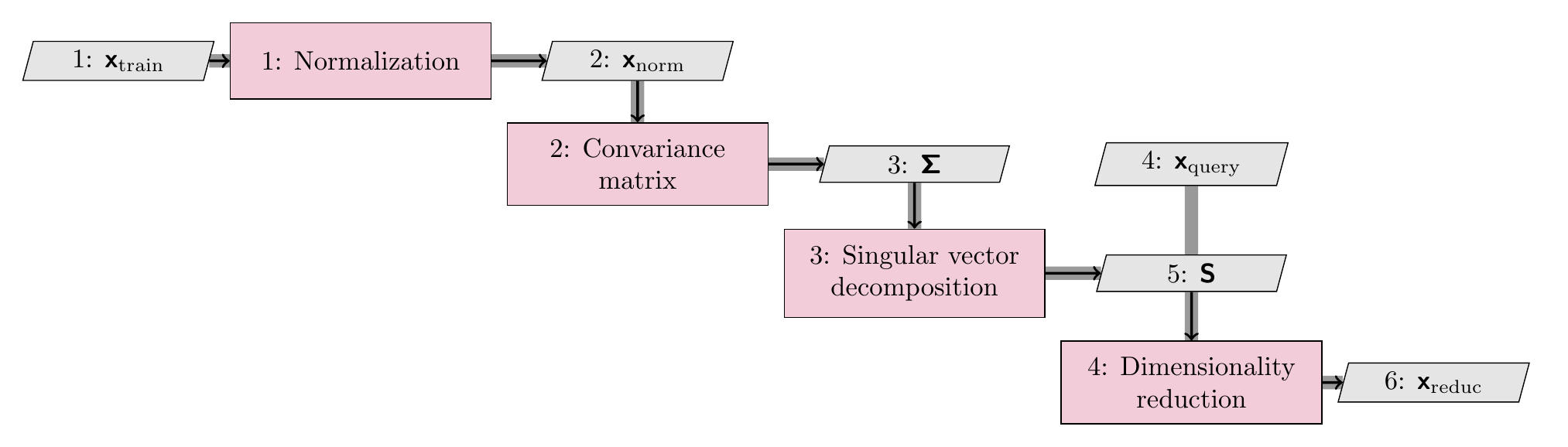}
\caption{PCA construction requires normalizing original data ($\mathbf{x}_\text{train}$) to the same order of magnitude ($\mathbf{x}_\text{norm}$) whose convariance matrix ($\mathbf{\Sigma}$) reveals the correlation among $\mathbf{x}_\text{norm}$.
SVD is a commonly used method to identify the eigenvectors and select the principal components ($\mathbf{S}$) based on the amount of physics information to preserve.
In the end, we can complete dimensionality reduction through matrix multiplication between $\mathbf{S}$ and query data ($\mathbf{x}_\text{query}$) and obtain reduced-dimension data ($\mathbf{x}_\text{reduc}$).
}
\label{fig:PCA}
\end{figure}

PCA is a mathematical algorithm that transforms high-dimensional data to low dimensions while maintaining the maximum variation in the data set~\cite{Jolliffe2016PrincipalCA,Shlens2014tutorial}.
The dimensionality reduction by PCA transformation simplifies the data visualization and improves the predictive modeling process.
PCA accomplishes such transformation by identifying directions (\emph{i.e}, principal components) along which the data variation is maximal to guarantee minimal information loss.
Identifying principal components is reduced to a problem of finding singular values and singular vectors, which a singular value decomposition (SVD) algorithm can solve.
 {SVD generalizes the eigendecomposition of a square normal matrix with an orthonormal eigenbasis to any matrix.
In particular, SVD of any $m \times n$ matrix $\boldsymbol{M}$ can be expressed as}
\begin{equation}
    \boldsymbol{M} = \boldsymbol{U} \boldsymbol{\Sigma} \boldsymbol{V}^\intercal,
\end{equation}
 {where $\boldsymbol{U}$ is an $m \times m$ unitary matrix, $\boldsymbol{\Sigma}$ is an $m \times n$ rectangular diagonal matrix with non-negative real numbers in the diagonal, and $\boldsymbol{V}$ is an $n \times n$ unitary matrix.
The diagonal elements of $\boldsymbol{\Sigma}$ are the singular values of $\boldsymbol{M}$, the columns of $\boldsymbol{U}$ are eigenvectors of $\boldsymbol{M}\boldsymbol{M}^\intercal$, and the rows of $\boldsymbol{V}$ are eigenvectors of $\boldsymbol{M}^\intercal\boldsymbol{M}$.}
 {The} singular values measure the variance retained by each principal component, while the eigenvectors with the highest singular values are the principal components.
Figure~\ref{fig:PCA} presents the key steps of a typical PCA algorithm.



\begin{figure}[h]
\begin{center}
    \subfigure[] {\includegraphics[   width=.49\textwidth]{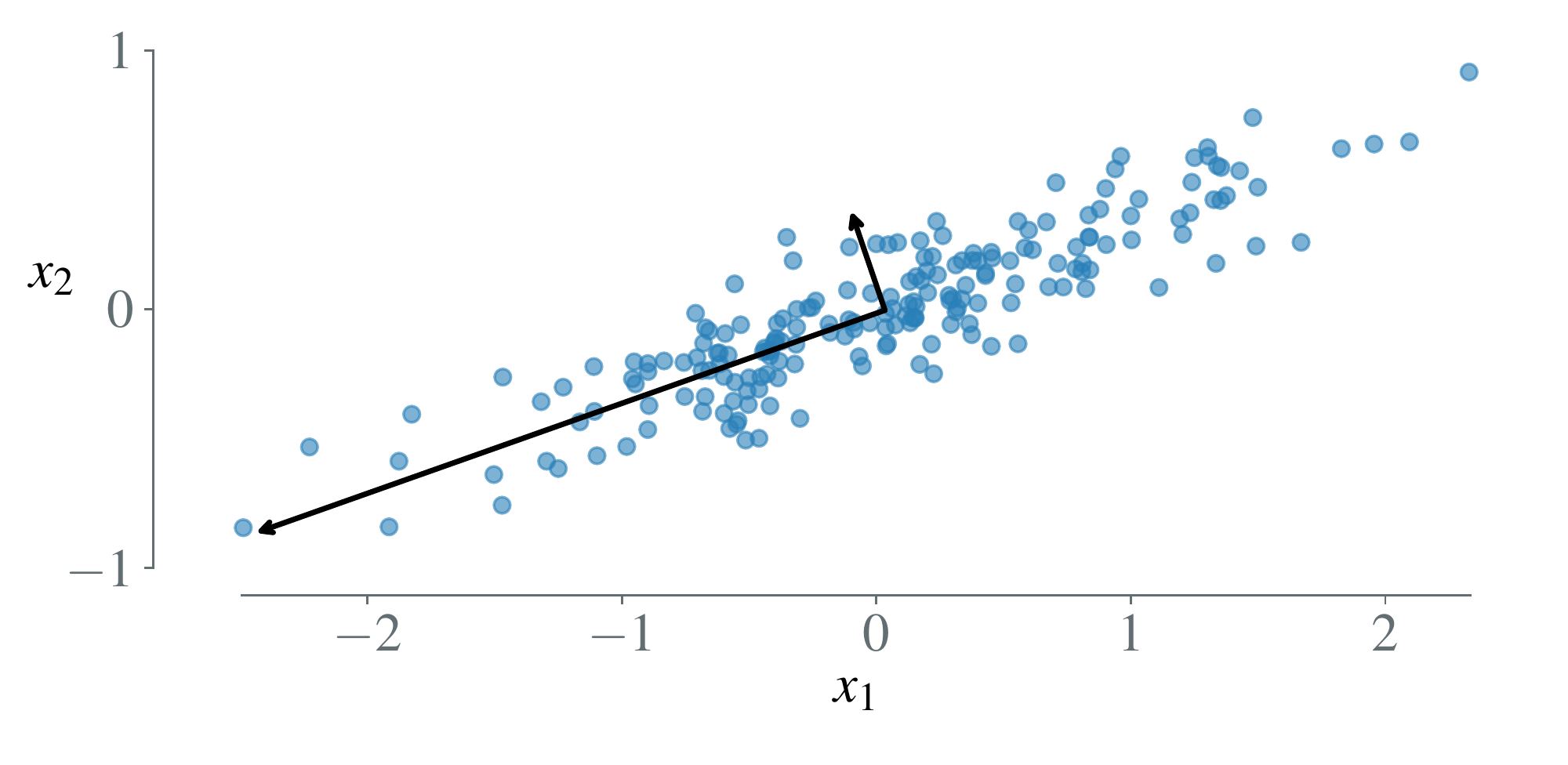}\label{}}
    \hspace{0.3mm}
    \subfigure[] {\includegraphics[
    width=.49\textwidth]{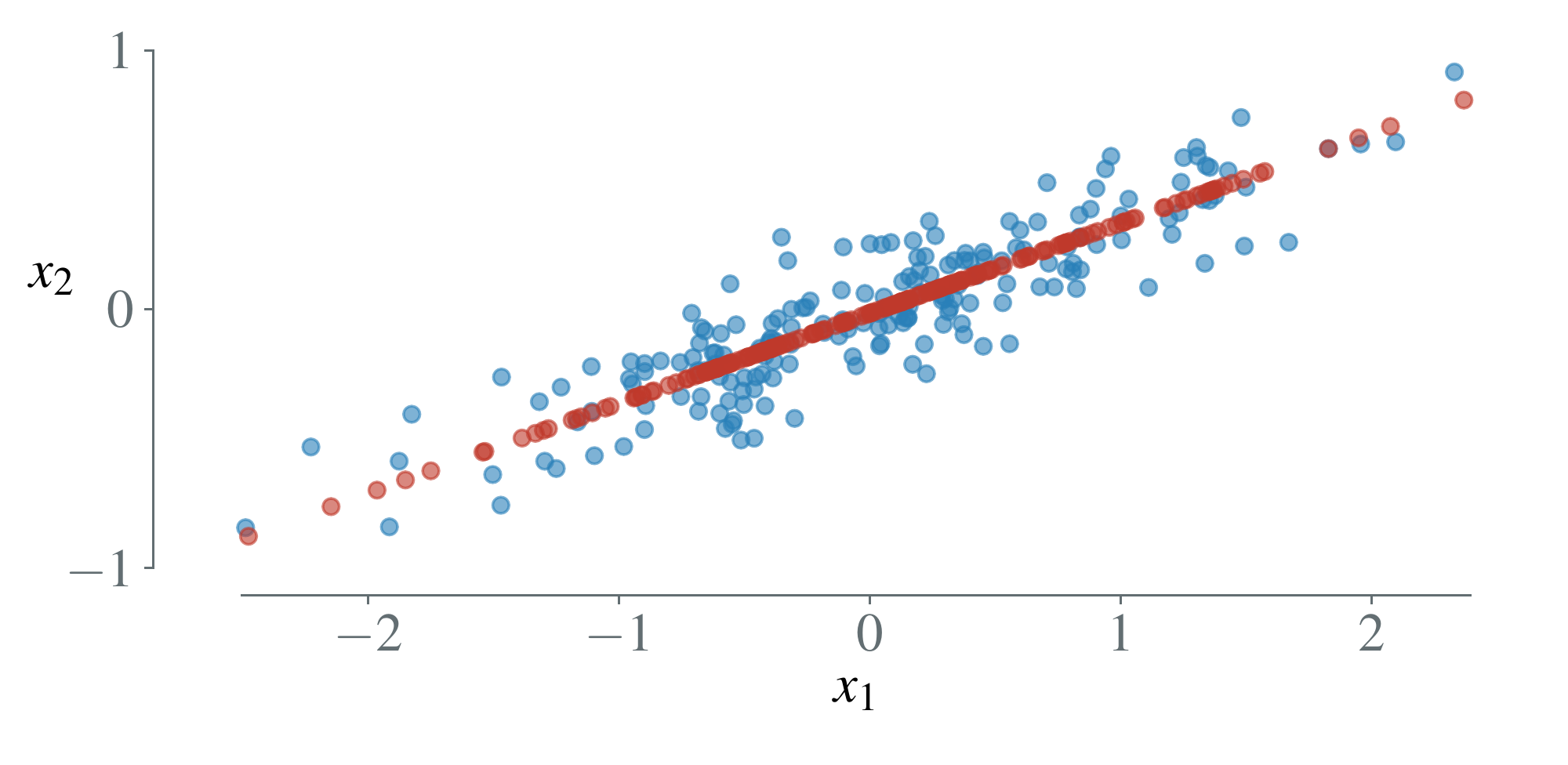}\label{}}
\end{center}
\vspace{-0.5cm}
\caption{Example illustrating how PCA transforms data~\cite{VanderPlas2016}: (a) blue dots are original data samples, the black arrows represent the principal axes of the data, and the length of the arrows is a measure of the variance of the data when projected onto that axis; (b) the red dots are projected data along the longest axis selected in (a) while short axis is removed such that PCA transforms the original two dimensions to one dimension.
}
\label{fig:PCA_example}
\end{figure}

In sum, PCA reduces dimensionality by finding a few orthogonal linear combinations of principal components  {(Fig.~\ref{fig:PCA_example})}.
PCA can find hidden patterns in a data set by identifying the correlated variables and reducing dimensionality by removing noisy and redundant information.

Standard PCA is a linear method that works well with linearly separable data sets; however, some data structures cannot be represented well in a linear subspace.
Kernel PCA extends standard PCA to efficient nonlinear dimensionality reduction by introducing a kernel function to measure the distance~\cite{Michael2001,Wang2014kernel}.
A commonly used kernel is the Gaussian kernel $\kappa(\boldsymbol{x}, \boldsymbol{x}^\prime) = \exp \big( - \| \boldsymbol{x} - \boldsymbol{x}^\prime \| ^2 / 2\sigma^2 \big)$, where $\sigma$ is an adjustable parameter.
For a data set with $n$ training points, the process process to construct kernel PCA can be summarized as follows:
\begin{enumerate}
    \item Construct the kernel matrix $\boldsymbol{K}_{i,j}=\kappa(\boldsymbol{x}_i, \boldsymbol{x}_j)$ directly from the training data.
    \item Compute the Gram matrix $\Tilde{\boldsymbol{K}} = \boldsymbol{K} - \boldsymbol{1}_n \boldsymbol{K} - \boldsymbol{K} \boldsymbol{1}_n + \boldsymbol{1}_n \boldsymbol{K} \boldsymbol{1}_n$ to ensure that the projected features have zero means, where $\boldsymbol{1}_n$ is an $n \times n$ matrix with all elements equal to $1/n$.
    \item Solve for $\boldsymbol{a}$ via performing eigenvalue decomposition ($\Tilde{\boldsymbol{K}} \boldsymbol{a}_k = \lambda_k n \boldsymbol{a}_k$).
\end{enumerate}
Then, the kernel-based principal components $y(\boldsymbol{x})$ can be calculated by $y_k(\boldsymbol{x}) = \phi(\boldsymbol{x})^\intercal \boldsymbol{v}_k = \sum\limits_{i=1}^n a_{k,i} \kappa(\boldsymbol{x}, \boldsymbol{x}_i)$.

Both standard PCA and kernel PCA methods have been successfully introduced into ASO.
For example, \citet{Asouti2016} applied standard PCA to reduce the input-space dimension to make the RBF network training easier and predictive performance more dependable.
They combined the trained RBF network with an evolutionary algorithm and managed to complete multiple ASO cases.
\citet{Gaudrie2020} applied kernel PCA to unveil the low-dimensional manifold of the high-dimensional CAD design parameters and facilitated the surrogate modeling of the Gaussian process, which was  {further} utilized for Bayesian optimization.

\subsubsection{Nonlinear manifold learning}
\label{sec:mlNml}

Manifold learning is a family of linear or nonlinear dimensionality reduction methods that learns the inherent low-dimensional structure from high-dimensional data~\cite{Izeman2012}.
We have already described linear manifold learning through PCA, so in this section, we focus on the nonlinear manifold learning approaches that are also widely used in ASO.
Manifold is a topological space with the property that the neighborhood space of each point on it resembles Euclidean space.
For example, the Grassmannian manifold $Gr(k, \mathbb{R}^{n})$ describes all $k$-dimensional linear subspaces in $\mathbb{R}^{n}$,
and the Stiefel manifold $\boldsymbol{V}_{k}(\mathbb{R}^{n})$ describes all orthonormal $k$-frames in $\mathbb{R}^{n}$.
Manifold learning hypothesizes that the high-dimensional data lies on a low-dimensional manifold.

\begin{figure}[h]
\centering
\includegraphics[width=\linewidth]{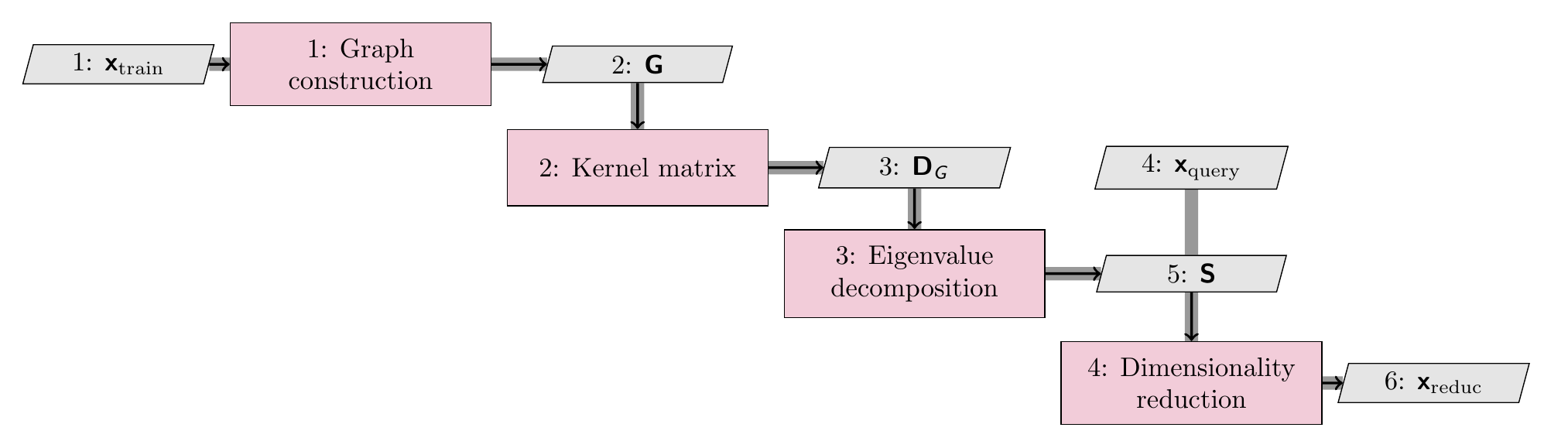}
\caption{Isomap first constructs neighborhood graph ($\mathbf{G}$) using KNN, then sets up the kernel matrix ($\mathbf{D}_G$) through the shortest path between any two training samples.
SVD is a commonly used method to identify the eigenvectors and select the principal components ($\mathbf{S}$) based on how much of the physics information we want to preserve.
In the end, we can complete the dimensionality reduction through matrix multiplication between $\mathbf{S}$ and the query data ($\mathbf{x}_\text{query}$) to obtain reduced-dimension data ($\mathbf{x}_\text{reduc}$).
}
\label{fig:Isomap}
\end{figure}

It is generally difficult to define the underlying low-dimensional manifolds of high-dimensional data  {analytically}, and thus manifold learning is needed.
The isometric feature mapping (Isomap)~\cite{Tenenbaum2000a} is one of the popular approaches for manifold learning.
For a data set with $N$ training points taken from the high-dimensional space $X$, Isomap can be fulfilled as shown in Fig.~\ref{fig:Isomap}.
 {Isomap enhances classical multidimensional scaling (MDS) by incorporating the geodesic distances imposed by a weighted graph.
Specifically, Isomap constructs the neighborhood graph over all observations by connecting $i$th and $j$th data points if point $i$ is one of the $k$ nearest neighbors of point $j$, sets the lengths of the edges equal to $d_{i,j}$, and then calculates the shortest paths between points.
Then we can compute lower-dimensional embedding through classical MDS, as follows:}
\begin{equation}
    \boldsymbol{X} = \boldsymbol{E}_r \boldsymbol{\Uplambda}^{1/2}_r,
\end{equation}
 {where $r$ is the required number of eigenvectors for lower-dimensional representation, and $\boldsymbol{E}_r$ is the matrix of $r$ eigenvectors.
The matrix $\boldsymbol{\Uplambda}_r$ is a diagonal matrix whose entries are the $r$ eigenvalues of the following matrix:}
\begin{equation}
    \boldsymbol{B} = -\frac{1}{2} \boldsymbol{C} \boldsymbol{D} \boldsymbol{C} ,
\end{equation}
 {where $\boldsymbol{D}$ is the squared distance matrix of $d_{i,j}^2$ and $\boldsymbol{C}$ is the centering matrix. This is defined as}
\begin{equation}
    \boldsymbol{C} = \boldsymbol{I} - \frac{1}{n} \boldsymbol{J}_n,
\end{equation}
 {where $\boldsymbol{I}$ is the $n \times n$ identity matrix and $\boldsymbol{J}_n$ is an $n \times n$ matrix of all ones.}

\citet{Orsenigo2013} compared PCA with Isomap in their ability to improve credit rating predictions of banks.
In the majority of cases, Isomap's low-dimensional representation resulted in the highest classification accuracy.
\citet{Ripepi2018} performed reduced-order modeling via PCA and Isomap to predict surface pressure distributions based on high-fidelity CFD simulations but at lower evaluation time and storage.
In most cases, Isomap shows higher predictive accuracy, especially near the shock-wave region.

Locally linear embedding (LLE)~\cite{Roweis2000a} is another manifold learning approach, which follows the same general process as Isomap (Fig.~\ref{fig:Isomap}) except that it computes the kernel matrix in a different way.
The kernel matrix in LLE corresponds to the weights $W_{ij}$ that best linearly reconstruct $x_i$ from its neighbors, which can be solved by minimizing the cost function 
\begin{equation}
    \mathcal{L} = \sum_i (x_i - W_{ij}x_j)^2,
\end{equation}
 {where each weight $W_{ij}$ corresponds to the amount of contribution the point $x_j$ has when reconstructing the point $x_i$.
The cost function is subject to two constraints: 
$W_{ij}$ is zero if $X_j$ is not one of the $k$ nearest neighbors of the point $X_i$; 
the sum of every row of the weight matrix equals 1, \emph{i.e.}, $\sum_j W_{ij}=1$}.
The manifold hypothesis that high-dimensional data tends to lie in the vicinity of a low dimensional manifold enables the local distance measurement in high-dimensional space~\cite{Narayanan2010,Fefferman2013}.
 {Therefore, LLE reduces the original data point collected in the original $D$ dimensional space to $d$ dimensions ($d \ll D$).
LLE finds the low-dimensional representation by minimizing the cost function}
\begin{equation}
    \mathcal{L} = \sum_i (y_i - W_{ij}y_j)^2,
\end{equation}
 {where the weights ($W_{ij}$) obtained in the original high-dimensional space are fixed and the minimization is performed by varying the points $y_i$ to optimize the coordinates.
Thus, LLE has several advantages over Isomap, including faster optimization when implemented to take advantage of sparse matrix algorithms and better results for many problems}~\cite{Roweis2000a}.
\citet{Decker2020} presented a study to compare dimensionality reduction methods, including PCA, Isomap, and LLE on analytical cases and a CFD application.
They concluded that nonlinear dimensionality reduction approaches performed better in the vicinity of shock and discontinuous regions while the linear method outperformed nonlinear methods for steady-state prediction.
In addition, they pointed out that nonlinear approaches discovered a lower-dimensional representation resulting in lower evaluation cost of nonlinear reduced-order models.

\subsubsection{Dynamic Mode Decomposition}

\begin{figure}[h]
\centering
\includegraphics[width=\linewidth]{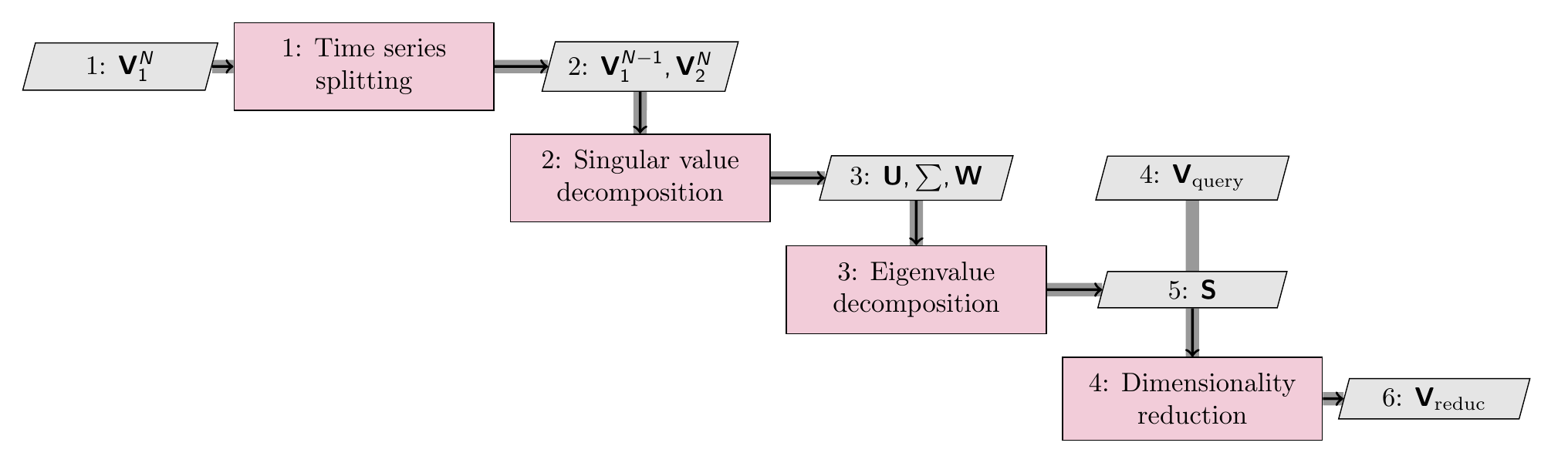}
\caption{DMD splits the time-series data ($\mathbf{V}_1^N$) into $\mathbf{V}_1^{N-1}$ and $\mathbf{V}_2^N$, completes SVD on $\mathbf{V}_1^{N-1}$, and then calculates the eigenvalues and eigenvectors of the SVD-formed matrix ($\mathbf{U}^\intercal V_2^NW\sum^{-1}$).
The selected modes ($\mathbf{S}$) can be used for dimensionality reduction.
}
\label{fig:DMD}
\end{figure}

Dynamic mode decomposition (DMD) is a data-driven decomposition approach to revealing spatio-temporal features of high-dimensional time-series data  {(Fig.~\ref{fig:DMD})}, such as unsteady flow fields.
For a time-series data set $\boldsymbol{V}_1^{N} = \boldsymbol{v}_1, \boldsymbol{v}_2, \dots, \boldsymbol{v}_{N}$ generated with a time interval of $\Delta t$, a linear mapping $\boldsymbol{A}$ is assumed to connect the data snapshot $\boldsymbol{v}_i$ to the subsequent snapshots $\boldsymbol{v}_{i+1}$, that is, $\boldsymbol{v}_{i+1} = \boldsymbol{A} \boldsymbol{v}_i$.
Then, we have $\boldsymbol{A}\boldsymbol{V}_1^{{N}-1} = \boldsymbol{V}_1^{{N}-1}\boldsymbol{S} + \boldsymbol{r}\boldsymbol{e}_{{N}-1}^{T}$, where $\boldsymbol{r}$ is the residual vector and $\boldsymbol{e}_{{N}-1} \in \mathbb{R}^{{N}-1}$ is a $({N}-1)\text{th}$ unit vector, and the eigenvalues of $\boldsymbol{S}$ can be used to approximate the eigenvalues of $\boldsymbol{A}$.
To improve the robustness, a ``full'' matrix $\hat{\boldsymbol{S}}$ can be used to perform eigenvalue decomposition, and $\hat{\boldsymbol{S}}$ is related to $\boldsymbol{S}$ via a similarity transformation of the following form:
\begin{equation}
  \hat{\boldsymbol{S}} = \boldsymbol{U} \boldsymbol{V}_2^{N} \boldsymbol{W} \boldsymbol{\Sigma}^{-1},
\end{equation}
where $\boldsymbol{U}$, $\boldsymbol{W}$, and $\boldsymbol{\Sigma}$ are obtained by performing a singular value decomposition of $\boldsymbol{V}_1^{{N}-1} = \boldsymbol{U} \boldsymbol{\Sigma} \boldsymbol{W}^{H}$.
The dynamic modes can be computed by $\boldsymbol{\Phi}_i = \boldsymbol{U}  \boldsymbol{y}_i$, where $\boldsymbol{y}_i$ is the $i\text{th}$ eigenvector of $\hat{\boldsymbol{S}}$, i.e., $\hat{\boldsymbol{S}} \boldsymbol{y}_i = \mu_i \boldsymbol{y}_i$.
The frequencies of dynamic modes can be obtained by the logarithmic mapping of corresponding DMD eigenvalues.
For the $i\text{th}$ mode with eigenvalue $\mu_i$, the frequency and growth rate are $f_i = \text{Im}({\ln{\mu_i}})/2\pi \Delta t$ and $g_i =\text{Re}({\ln{\mu_i}})/\Delta t$, which can also be expressed as $f_i = \angle \mu_i /2 \pi \Delta t$ and $f_i = |\mu_i|/ \Delta t$~\cite{Taira2019a}.

\citet{Schmid2010} applied DMD to extract dynamic information from flow fields for a better understanding of fluid-dynamical and transport processes.
They pointed out that DMD was capable of processing subdomains of the full computational or experimental domain.
This advantage enables DMD to focus on particular flow features and instability mechanisms, especially for flows containing a multitude of instability mechanisms or multiphysics phenomena.

\subsection{Semi-Supervised Learning}
\label{sec:mlSSL}

\begin{figure}[h]
\centering
\includegraphics[width=\linewidth]{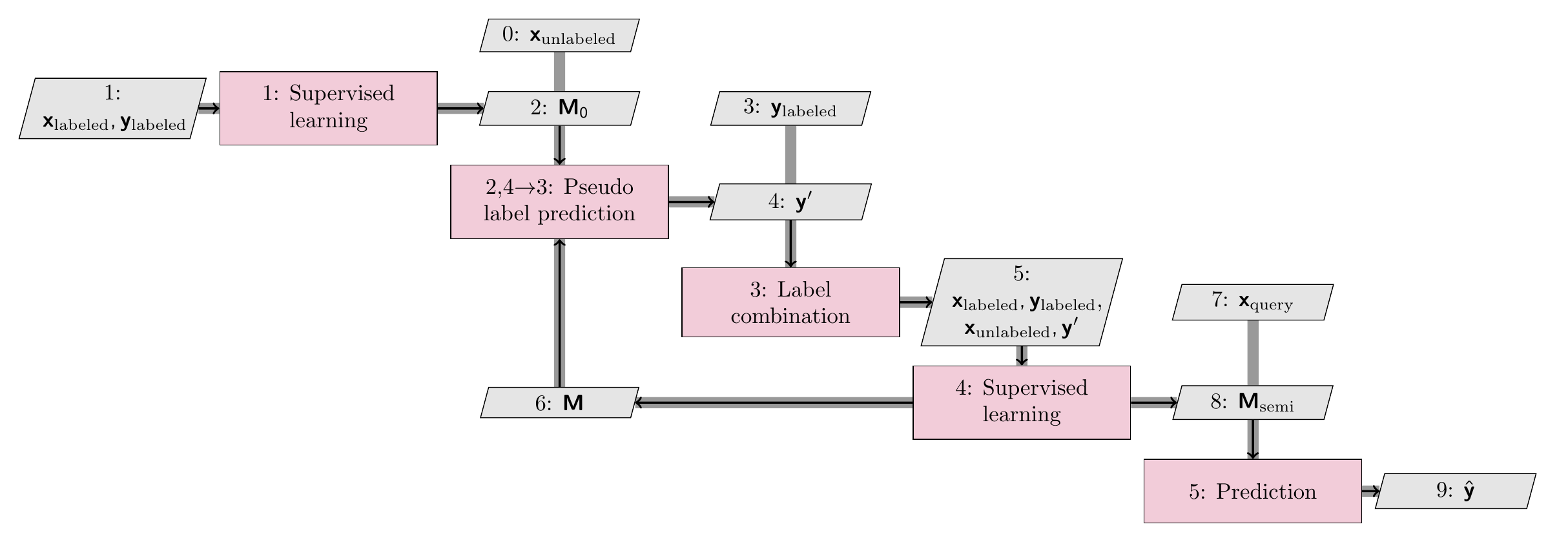}
\caption{Semi-supervised learning constructs an initial supervised learning model ($\mathbf{M}_0$) on labeled data and and predicts peudo labels ($\mathbf{y}'$) for unlabeled data using $\mathbf{M}_0$.
Then real and pseudo labels and the corresponding data are combined to train another supervised learning model ($\mathbf{M}$) to predict pseudo labels.
This step is iterated until predictive performance does not change to obtain the trained model $\mathbf{M}_\text{semi}$ for further regression tasks on query data ($\mathbf{x}_\text{query}$).
}
\label{fig:Semi}
\end{figure}

As stated previously, supervised learning works with labeled data, while unsupervised learning works with unlabeled data.
However, there are circumstances where data labels (e.g., experimental aerodynamic data) are incomplete and costly to obtain, leading to semi-supervised learning.
Semi-supervised learning methods work with a small amount of labeled data and a large amount of unlabeled data~\cite{Ouali2020overview,Yang2021survey}.
Semi-supervised learning manages to train models by combining the few existing labels and pseudo labels (Fig.~\ref{fig:Semi}).





\begin{figure}[h]
\centering
\includegraphics[trim=20 120 20 120, clip, width=1.0\linewidth]{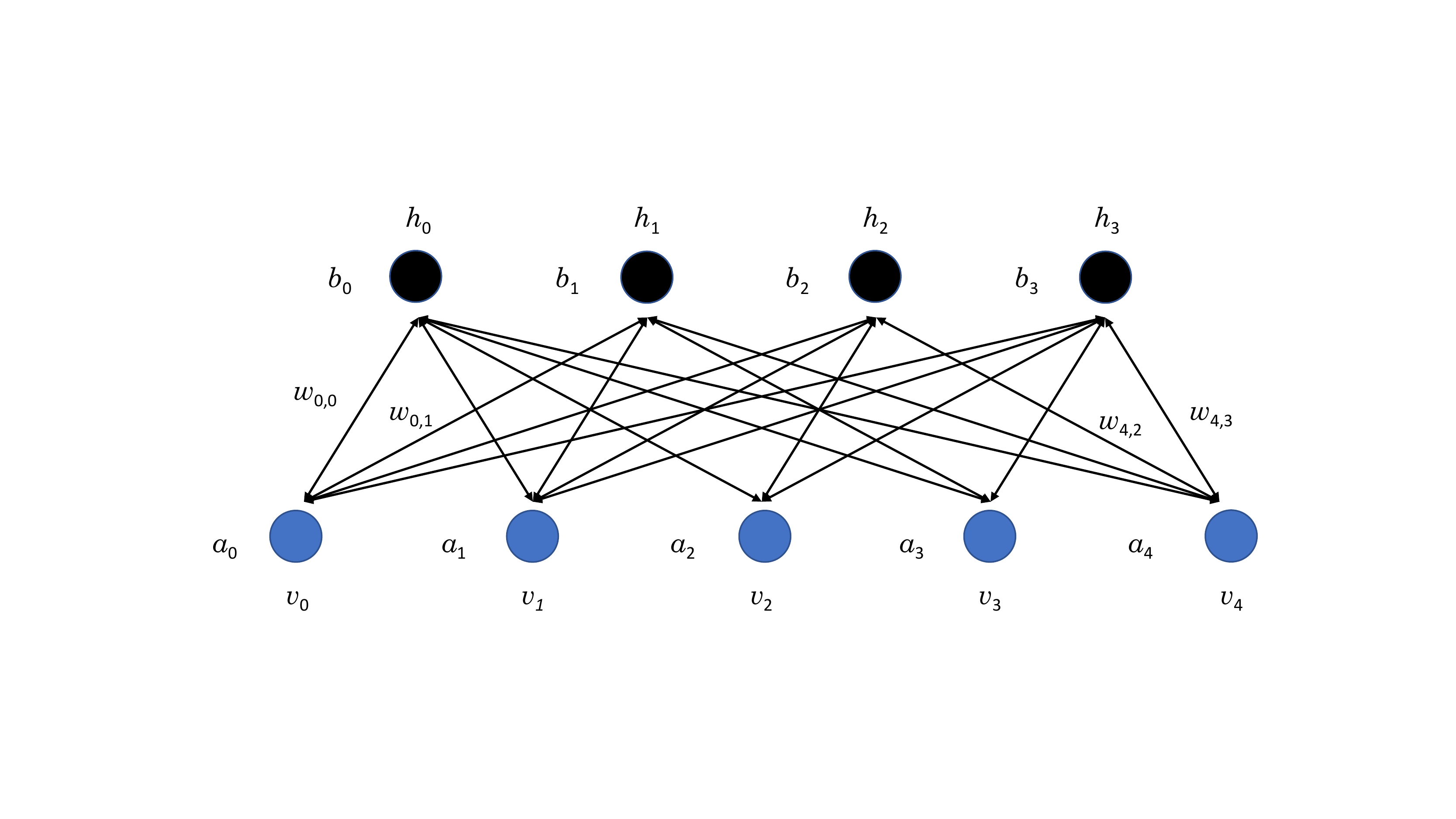}
\caption{RBM is a shallow, two-layer neural network.
The first layer (blue) is called visible layer or input layer with visible neurons ($v_0,v_1,\ldots,v_4$) while the second layer (black) is called hidden layer with hidden neurons ($h_0,h_1,h_2,h_3$).
The weights ($w_{i,j}$) represents the connection between $i$th visible neuron and $j$th hidden neuron while $a_i$ and $b_j$ are biases on visible and hidden neurons, respectively.
}
\label{fig:RBM}
\end{figure}

\begin{figure}[h]
\centering
\includegraphics[width=\linewidth]{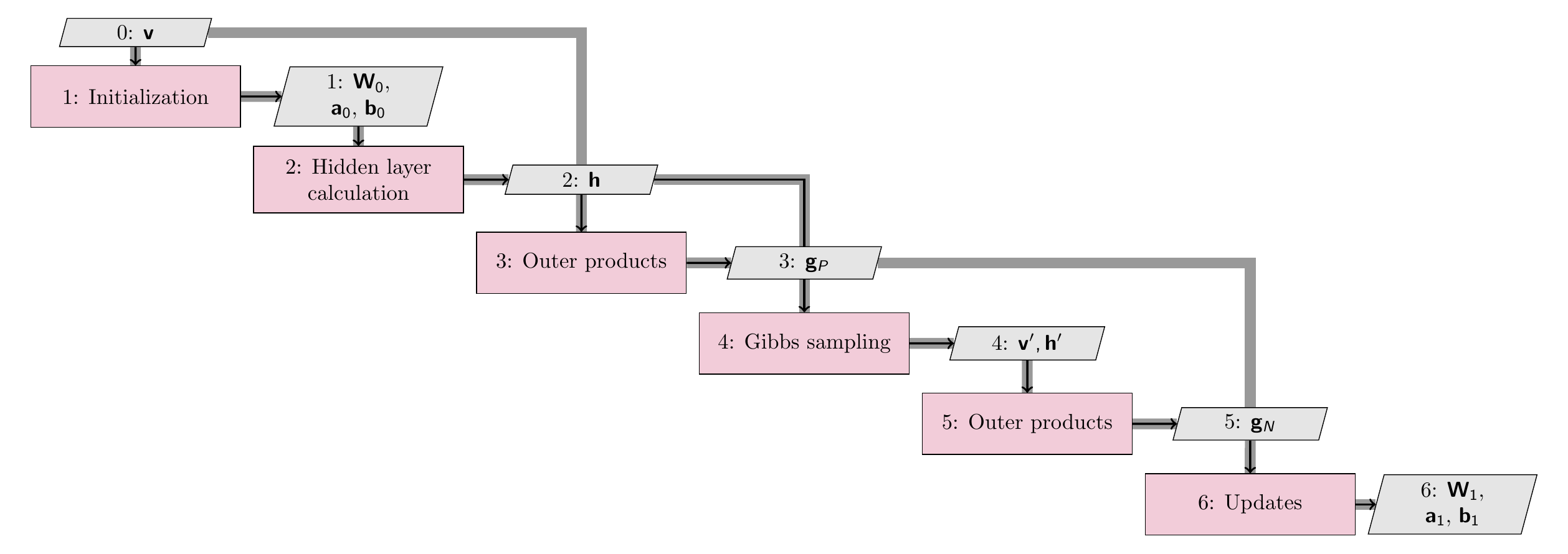}
\caption{One step of the contrastive divergence-based RMB training shows how weights and biases get updated.
Specifically, the forward pass computes hidden vector $\mathbf{h}$, which leads to the positive gradient ($\mathbf{g}_P$) through outer products between $\mathbf{v}$ and $\mathbf{h}^\intercal$.
The Gibbs sample strategy produces $\mathbf{v}'$ through a backward pass and $\mathbf{h}'$ through another forward pass on $\mathbf{v}'$, which leads to the negative gradient ($\mathbf{g}_N$).
Eventually, we can update the weights and biases using Eqns.~\ref{eqn:W}, \ref{eqn:a}, and~\ref{eqn:b}.
}
\label{fig:RBM_CD}
\end{figure}

A semi-supervised learning method used in ASO is the deep belief network (DBN), which is a stack of restricted Boltzmann machines (RBM).
RBM is a two-layer ANN (Fig.~\ref{fig:RBM}) with generative capabilities to learn a probability distribution over its input~\cite{Montufar2018restricted,Upadhya2019}.
``Restricted" refers to the only connections between the visible and hidden layers.
Specifically, every node in the visible layer is connected to every node in the hidden layer, while the nodes in the same layer have no connections between each other.
Such ``restriction" allows for easy implementation and efficient training.
Multiple stacked RBM models can be fine-tuned through the process of gradient descent and backpropagation to make a DBN model.
The hidden layer activates the forward pass (from a visible layer to a hidden layer).
In contrast, visible layers reconstruct the inputs on the backward pass (from hidden layer to visible layer).
RBM is undirected, so it cannot adjust the weights and biases via gradient descent and backpropagation.
An alternative training approach is to approximate the gradient information through Markov chain Monte Carlo (MCMC) but MCMC requires many steps to reach the equilibrium state.
Contrastive divergence truncates the MCMC method at its $k$-th step to approximate the gradient information to adjust the unknown parameters~\cite{Perpinan2005,Hinton2002}.
The $k$ can be as small as 1 leading to the 1-step contrastive divergence to train RBM as follows (Fig.~\ref{fig:RBM_CD}):
\begin{enumerate}
    \item Randomly initialize the unknown weights and hidden-layer biases.
    \item Take a training sample $\boldsymbol{v}$, compute the probabilities of the hidden units and sample a hidden activation vector $\boldsymbol{h}$.
    \item Compute the outer product of $\boldsymbol{v}$ and $\boldsymbol{h}$ and call it the positive gradient.
    \item Sample the reconstructed $\boldsymbol{v}'$ of the visible units from $\boldsymbol{h}$, then resample the hidden vector $\boldsymbol{h}'$ from this. This step is also called Gibbs sampling step.
    \item Compute the outer product of $\boldsymbol{v}'$ and $\boldsymbol{h}'$ and call it the negative gradient.
    \item Update the weight matrix as follows:
    \begin{equation}
        \boldsymbol{W} = \boldsymbol{W} + \lambda(\boldsymbol{v}\boldsymbol{h}^\intercal - \boldsymbol{v}'\boldsymbol{h}'^\intercal),
    \label{eqn:W}
    \end{equation}
    where $\lambda$ is a pre-set learning rate.
    Similarly, we update biases by
    \begin{equation}
        \boldsymbol{a} = \boldsymbol{a} + \lambda(\boldsymbol{v}-\boldsymbol{v}'),
    \label{eqn:a}
    \end{equation}
    and
    \begin{equation}
        \boldsymbol{b} = \boldsymbol{b} + \lambda(\boldsymbol{h}-\boldsymbol{h}').
    \label{eqn:b}
    \end{equation}
\end{enumerate}

RBM is computationally efficient.
Its training is faster than traditional Boltzmann machine because of the restrictions on connections between intra-layer nodes.
Activations of the hidden layer can be used as input to other models as useful features to improve performance.
However, the training is still challenging, even with the contrastive divergence algorithm.

\begin{figure}[!h]
\centering
\includegraphics[trim=00 20 00 20, clip, width=0.8\linewidth]{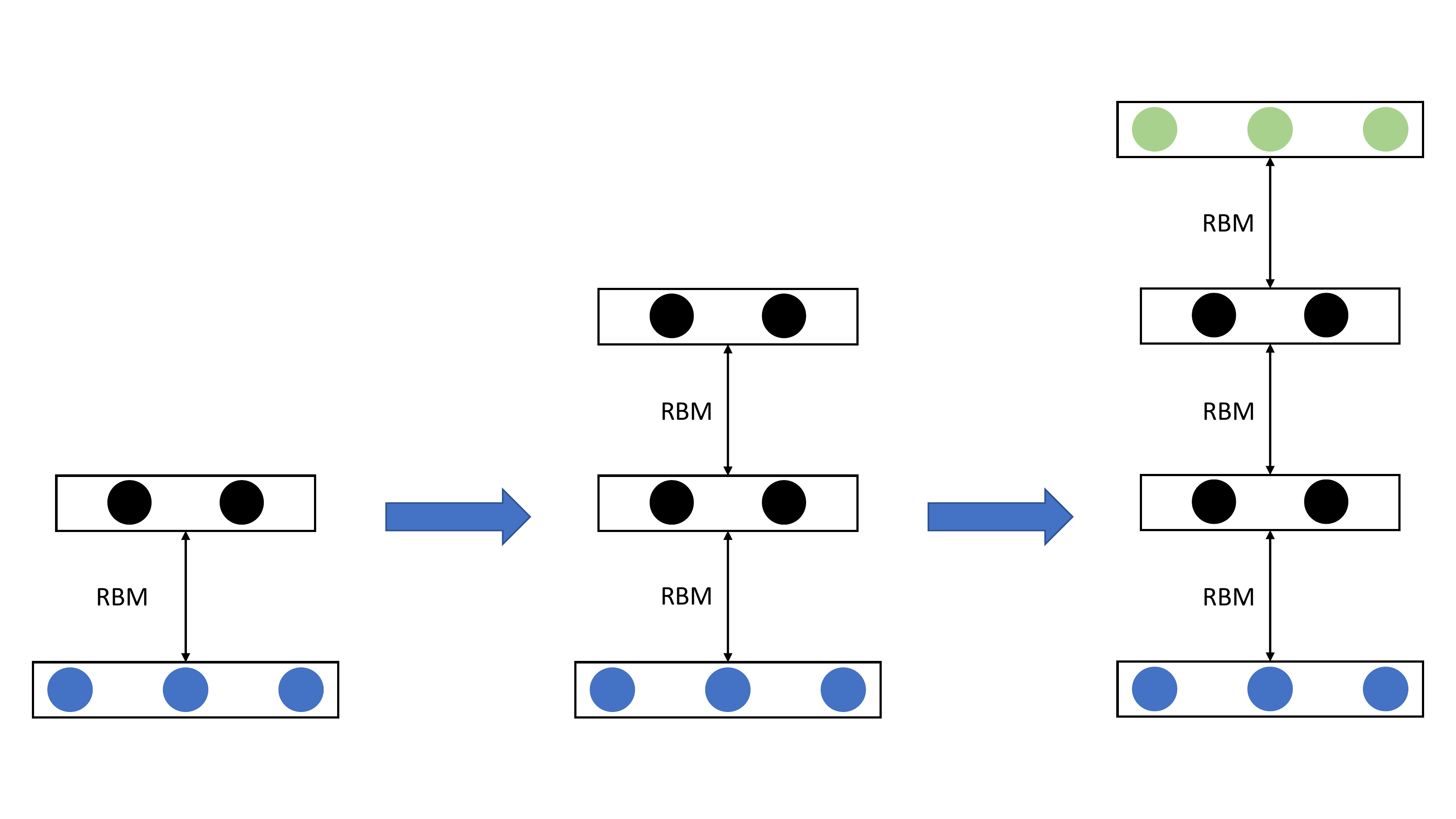}
\caption{RBM models are sequentially connected to formulate DBN.
Each RBM is trained until convergence, then frozen;
the result of the ``output'' layer of the machine is then fed as an input to the next RBM in the sequence, which is then trained until convergence, and so on until the entire network has been trained.
Blue is the input layer, black is a hidden layer, and green is the output layer.
}
\label{fig:DBN}
\end{figure}

DBN stacks RBM models on top of one another sequentially trains the RBM models in an unsupervised manner and then fine-tunes the model using supervised learning techniques~\cite{Hinton2006,Hinton2009DeepBN} (Fig.~\ref{fig:DBN}).
This deeper stacking setup of DBN improves RBM's performance.
\citet{Mohamed2012} claimed DBN as a very competitive alternative to GMM because of the following reasons:
DBN can be fine-tuned as neural networks;
DBN has many nonlinear hidden layers;
and DBN is generatively pre-trained.
DBN training avoids backpropagation, which may result in local optima or ``vanishing gradient" (the gradient will be vanishingly small due to the backpropagation, effectively preventing the weight from changing its value).
We notice successful DBN application in multi-fidelity robust aerodynamic design optimization where DBN was trained and used as a low-fidelity model~\cite{Tao2019}.

\subsection{Reinforcement Learning}
\label{sec:mlRl}

Reinforcement learning~\cite{Sutton2018,Kaelbling1996} (RL) is a kind of ML approach aiming to solve sequential decision-making problems.
In other words, RL aims to learn an optimal policy guiding an agent to move in an environment, which is normally formulated as a Markov decision process.
 {In this section, we describe the general RL architecture and basic terminology, followed by deep RL (DRL) enabled through deep learning.}

\subsubsection{Reinforcement Learning Fundamentals}
\label{sec:RL}

\begin{figure}[h]
\centering
\includegraphics[width=0.65\linewidth]{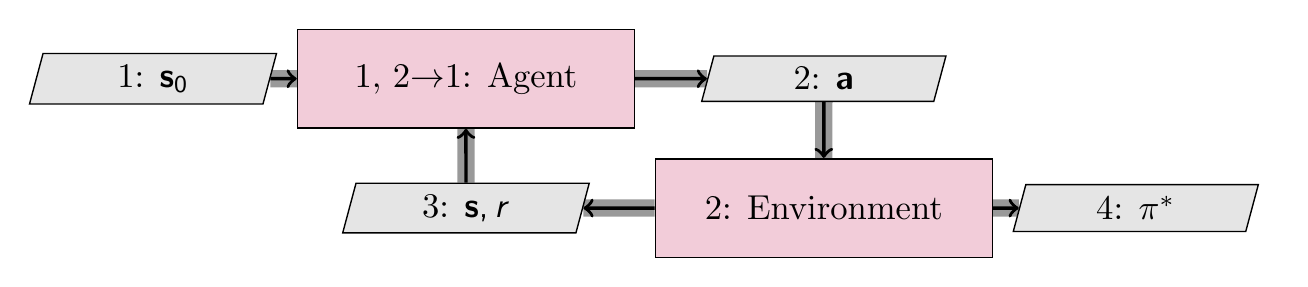}
\caption{RL starts with an initial observation of the environment ($\mathbf{s}_0$), based on which the agent takes actions ($\mathbf{a}$).
The action is executed in the environment, and then an updated state ($\mathbf{s}$) is transported back to the agent to take further action decisions.
The reward $r$ is computed after each action execution to guide how to modify the agent's policy.
RL iterates this process until an optimal policy ($\mathbf{\pi}^*$) is learned.}
\label{fig:RL}
\end{figure}

During the RL model training (Fig.~\ref{fig:RL}), the agent learns to take the right actions by interacting with the environment.
Some key concepts in RL are explained as follows.

\begin{enumerate}
    \item Agent: An agent controls the action to take based on the environment observation.
    In real life, the agent (artificial intelligence) can be a drone that makes a delivery.

    \item Environment: An environment is where the agent moves through, executing the agent's actions and responding to the agent with the reward and next state.
    For a drone delivery application, for example, the environment updates a drone's location and outputs the reward depending on the drone's movement.

    \item State: A state  {($s \in \mathcal{S}$)} is an observation of the environment, which the agent will use to take action.

    \item Action: An action  {($a \in \mathcal{A}$)} is a possible move the agent can make at each time step.
    In the drone delivery example, an action can be turning right or left, cruising, or accelerating.

    \item Trajectory: A sequence of states and actions that influence those states.
    
    \item Reward: A reward  {($r \in \mathcal{R}$)} is feedback used to measure the quality of the agent's action.
    A reward can be immediate or delayed to evaluate the agent's actions.

    \item Discount factor: A discount factor ( {$\gamma$}, generally smaller than 1) is multiplied by future rewards as a mathematical trick to make an infinite sum finite.
    In addition, a discount factor smaller than 1 also reveals we value future rewards less than immediate rewards.

    \item Policy: The policy ($\pi(s)$) is the core algorithm that the agent follows to take actions. 
    The policy is a map from the state space to the action space. 
    At each time step, the agent will follow the action  {($a$)} generated from the policy according to the measured state  {($s$)}.
     {The policy mapping can be either deterministic: $\pi(s) = a$, or stochastic: $\pi(a|s)=P_{\pi}[A=a | S=s]$, where $P_{\pi}$ is the probability of action ($a$) is taken in state $s$ following the current policy $\pi$.}

    \item  {Return: $G$ measures the future reward as a total sum of discounted rewards going forward.
    For example, the return starting from time step $t$ is computed as}
    \begin{equation}
        G_t = R_{t+1} + \gamma R_{t+2} + \dotsi = \sum_{k=0}^{\infty} \gamma^k R_{t+k+1}.
    \end{equation}

    \item Value:  {Each state is associated with a value function $V_{\pi}(s)$, or state-value, predicting} the expected future cumulative discounted reward under the constructed policy.
     {In other words, $V_{\pi}(s)$ quantifies how good a state is and formulates as}
    \begin{equation}
        V_{\pi} (s) = \mathds{E}_{\pi}[G_t | S_t = s].
    \end{equation}

    \item $Q$-value: In contrast to the state-value, the $Q$-value  {($Q_{\pi}(s,a)$, also known as action-value)} refers to the long-term return of the action under the policy from the current state.
     {Similarly as state-value, $Q$-value formulates as}
    \begin{equation}
        Q_{\pi} (s, a) = \mathds{E}_{\pi}[G_t | S_t = s, A_t = a],
    \end{equation}
     {through which we can see the relationship between $V_{\pi}(s)$ and $Q_{\pi}(s,a)$} given by
    \begin{equation}
        V_{\pi}(s) = \sum_{a \in \mathcal{A}} Q_{\pi}(s,a) {\pi}(s,a).
    \end{equation}
    
\end{enumerate}

The two main types of RL algorithms are model-based and model-free algorithms.
 {Here, ``model'' refers to a transition probability function ($P$) and reward function ($R$).
The transition function records the probability of transitioning from state $s$ to $s^\prime$ after taking action $a$ while obtaining reward $r$}
\begin{equation}
    P(s^\prime, r | s, a) = \mathds{P}[S_{t+1}=s^\prime, R_{t+1}=r | S_t=s, A_t=a],
\end{equation}
 {which leads to state-transition function}
\begin{equation}
    P_{ss^\prime}^a = P(s^\prime|s,a) = \mathds{P}[S_{t+1}=s^\prime | S_t=s, A_t=a] = \sum_{r \in \mathcal{R}} P(s^\prime, r | s, a).
\end{equation}
 {The reward function $R$ predicts the next expected reward triggered by one action}
\begin{equation}
    R(s,a) = \mathds{E} [R_{t+1}|S_t=s,A_t=a] = \sum_{r \in \mathcal{R}} r \sum_{s^\prime \in \mathcal{S}} P(s^\prime,r|s, a).
\end{equation}
 {Model-based RL algorithms rely on a model that predicts the outcomes of actions to learn optimal policies.
Model-free RL algorithms directly interact with the environment without constructing the model}~\cite{Sutton2018,Garnier2021}.
 {In either method, RL aims to train an agent to formulate optimal policy to drive actions that maximize the total reward.
There can be more than one optimal policy; all the optimal policies are donated as $\pi^*$.
The optimal policies share the same state-value, which is the optimal state-value function,} 
\begin{equation}
    V^* \equiv \max_{\pi} V_{\pi}(s),
\end{equation}
 {for all $s \in \mathcal{S}$.
Similarly, we denote optimal action-value function as}
\begin{equation}
    Q^*(s,a) \equiv \max_{\pi} Q_{\pi} (s,a),
\label{eqn:optQ}
\end{equation}
 {for all $s \in \mathcal{S}$ and $a \in \mathcal{A}$.
The optimal action-value function gives the expected return for taking action $a$ in state $s$ and thereafter following an optimal policy.
Thus, we can represent $Q^*(s,a)$ with $V^*(s)$ as~\cite{Sutton2018}}
\begin{equation}
    Q_*(s,a) = \mathds{E} [R_{t+1} + \gamma V^*(S_{t+1}) | S_t=s,A_t=a].
\end{equation}

Compared with supervised, unsupervised, and semi-supervised learning, which aim to learn data patterns from a training set and then apply them to a new data set, RL is the process of dynamically learning by adjusting actions based on continuous feedback to maximize accumulated reward. 
Because of the different principles in nature, RL has the following major advantages and disadvantages.
On one hand, RL solves complex problems, such as optimal control problems~\cite{Bertsekas2019reinforcement}, that  {are intractable} when using conventional techniques;
RL can achieve long-term good performance as the long-term accumulated reward is used in training it;
RL is similar to human learning, which makes RL particularly powerful. 
On the other hand, too much RL can lead to an overload of states, which can diminish the results.
On the other hand, RL requires a lot of data and computational budget and is subject to the ``curse of dimensionality" issue for real physical systems.
Combining RL and deep learning to be introduced in the following section alleviates the above-mentioned drawbacks.

\subsubsection{Deep Reinforcement Learning}
\label{sec:mlDRL}

Deep RL (DRL) improves the standard RL method by using DNN to model the agent policy and thus facilitates RL to handle large-scale complex problems.
Like the training of other DNNs in supervised learning, the training process of DRL iteratively adjusts the weights and biases of the agent DNN.
The difference is that supervised learning has the ground-truth labels to be predicted beforehand, while RL needs to wait for the environment-returned rewards, which can be varied, delayed, or affected by unknown variables.
Based on the way to obtain the optimal RL policy, there are two categories of DRL: value-based methods and policy-based methods~\cite{Mnih2015,Francois2018,Garnier2021}.

\begin{figure}[h]
\centering
\includegraphics[width=0.8\linewidth]{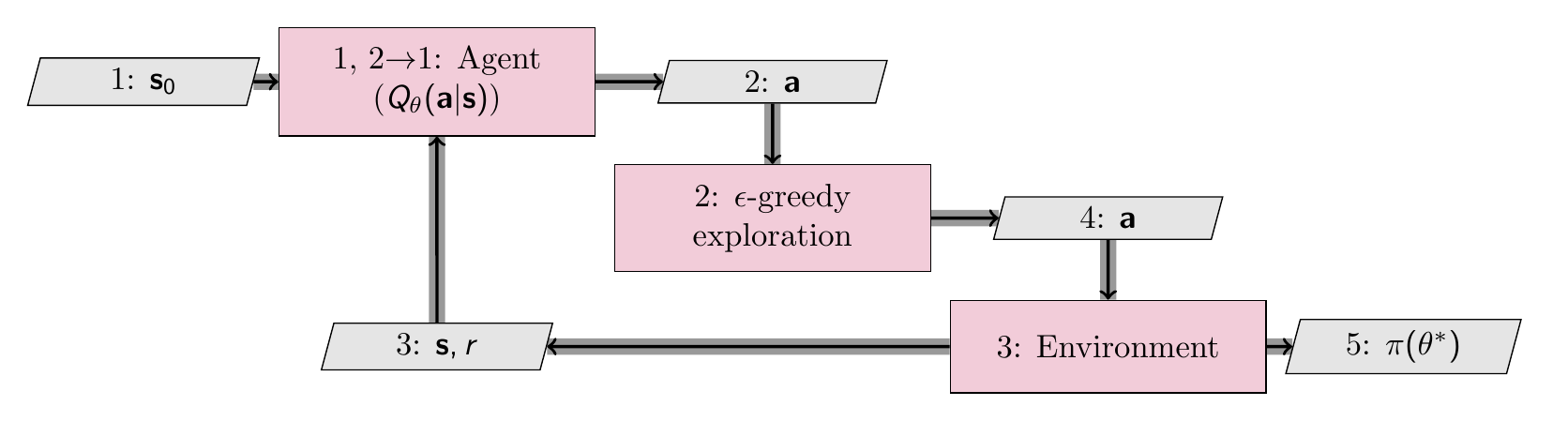}
\caption{
As a value-based RL, deep $Q$-network trains a value function ($Q_\mathbf{\theta} (\mathbf{a} | \mathbf{s})$) using DNN to evaluate the agent policy $\pi$.
The $\epsilon$-greedy strategy is typically used to ensure a suitable exploration that takes either the action with the highest value or a random action with a user-defined probability.
The hyperparameters of the DNN ($\mathbf{\theta}$) can be updated after each action.
}
\label{fig:DQN}
\end{figure}

In value-based methods, DNN is used to approximate a value function that estimates the expected reward of any state-action pairs.
The agent policy is determined by selecting the action with the highest value.
These methods are generally applicable to problems with a discrete action space.
Deep $Q$-Network~\cite{Mnih2015} is a popular value-based RL (Fig.~\ref{fig:DQN}), which shares a similar structure as vanilla $Q$-learning.
The core idea of $Q$-learning is to keep track of the $Q$-value for the different state-action pairs the agent can take~\cite{Hasselt2015deep,Mnih2015}.
If there are a massive number of intermediate states, the required memory rapidly expands and $Q$-learning becomes practically impossible.
Deep $Q$-networks solve this issue by estimating the $Q$-value functions with DNN.
In particular, the DNN reads states as the input and predicts $Q$-values for all possible state-action pairs.
The loss function is usually obtained from the Bellman equation~\cite{Bellman1962} to minimize the $Q$ value prediction error from the environment's feedback.
Training the DNN requires an adequate exploration of the environment, which is typically the $\epsilon$-greedy strategy.

\begin{figure}[h]
\centering
\includegraphics[width=0.8\linewidth]{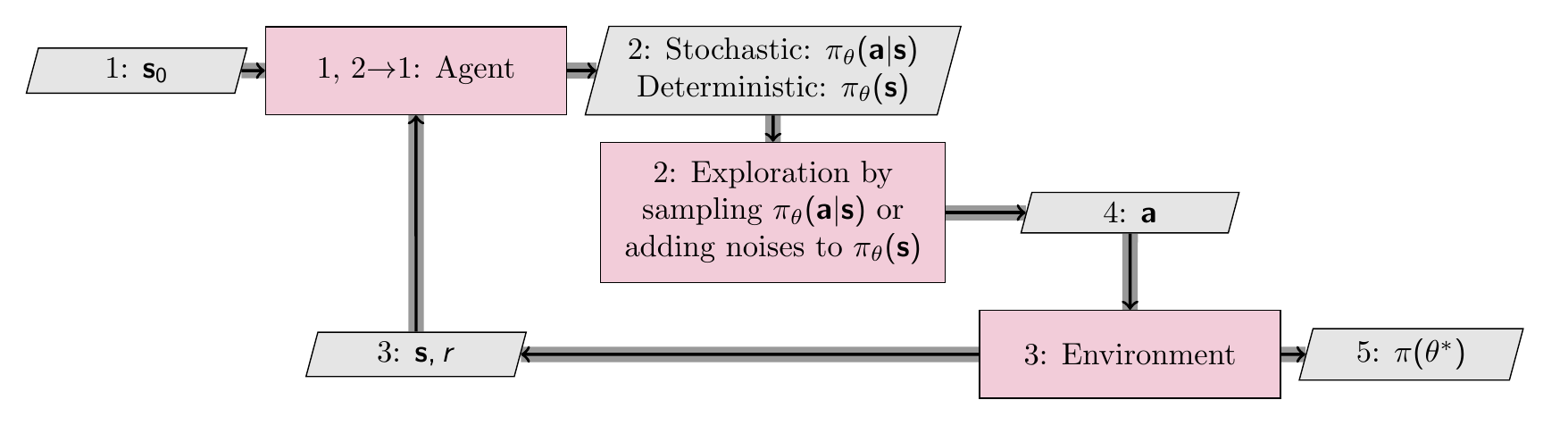}
\caption{
Policy-based RL directly models the policy ($\pi$) using DNN, which can be stochastic or deterministic.
The actions are generated by sampling the stochastic policy or adding noise to the deterministic action to ensure an adequate exploration.
}
\label{fig:DPG}
\end{figure}

In policy-based methods (Fig.~\ref{fig:DPG}), DNN is directly used to model the policy to take actions, and this makes these methods applicable to continuous action spaces.
The policy-based method will firstly run an episode (containing all states that come in between an initial state and a terminal state) to maximize the cumulative reward.
Then, it increases the probability of high-return actions and decreases the probability of low-return actions.
Policy-based DRL can be stochastic or deterministic.
A stochastic policy (such as the proximal policy optimization method~\cite{schulman2017proximal}) models the probability distribution of action (with respect to a state) to maximize the expected cumulative reward, which integrates over both state and action spaces.
In the training process of a stochastic policy, to explore the environment, the action is generated by sampling the distribution governed by the policy.
A deterministic policy (such as the deterministic policy gradient method~\citep{silver2014deterministic}) directly maps the state to the optimal action, so it merely integrates over the state space.
The exploration in training a deterministic policy can be realized by adding noise to the action.
Training a deterministic policy usually requires fewer episodes than training a stochastic one~\cite{silver2014deterministic}.

As summarized by \citet{Garnier2021}, policy-based methods converge on optimal parameters more quickly and reliably than value-based RL.
Nevertheless, policy-based methods may get stuck on local optima instead of the global optimum.
Policy-based methods work better with high-dimensional action spaces because deep $Q$-networks have to assign a score to every possible action for all time steps.

DRL incorporates deep learning into RL, allowing agents to make decisions from unstructured input data without manual engineering of state space.
Making use of deep learning enables DRL to deal with high-dimensional states (such as pixels rendered to the screen in a video game) and decide what actions to perform to optimize an objective (eg. maximizing the game score).
We have noticed a broad range of DRL applications for video games~\cite{Torrado2018,Shao2019survey}, robotics~\cite{Gu2016deep,Nguyen2019}, transportation~\cite{Liu2018deep,Farazi2021}, flow control~\cite{Rabault2019a,Li2022b}, \emph{etc}.
DRL can mimic human intuition to solve ASO problems.
\citet{Li2020d} applied DRL to learn a policy to reduce the aerodynamic drag of supercritical airfoils.


\subsection{Artificial Neural Networks}
\label{sec:mlFdm}

\begin{figure}[h]
\centering
\includegraphics[trim=00 50 00 50, clip, width=0.8\linewidth]{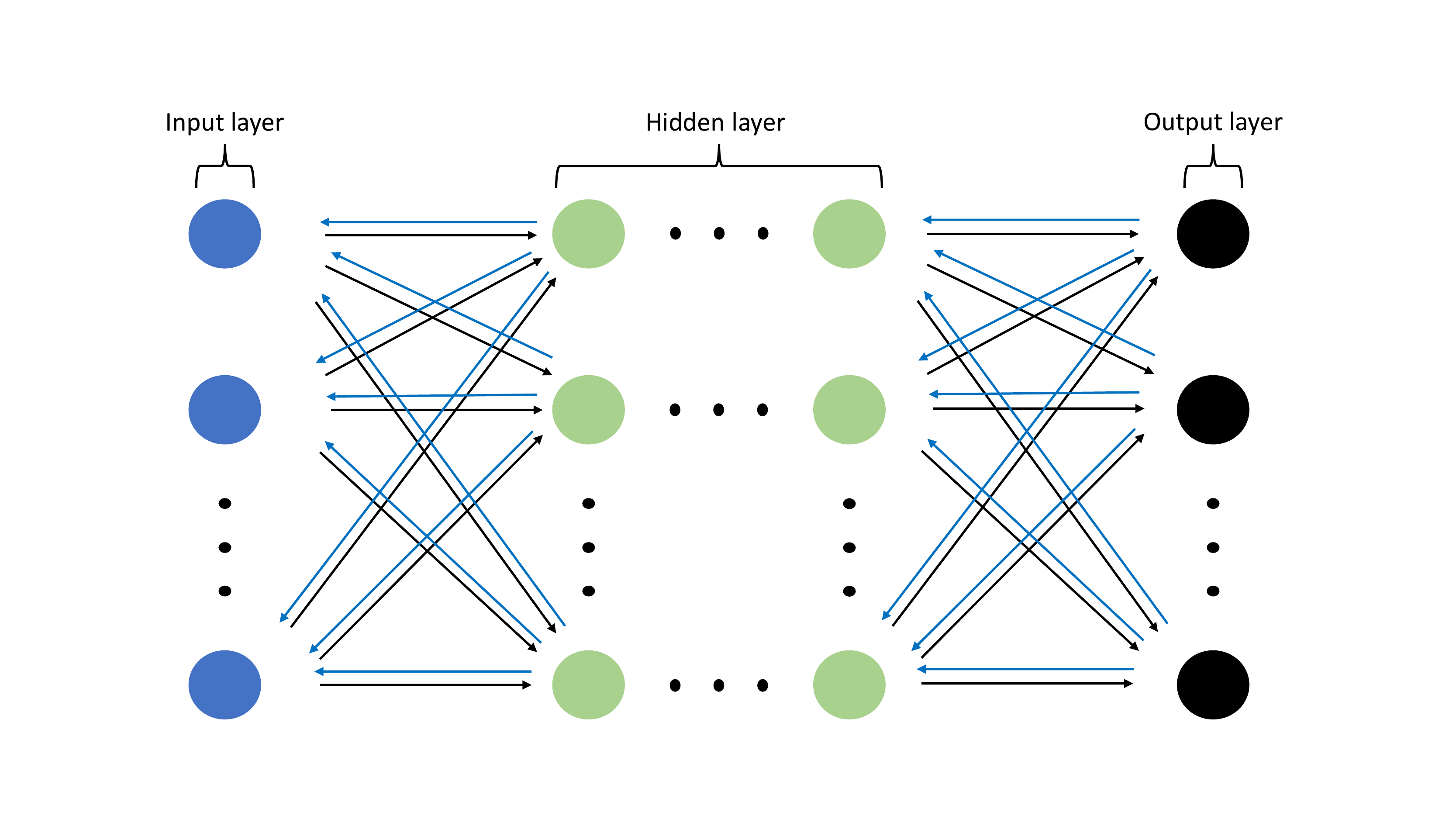}
\caption{ANN consists of input layer (blue dots), hidden layer (green dots), and output layer (black dots), and each dot is called neuron.
Forward pass (black arrows) maps input layers to output layer through multiplication and addition operations, followed by activation function for nonlinearity within each hidden-layer neuron.
Backpropagation (blue arrows) tracks the gradient information of the output layer with respect to hidden layers and hidden layers with respect to the input layer.
Thus, the forward pass predicts outputs given inputs, whereas backpropagation leads the model training to adjust unknown weight and bias parameters.
}
\label{fig:ANN}
\end{figure}

ANN is the core component of deep learning.
They are powerful and scalable, making them ideal for tackling large-scale and highly complex ML tasks~\cite{Emmert2020,Mocanu2018} (Fig.~\ref{fig:ANN}).
ANN has become the most popular ML model in various areas.
Most of the recent advances in ML-based ASO utilize ANN models.
In this section, we introduce the ANN fundamentals followed by a few typical and novel ANN architectures, including convolutional neural networks (CNN), recurrent neural networks (RNN), autoencoders, generative adversarial networks (GAN), and the self-organizing maps (SOM).
In addition, we also elaborate on the PINN model.

\subsubsection{Basic Setup}
\label{sec:mlBS}

The ANN architecture consists of one input layer, one or more hidden layers, and one output layer.
A deep stack of hidden layers makes ANN a DNN.
The input layer is associated with the input parameters, while the output layer is associated with quantities of interest to be predicted.
Each layer contains one or multiple neurons.
Within one neuron, we have the following operation:
\begin{equation}
    h = a(w_1 x_1 + w_2 x_2 + \dots + w_n x_n + b) = a( \boldsymbol{x}^\intercal \boldsymbol{w} + b),
\end{equation}
where $\boldsymbol{x}$ is an input vector, $\boldsymbol{w}$ is a vector of weights to be determined, $b$ is an unknown bias, $a$ is an activation function that injects nonlinearity into neural networks, and $h$ is the output of this neuron that is used as an input for the next hidden layer.
Typical activation functions include the sigmoid function,
\begin{equation}
    \sigma(z) = \frac{1}{1 + e^{-z}},
\end{equation}
which is monotonic and differentiable; its output exists within $[0, 1]$ and its derivative is not monotonic.
The hyperbolic tangent activation function is
\begin{equation}
    \text{tanh}(z) = \frac{e^z - e^{-z}}{e^z + e^{-z}},
\end{equation}
which is monotonic and differentiable; its output exits within [-1, 1] and itsderivative is not monotonic.
The \texttt{ReLU} activation functions is defined as
\begin{equation}
    R(z) = \max(0, z),
\end{equation}
which is monotonic and differentiable; its output is within $[0, \infty]$ and derivative is monotonic.
The \texttt{Leaky ReLU} activation function is formulated as
\begin{equation}
L(z) =
\begin{cases}
    z,& \text{if } z\geq 0\\
    \gamma z,    & \text{otherwise},
\end{cases}
\end{equation}
where $\gamma$ is a small-value constant defining the ``leakage" such that \texttt{Leaky ReLU} is monotonic and differentiable.
Its output exits within $[-\infty, \infty]$ and its derivative is monotonic.
It is important to understand the monotonicity, range, and differentiability of activation functions to select them intelligently.


\subsubsection{Model Training}
\label{sec:mlMT}

The key principle of neural network model training lies in a maximum likelihood estimate where we optimize the unknown parameters to maximize the probability of observing output data conditioned on the inputs~\cite{Martins2021}.
Thus, we formulate the objective function (also known as loss function) as a sum of the squared errors between model predictions and real observations:
\begin{equation}
    \min_{\boldsymbol{\theta}} \sum_{i=1}^N (\hat{y}_i(\boldsymbol{\theta}) - y_i)^2,
\end{equation}
where $\boldsymbol{\theta}$ is the unknown parameter vector, $y$ is the real observation in training data, $\hat{y}$ is the neural network model prediction.
The unknown weights and biases in ANN are usually randomly initialized and then iteratively adjusted to minimize the loss function.
ANN models involve matrix multiplications and activation functions, which are all differentiable.
In addition, ANN models typically involve a large number of unknown parameters.
Therefore, gradient-based optimization algorithms prevail.
One way of calculating the gradient is to analytically compute the derivative for each neuron; however, it may be practically impossible since modern ANN models can have thousands of neurons.
Using the finite-difference method is also computationally expensive and suffers from issues like step length selection~\cite[Sec. 6.4]{Martins2021}.
Because ANN typically has many inputs and few outputs, and the network structure consists of many differentiable simple functions chained together, they are well suited for reverse-mode algorithmic differentiation (AD)~\cite[Sec. 6.6]{Martins2021}.
The ML community refers to the reverse AD as backpropagation, which is built into modern ML software packages, such as \texttt{Tensorflow}~\cite{Abadi2016}.

Among the available gradient-based optimization algorithms, the most popular choice in ANN training is the steepest descent algorithm (also known as gradient descent), even though it is the simplest and not the most efficient in general.
However, gradient descent works well for training ML models.
Finding the global minimum of the loss function for a large-scale ANN is hard; however, it is more important to find a good enough solution quickly than to find the real global minimum.
Gradient descent in ML uses a pre-selected step size (called the learning rate) instead of a line search to make the training more efficient.
The learning rate is usually gradually decreased when approaching the end of training to avoid missing the minimum.

The training can take all available training data at every step, which is called batch gradient descent (BGD).
BGD uses the mean gradient of the whole data set to update the unknown weights that move directly towards a local or global optimum solution for convex or relative smooth error manifolds.
BGD can guarantee a minimum within the basin of attraction given an annealed learning rate (decaying learning rate).
Nevertheless, using the whole training data set at each step is computationally expensive and makes the algorithm more likely to get trapped in local optimal.

In contrast to BGD, stochastic gradient descent (SGD) takes one random training sample at every step and computes the gradients~\cite{Bottou2012stochastic}.
SGD accelerates the algorithm because the stochasticity helps avoid local optima.
The drawback of the stochasticity is that the loss function does not decrease monotonically and never settles at the minimum.
One solution is to gradually reduce the learning rate during the training to settle at the global minimum.
Mini-batch gradient descent (MGD)~\cite{Khirirat2017,Hinton2012} is a strategy between BGD and SGD.
MGD takes a random subset of the whole training data to compute the gradients.
Generally, MGD is better than BGD at getting out of local minima but not as good as SGD.
MGD converges more smoothly than SGD.

\subsubsection{Convolutional Neural Networks}
\label{sec:mlCNN}

\begin{figure}[h]
\centering
\includegraphics[width=\linewidth]{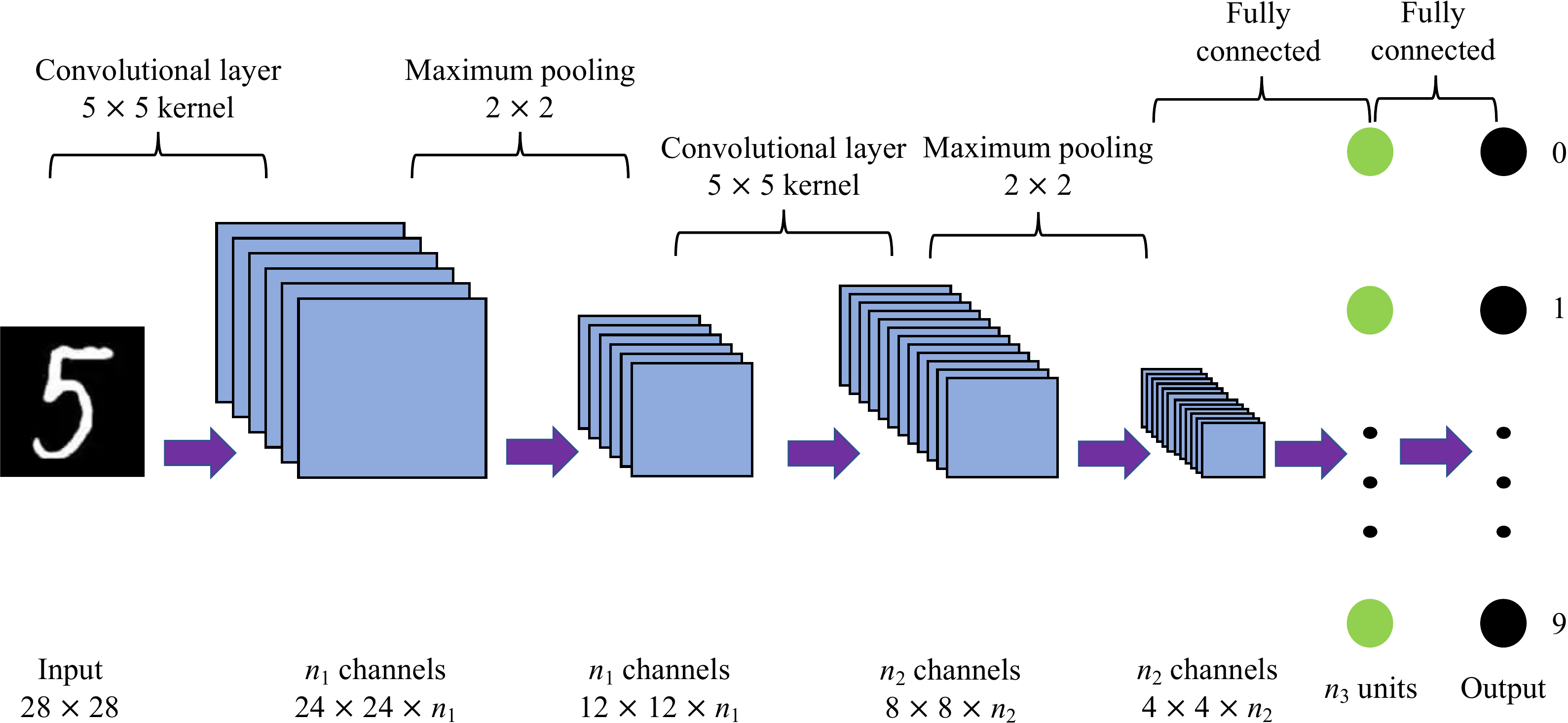}
\caption{Taking as an example an MNIST digit with a resolution of $28 \times 28$, convolutional layers are in charge of converting image to multiple channels and extracting high-level features through local kernels.
Pooling reduces the number of unknown parameters further.
Then, the resulting channels are converted to one fully-connected layer to complete the classification task.
}
\label{fig:CNN}
\end{figure}

CNN is a specialized type of DNN designed for large-scale structured data such as images.
Its architecture makes the implementation more efficient and vastly reduces the number of parameters in the network~\cite{Krizhevsky2012ImageNetCW,Albawi2017}.
Therefore, CNN  {is the state-of-the-art in computer vision field applications}. 
A CNN architecture is mainly composed of convolutional layers, pooling layers, and fully connected layers (Fig.~\ref{fig:CNN}).

Convolutional layers achieved by filters (also known as convolutional  {kernels}) capture the global and local information of the input data.
The term convolution refers to the mathematical combination of two functions to produce a third function to merge information.
CNN performs convolution on the input data and a convolutional layer to produce a feature map.
Mathematically, the convolutional layer can be formulated as
\begin{equation}
    \text{ConvL} = (w \cdot I)_{i,j} = a \left(\sum\limits_{m=0}^{l_1-s} \sum\limits_{n=0}^{l_2-s}
                            w_{m,n} \cdot I_{i+m,j+n} + b \right),
\end{equation}
where $I$ is a two-dimensional input with a of $L_1 \times L_2$ matrix, $w$ is unknown weight matrix with a size of $l_1 \times l_2$, $s$ is the stride length, $b$ is the bias, $a$ is the activation function, and $\text{ConvL}$ is the output of a convolutional layer with the size of $(L_1 + 2 \times p - l_1 + s) \times (L_2 + 2 \times p - l_2 + s)$ where $p$ is the padding value.

A pooling layer usually follows a convolutional layer to reduce the dimension of the data to avoid overfitting~\cite{Rao2018Over,Xiao2021Over}.
The most common pooling operations are maximum and average pooling.
This process is completed via a filter striding through the output of a convolutional layer.
The maximum pooling selects the maximum value within the local filter region and then strides to the next local region to do the same operation.
Similarly, the average pooling calculates the average value within the local filter region.
A fully connected layer is simply a stack of multiple neurons followed by activation functions.

When dealing with large-scale structured inputs, CNN has the following advantages over normal  {ANN}:
(1) fewer unknown parameters due to the use of filters;
(2) more effective in recognition problems because CNN learns a spatial hierarchy of patterns (\emph{i.e.}, forming higher CNN layers by combing lower layers);
(3) translation invariant property, that is, once a pattern is learned at one location, CNN can identify this pattern at any other locations because the learned weights are reusable even if the input is shifted or rotated.

However, a CNN typically requires a large amount of data to be well trained, and not all tasks can be formulated with structured inputs.
Some applications of CNN in aerodynamic predictions are based on converting the unstructured data to structured images\cite{bhatnagar2019,Zhang2018b}.
This strategy may not be preferable for high-fidelity modeling despite the proof-of-concept.

\subsubsection{Recurrent Neural Networks}
\label{sec:mlRNN}

\begin{figure}[h]
\centering
\includegraphics[trim=20 120 20 120, clip, width=1.0\linewidth]{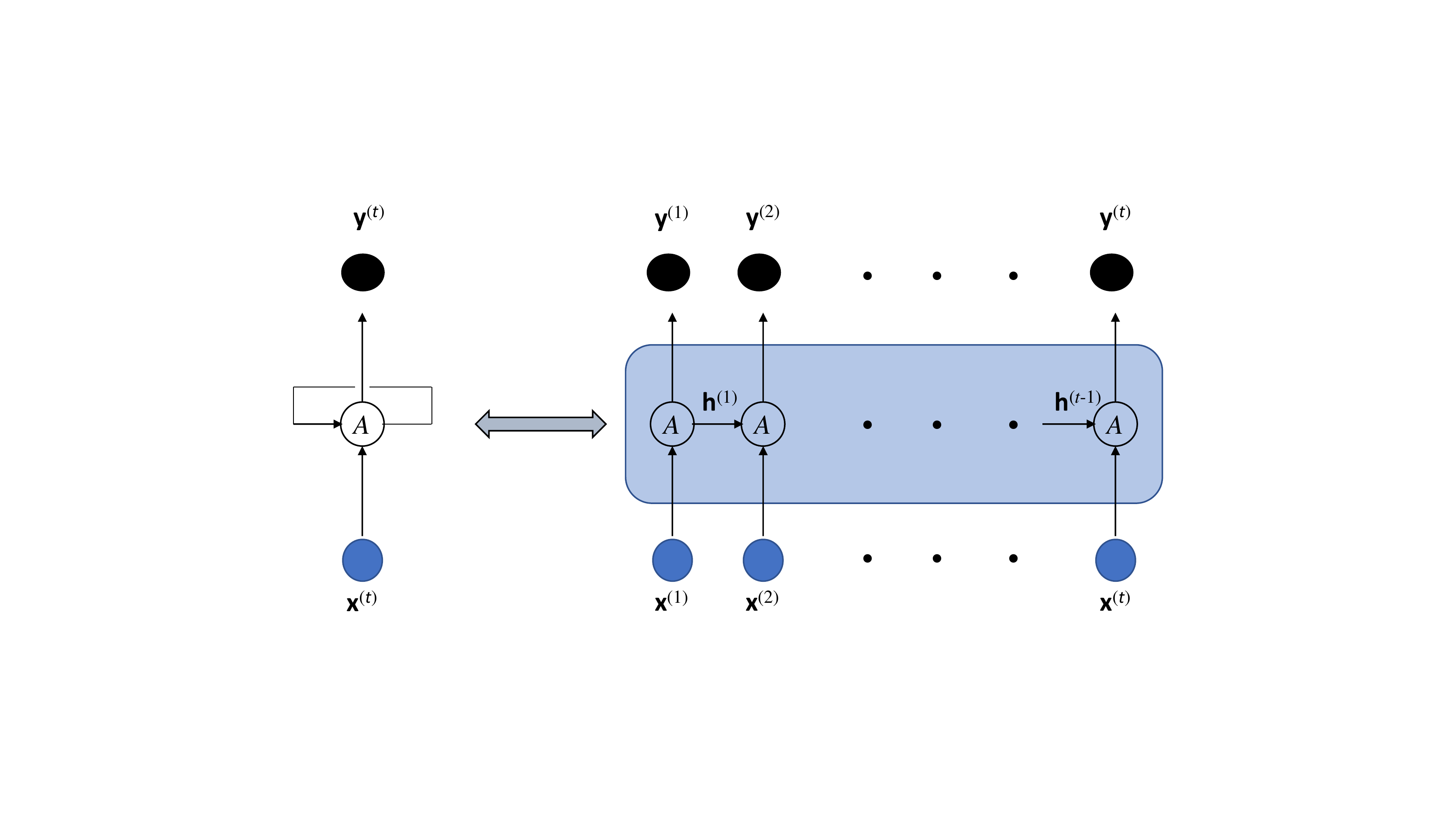}
\caption{RNN is widely used for time-dependent tasks. They consist of a series of recurrent structures, where each structure takes as the input the corresponding time-step input ($\mathbf{x}^{(t)}$) and the hidden state ($\mathbf{h}^{(t-1)}$) from the previous structure, and then produces the current-structure hidden state, which predicts the output and proceeds to the subsequent structure.
}
\label{fig:RNN}
\end{figure}

RNN gets its name from a feedback loop that points a hidden neuron back to itself, constituting a recurrent structure~\cite{Li_2018_CVPR,Sherstinsky2020}.
This recurrent structure allows previous outputs to be used as inputs while having hidden states (Fig.~\ref{fig:RNN}).
Therefore, RNN has the memory of historical information, which makes it suitable to handle time sequence data.
A traditional RNN connects the inputs and outputs as follows:
\begin{equation}
    h^{t} = a_h \left( W_{xh}x^{t} + W_{hh}h^{t-1} + b_h \right),
\end{equation}
\begin{equation}
    y^{t} = a_o \left( W_{ho}h^{t} + b_o \right), 
\end{equation}
where $x^t$ is the time-sequence instance at step $t$, $h$ is the hidden neuron output, $y$ is the model output, $W$ is the weight matrix, $a_h$ is the activation function for $h$ while $a_o$ is the activation function for $y$.

\begin{figure}[]
\centering
\includegraphics[width=0.5\linewidth]{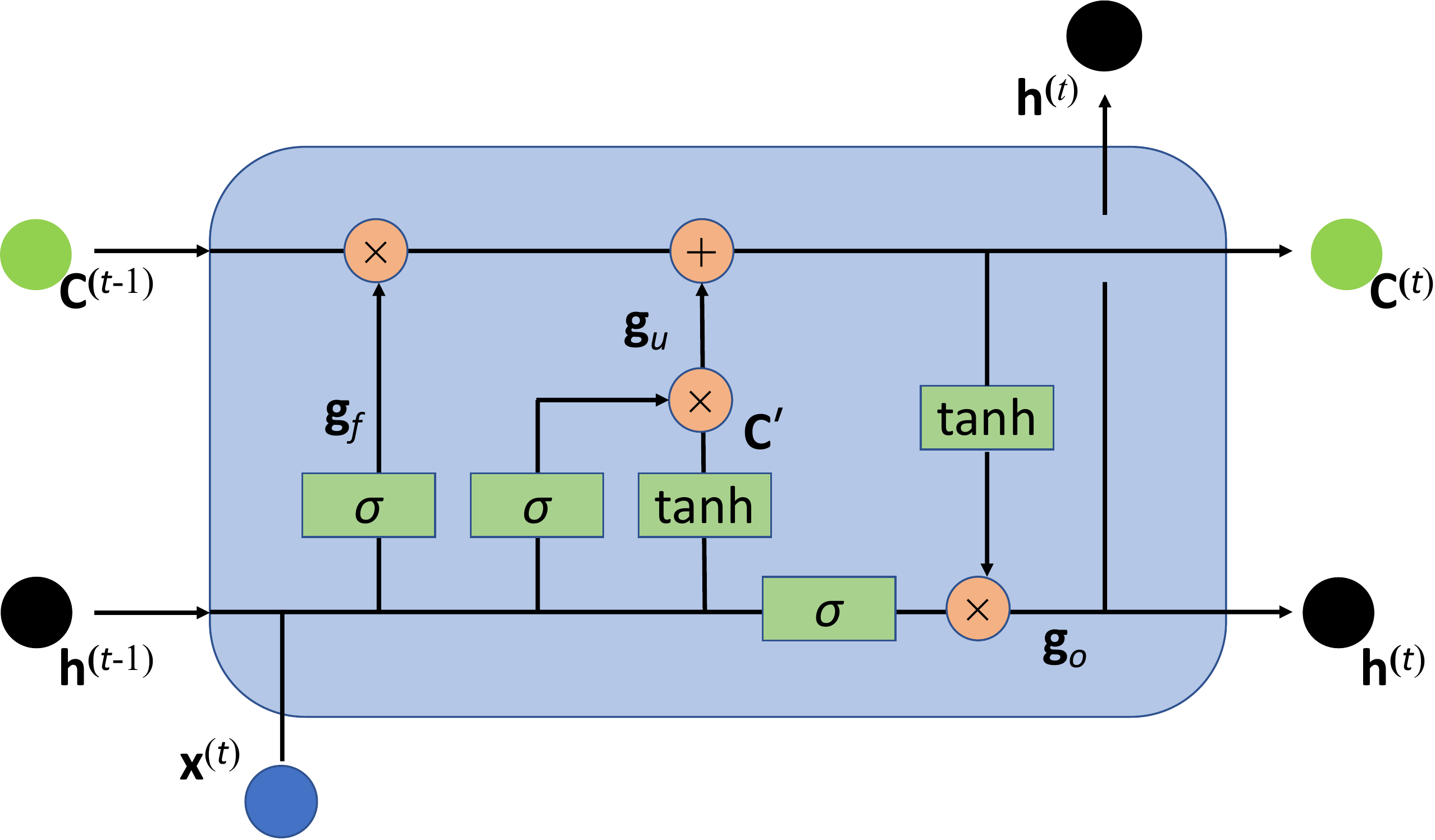}
\caption{An LSTM cell takes current-step input parameters ($\mathbf{x}^{(t)}$) and previous-step outputs (cell state $\mathbf{C}^{(t-1)}$ and hidden state $\mathbf{h}^{(t-1)}$) and operates through \emph{forget gate} ($\mathbf{g}_f$), \emph{update gate} ($\mathbf{g}_u$), and \emph{output gate} ($\mathbf{g}_o$).
A hidden state ($\mathbf{h}^t$) is produced for prediction tasks.
The current-step cell state ($\mathbf{C}^t$) and $\mathbf{h}^t$ are passed onto the next time step.
Operation symbols in circles represent point-wise operations.
}
\label{fig:LSTM}
\end{figure}

RNN performs well in various engineering applications, however, it suffers from \emph{long-term dependencies}.
Taking a language generation model as an example, the long-term dependencies refer to situations where the word to be predicted is far away in memory from the relevant information.
In theory, RNN is capable of handling the long-term dependencies problem.
However, in practice, RNN is unable to connect the information~\cite{Bengio1994}.
The long-short term memory (LSTM) algorithm (Fig.~\ref{fig:LSTM}) improves RNN by adding a \emph{forget gate} ($\boldsymbol{g}_f$), an \emph{update gate} ($\boldsymbol{g}_u$), and an \emph{output gate} ($\boldsymbol{g}_o$)~\cite{Hochreiter1997,Sak2014long}.
We mathematically describe LSTM updates through the following equations:
\begin{align}
    & \boldsymbol{g}_f = \sigma \left(\boldsymbol{w}_f \cdot [\boldsymbol{h}^{(t-1)}, \boldsymbol{x}^{t}] + \boldsymbol{b}_f\right), \\
    & \boldsymbol{g}_\text{u} = \sigma \left(\boldsymbol{w}_\text{u} \cdot [\boldsymbol{h}^{(t-1)}, \boldsymbol{x}^{(t)}] + \boldsymbol{b}_\text{u}\right), \\
    & \boldsymbol{C}^\prime = \tanh \left(\boldsymbol{w}_\text{C} \cdot [\boldsymbol{h}^{(t-1)}, \boldsymbol{x}^{(t)}] + \boldsymbol{b}_\text{C}\right), \\
    & \boldsymbol{C}^{(t)} = \boldsymbol{g}_f \cdot \boldsymbol{C}^{(t-1)} + \boldsymbol{g}_u \cdot \boldsymbol{C}^\prime, \\
    & \boldsymbol{g}_o = \sigma \left(\boldsymbol{w}_o \cdot [\boldsymbol{h}^{(t-1)}, \boldsymbol{x}^{(t)}] + \boldsymbol{b}_o\right), \\
    & \boldsymbol{h}^{(t)} = \boldsymbol{g}_o \cdot \text{tanh}(\boldsymbol{C}^{(t)}),
\end{align}
where $\boldsymbol{w}$ contains the weighting parameters, and the subscripts represent the corresponding variables.
\emph{Forget gate} uses a $\sigma$ activation function to scale previous-step hidden state $\boldsymbol{h}^{(t-1)}$ and current-step input $\boldsymbol{x}^{(t)}$ within the range of 0 and 1 multiplied by each number in the previous-step cell state $\boldsymbol{C}^{(t-1)}$.
Thus, a 1 represents ``keep this number", and a 0 represents ``forget this number".
Similarly, \emph{update gate} scales $\boldsymbol{h}^{(t-1)}$ and $\boldsymbol{x}^{(t)}$ within the range of 0 and 1 using a $\sigma$ function to decide how much we want to update the previous-step cell state $\boldsymbol{C}^{(t-1)}$ via a candidate cell state $\boldsymbol{C}^\prime$.
The candidate cell state $\boldsymbol{C}^\prime$ takes as inputs the $\boldsymbol{h}^{(t-1)}$ and $\boldsymbol{x}^{(t)}$ followed by a \texttt{tanh} activation function.
We achieve the current-step cell state $\boldsymbol{C}^{(t)}$ through $\boldsymbol{C}^{(t-1)}$ and $\boldsymbol{C}^\prime$, which are multiplied by the \emph{forget gate} and \emph{update gate} values, respectively.
On the one hand, $\boldsymbol{C}^{(t)}$ directly goes into the next-time-step LSTM cell with the values as a memory from previous steps.
One the other hand, we put $\boldsymbol{C}^{(t)}$ through \texttt{tanh}, which combines with the \emph{output gate} to guide the computation of $\boldsymbol{h}^{(t)}$.
\emph{Output gate} again uses $\sigma$ to scale $\boldsymbol{h}^{(t-1)}$ and $\boldsymbol{x}^{(t)}$ within the range of 0 and 1.
We obtain $\boldsymbol{h}^{(t)}$ by multiplying \texttt{tanh}($\boldsymbol{C}^{(t)}$) by \emph{output gate} values.

LSTM is able to model long-term sequence dependencies and is more robust to the problem of short memory than the vanilla RNN model due to the added gates.
However, the improved LSTM architecture increases the computing complexity and requires higher memory compared with vanilla RNN.
LSTM has been introduced into aerospace engineering mainly for predicting unsteady aerodynamics, which makes full use of the memory cells.
\citet{Wang2020MultivariateRN} showed promising multivariate LSTM predictive performance on scalar and distribution aerodynamic quantities in unsteady aerodynamics.

\subsubsection{Autoencoder}
\label{sec:mlAE}

\begin{figure}[h]
\centering
\includegraphics[width=0.8\linewidth]{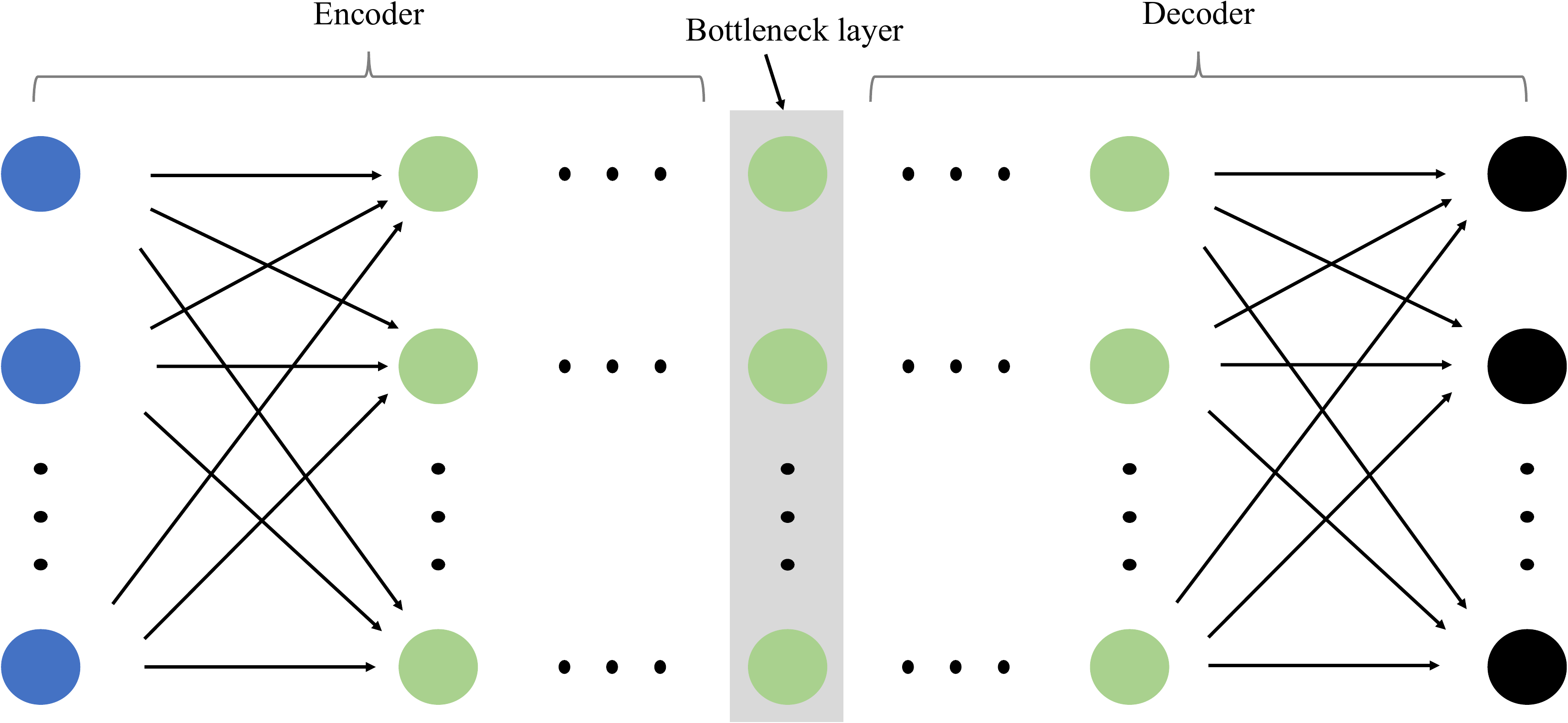}
\caption{An autoencoder consists of encoder and decoder networks connected through the bottleneck layer.
Input-layer features are automatically extracted into the bottleneck layer  {(which typically has a much lower dimension than the input dimension)} in an unsupervised way by minimizing the difference between the output layer (black) and the input layer (blue).
}
\label{fig:Autoencoder}
\end{figure}

An autoencoder is an ANN designed to learn the low-dimensional representation of unlabeled high-dimensional data.
We use as the low-dimensional representation a neural network layer that has fewer neurons than the input layer.
This ``narrowed" layer is also known as the bottleneck layer.
In sum, the bottleneck layer with few neurons is imposed in the autoencoder network architecture to force a compressed representation of the input.
The autoencoder is divided into an encoder functioning as a recognition model (from the input layer to the bottleneck layer) and a decoder functioning as a generative model (from the bottleneck layer to the output layer) (Fig.~\ref{fig:Autoencoder}).
The encoder can be a fully connected or convolutional network, which extracts low-dimensional features from the high-dimensional input.
The decoder converts the low-dimensional features to high-dimensional output, which can be done by fully connected or transposed convolutional layers.
In contrast to a convolutional layer used for downsampling (input dimension is larger than output dimension), transposed convolution is an upsampling strategy (output dimension is larger than input dimension).
Transposed convolution~\cite{Dumoulin2018guide} is similar to convolution except that we first preprocess inputs via padding operation (a process of adding layers of zeros to increase the height and width of an input image).
Then we apply convolution to the preprocessed inputs through which the output dimension is larger than the input dimension. 
The autoencoder is trained by minimizing the reconstruction loss $\mathcal{L}(\hat{\boldsymbol x}, \boldsymbol x)$ of output $\hat{\boldsymbol x}$ compared with the input $\boldsymbol x$.

\begin{figure}[!h]
\centering
\includegraphics[width=0.8\linewidth]{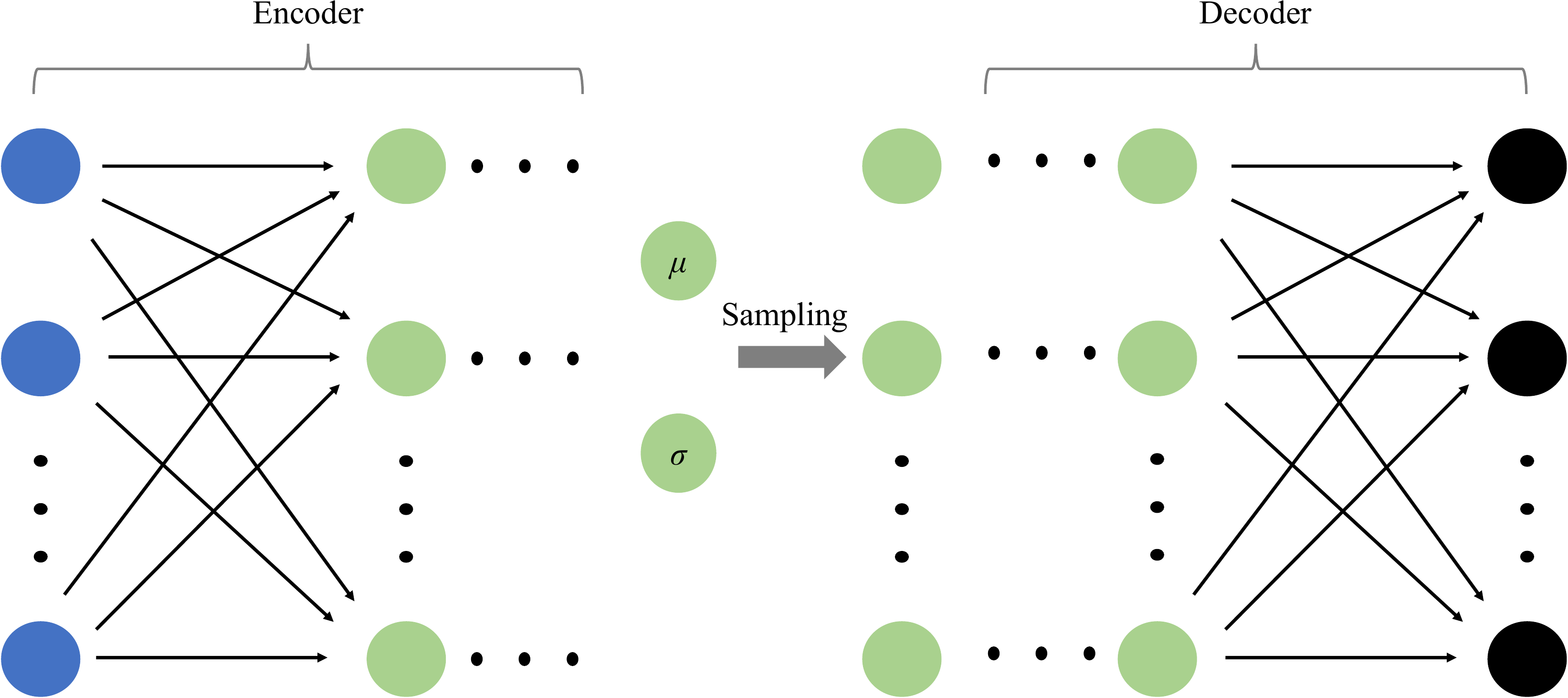}
\caption{VAE has a similar structure as autoencoder; however, VAE does not use a bottleneck layer.
Instead, the encoder of VAE produces the mean ($\mu$) and standard deviation ($\sigma$) of latent variables.
Gaussian distribution with the predicted $\mu$ and $\sigma$ is used to generate samples as the input layer of the decoder.
Once trained, a VAE model can generate similar data or shapes as input data using the Gaussian distribution with $\mu$ and $\sigma$.
}
\label{fig:VAE}
\end{figure}

A variational autoencoder (VAE) introduces isotropic Gaussian priors on the latent variables to obtain low-dimensional representations with disentangled latent variables~\cite{kingma2013a}.
This is realized by adding a layer containing a mean and a standard deviation for each latent variable (Fig.~\ref{fig:VAE}).
The loss function for VAE consists of two terms.
The first term penalizes the reconstruction error.
The second term encourages the learned distribution to be similar to the true prior distribution, assuming a unit Gaussian distribution.
A detailed introduction on VAE can be found~\cite{Kingma2019a}.

Despite the similar architecture to a basic autoencoder, a VAE belongs to the family of variational Bayesian methods~\cite{Tran2021practical}.
Specifically, an autoencoder is an unsupervised technique for dimensionality reduction by reading, compressing, and then recreating the original input.
In contrast, a VAE model assumes the source data belongs to implicit probability distribution and attempts to infer the distribution parameters, through which a VAE model generates new data related to the source data.
We have noticed some applications of autoencoder and VAE models in ASO.
\citet{Rios2021} applied PCA, kernel PCA, and autoencoder for a compact representation of 3D vehicle shape design space to advance the optimization performance.
They showed that they could modify the geometries more locally with the autoencoder than with the remaining methods and verified that the autoencoder representation improved the optimization performance.
\citet{Wang2021vae} extracted the VAE variables to represent flow fields of supercritical airfoils and applied an ANN model to capture the mapping between airfoil shapes and the VAE variables.

\subsubsection{Generative Adversarial Networks}
\label{sec:mlGAN}

\begin{figure}[h]
\centering
\includegraphics[width=0.6\linewidth]{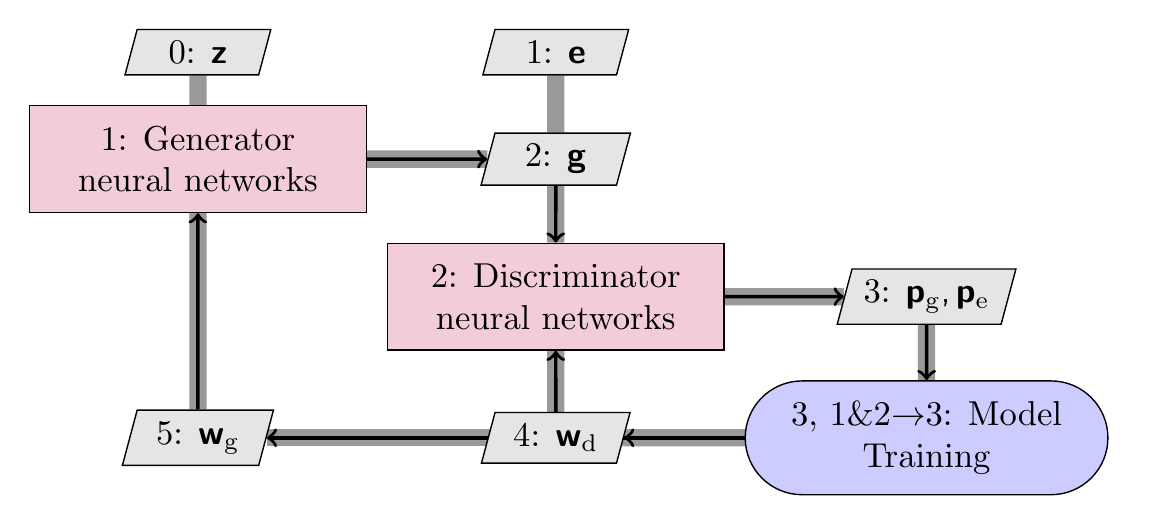}
\caption{A GAN incorporates the discriminator networks to compete with the generator networks, improving on other generative models, such as VAE.
The generator networks produce generated shapes ($\mathbf{g}$) corresponding to random variables ($\mathbf{z}$), which typically follow user-assigned uniform distributions.
The discriminator networks aim at differentiating the existing data ($\mathbf{e}$) and $\mathbf{g}$.
Training discriminator adjusts the weights ($\mathbf{w}_d$) and leads the probability of $\mathbf{g}$ ($\mathbf{p}_g$) being real towards $\mathbf{0}$ and the probability of $\mathbf{e}$ ($\mathbf{p}_e$) towards $\mathbf{1}$.
In contrast, training generator adjust the weights ($\mathbf{w}_g$) and leads the $\mathbf{p}_g$ towards $\mathbf{1}$.
At the end of the training, the generator generates new shapes similar to the existing data.
Ideally, $\mathbf{p}_g$ and $\mathbf{p}_e$ are around $\mathbf{0.5}$ for all $\mathbf{z}$ samples but it is difficult to achieve.
}
\label{fig:GAN}
\end{figure}

GANs consist of two ``adversarial'' models: a generator and a discriminator, competing against each other (Fig.~\ref{fig:GAN})~\cite{Goodfellow2014}.
The generator and discriminator can be any type of neural networks.
The goal of training the generator is to generate data that maintains similar patterns and properties as the provided training data set.
In contrast, the discriminator distinguishes between generated data and training data.

Generally, the generator reads random variables following a uniform distribution.
The discriminator reads data (both generated and training data) and tells the probability that these data are taken from the training data set.
Usually, the generated data and training data are labeled as  $\boldsymbol{0}$ and $\boldsymbol{1}$, respectively.
This process is mathematically formulated as a minimax problem~\cite{Goodfellow2014},
\begin{equation}
    \underset{\boldsymbol{w}_\text{g}}{\text{min}} \ \underset{\boldsymbol{w}_d}{\text{max}} \ V(\boldsymbol{w}_\text{g}, \boldsymbol{w}_d)= \E_{\boldsymbol{z}\sim P_\text{z}}[\text{log}(\boldsymbol{1} - \boldsymbol{p}_\text{g})] + \E_{\boldsymbol{e} \sim P_{\text{data}}}[\log(\boldsymbol{p}_\text{e})],
    \label{eqn:gan}
\end{equation}
where $\boldsymbol{e}$ is sampled from the training data distribution $P_{\text{data}}$.

\begin{figure}[!h]
\centering
\includegraphics[width=0.6\linewidth]{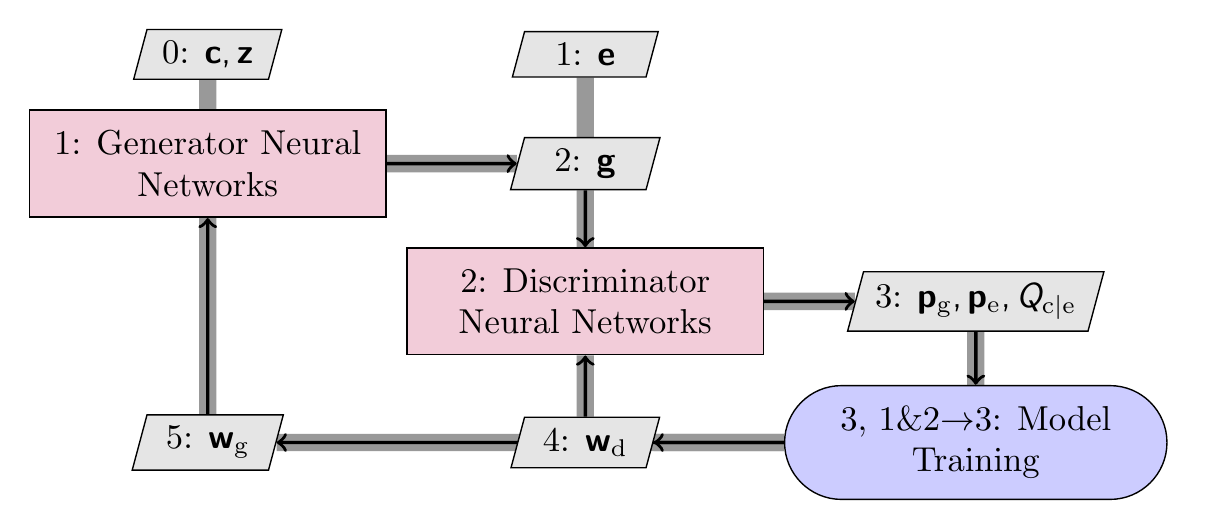}
\caption{InfoGAN uses two sets of input variables, the latent variables ($\mathbf{c}$) and noise variables ($\mathbf{z}$).
The discriminator outputs not only the probabilities ($\mathbf{p}_g$ and $\mathbf{p}_e$) but also the approximate distributions ($Q_{\mathbf{c}|\mathbf{e}}$) of $\mathbf{c}$ given the existing data samples ($\mathbf{e}$).
$Q_{\mathbf{c}|\mathbf{e}}$ approximating the real $\mathbf{c}$ distribution enables the convenient form of lower bounds of mutual information, which is called variational mutual information.
The training process is the same as GAN (Fig.~\ref{fig:GAN}), except that the loss function also considers the mutual information to be maximized between $\mathbf{c}$ and $\mathbf{e}$.
More mathematical details can be found in \citet{chen2016_info}.
}
\label{fig:InfoGAN}
\end{figure}

\begin{figure}[h]
\begin{center}
    \subfigure[] {\includegraphics[width=.49\textwidth]{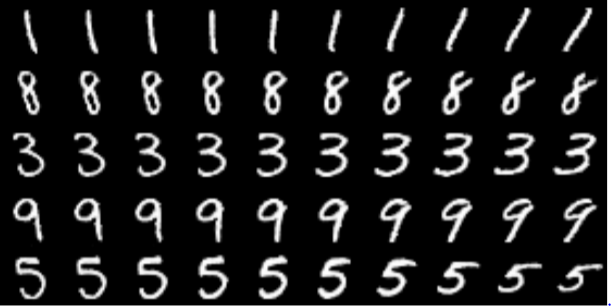}\label{}}
    \hspace{0.3mm}
    \subfigure[] {\includegraphics[
    width=.49\textwidth]{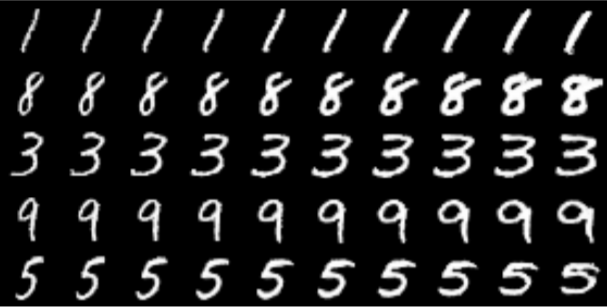}\label{}}
\end{center}
\vspace{-0.5cm}
\caption{\citet{chen2016_info} applied the InfoGAN model to MNIST data using 10 categorical latent variables and two uniform latent variables.
After training, the digit type is automatically captured by each categorical variable.
The two uniform variables of control are rotation (a) and width (b).
}
\label{fig:InfoGAN_example}
\end{figure}

Mutual information-based GAN (InfoGAN)~\cite{chen2016_info} extends GAN to extract structural data features, such as rotation and width of MNIST digits (Fig.~\ref{fig:InfoGAN}).
In probability theory and information theory, the mutual information between two random variables is a measure of the mutual dependence between the two variables.
In simple words, mutual information is a metric that quantifies how much one random variable ($X$) tells us about another ($Y$),  {which can be quantified through entropy as}
\begin{equation}
    I(X; Y) = E(X) - E(X|Y) = E(Y) - E(Y|X).
\end{equation}
InfoGAN model introduces latent variables into GAN while keeping the original GAN variables, also called noise variables.
Maximizing the mutual information between latent variables and generated data during training automatically assigns structural data features to latent variables based on the pre-defined probabilistic distributions  {(Fig.~\ref{fig:InfoGAN_example})}.
 {Therefore, we reformulate the original minimax objective function of GAN (Eq.~\eqref{eqn:gan}) as}
\begin{equation}
    \underset{\boldsymbol{w}_\text{g}}{\text{min}} \ \underset{\boldsymbol{w}_d}{\text{max}} \ V(\boldsymbol{w}_\text{g}, \boldsymbol{w}_d) - I(\boldsymbol{c}; \boldsymbol{g}),
\end{equation}
 {where $\boldsymbol{c}$ is the vector of latent variables and $\boldsymbol{g}$ is the vector of generated shapes.
Because the mutual information between $\boldsymbol{c}$ and $\boldsymbol{g}$ is challenging to obtain, the lower bound of mutual information is derived as~\cite{chen2016_info}}
\begin{equation}
        L(\boldsymbol{w}_\text{g}, \boldsymbol{w}_d) = \E_{\boldsymbol{g} \sim P_{G(\boldsymbol{c}, \boldsymbol{z})}} [\E_{\boldsymbol{c}^\prime \sim P(\boldsymbol{c}|\boldsymbol{e})} \left[ \log Q(\boldsymbol{c}^\prime| \boldsymbol{e}) \right]] + \text{const},
    \end{equation}
 {where $P_{G(\boldsymbol{c}, \boldsymbol{z})}$ is the distribution of generated airfoil shapes.
Thus, the training objective function of InfoGAN model becomes}
\begin{equation}
    \underset{\boldsymbol{w}_\text{g}}{\text{min}} \ \underset{\boldsymbol{w}_d}{\text{max}} \ V(\boldsymbol{w}_\text{g}, \boldsymbol{w}_d) - L(\boldsymbol{w}_\text{g}, \boldsymbol{w}_d).
\end{equation}

Compared with a VAE model, a GAN generates new data samples more correlated to a source data set because of incorporating the discriminator networks.
A GAN model does not introduce any deterministic bias, while variational methods introduce bias because they optimize a lower bound on the log-likelihood.
A GAN model generates a sample via one pass through the model, rather than an unknown number of Markov chain iterations compared with Boltzmann machines.

However, a GAN model is hard to train to reach the perfect equilibrium state, \emph{i.e.}, the discriminator-predicted probability of generated samples being realistic is 0.5 within the whole GAN variable space~\cite{Borji2021pros}.
Another commonly known problem of GAN is mode collapse.
Mode collapse happens when the discriminator gets stuck in a local minimum and the generator starts producing the same output (or a small set of outputs) over and over again.
Thus, a GAN model cannot generate a wide variety of outputs.
A common solution to mode collapse is using Wasserstein loss function, which helps the discriminator get rid of the vanishing gradient issue~\cite{arjovsky2017wasserstein}.
Therefore, the discriminator trained through Wasserstein loss function does not get stuck in local minima and learns to reject the outputs that the generator stabilizes on.
However, \citet{arjovsky2017wasserstein} also mentioned that ``weight clipping is a terrible way to enforce a Lipschitz constraint'', which leads to unstable training or slow convergence.

GAN and InfoGAN models combined with curve fitting methods (B\'ezier or B-spline curves) have been introduced to ASO to parameterize airfoil and wing sections.
After training, a curve fitting-based GAN model not only generates smooth and realistic airfoils but also shortens the optimization iterations with almost no penalty on the optimal performance~\cite{chen2018_bezier,chen2019_bezier}.

\subsubsection{Self-organizing Maps}
\label{sec:mlSOM}

\begin{figure}[h]
\centering
\includegraphics[width=0.7\linewidth]{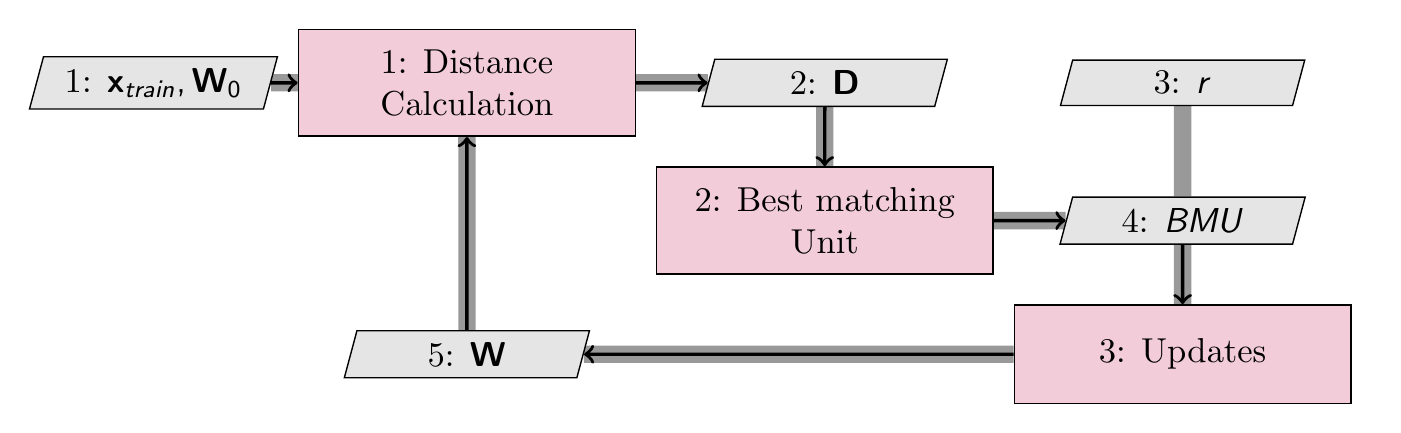}
\caption{A SOM is trained through competitive learning instead of error-correction learning, such as backpropagation with gradient descent.
A training sample ($\mathbf{x}_\text{train}$) is randomly selected, and then the Euclidean distance ($\mathbf{D}$) between $\mathbf{x}_\text{train}$ and weights of each SOM output node is calculated (the weights are initialized as ($\mathbf{W}_0$)).
The node with the shortest distance is called the best matching unit (BMU), around which a circle with a pre-set radius ($r$) is drawn.
BMU and its neighbor nodes need to be updated through the products of the difference between $\mathbf{x}_\text{train}$ and BMU weights and scaling parameters.
The scaling parameters ensure that the farther the neighbor nodes are, the smaller the change.
The training stops when the iteration limit is reached.
}
\label{fig:SOM}
\end{figure}

A SOM  {(Fig.~\ref{fig:SOM})} is unsupervised learning intended to find low-dimensional representations that preserve the topology of the high-dimensional input space~\cite{Kohonen1982a}.
Preserving topology means that neighbor observations in an input space should be projected as neighbor nodes in a mapping space~\cite{Phuoc2007}.
This topology preservation feature makes SOM outstanding for data analysis and data visualization because we can see more clearly in the mapping space the structure (such as clusters) hidden in the high-dimensional data~\cite{Uriarte2008TopologyPI}.
Unlike other ANN models, SOM is composed of single-layer structured neurons, and all the neurons are connected to the input.
Each neuron represents a coordinate of the low-dimensional map such as $(i,j)$, which computes the Euclidean distance between neurons to preserve the high-dimensional topology.
The parameters of the neurons are competitively updated to ensure that similar high-dimensional data are mapped closer.
Generative topographic mapping (GTM) is a probabilistic formulation of the SOM, which models the distribution of high-dimensional data by defining a density model in a low-dimensional space~\cite{Bishop1998a}.
Compared to SOM, the convergence of GTM (to local optima) is guaranteed by the EM algorithm~\cite{Bishop1998a} similar as the one used for GMM (Sec.~\ref{sec:mlGMM}).

We summarize several main advantages and disadvantages of a SOM model as follows.
Data can be easily interpreted and understood with the help of SOM.
SOM can deal with several types of classification problems while providing a valuable and intelligent summary from the data in the meantime.
However, SOM does not perform well on categorical data and even worse for mixed types of data.
The model preparation time is slow, meaning that SOM is hard to train.
SOM models have been used to visualize tradeoffs of Pareto solutions in the multi-objective function space in ASO.
\citet{Obayashi2003som} used a constructed SOM model to generate clusters of design variables, which indicated the roles of design variables for design improvements and tradeoffs.

\subsubsection{Physics-Informed Neural Networks}
\label{sec:mlPINN}

\citet{Raissi2019} developed PINNs as ``neural networks that are trained to solve supervised learning tasks while respecting any given laws of physics described by general nonlinear partial differential equations''~\cite{raissi2017physics1,raissi2017physics2}.
The key to PINNs' success is incorporating partial differential equations of interest into the loss function for neural network training.
In the training of PINNs, the residual of underlying partial differential equations and the residual derivatives are to be evaluated.
The loss function for PINN training reads:
\begin{equation}
    \min_{\boldsymbol{\theta}} \left( \sum_{i=1}^{N_u} (\hat{y}_{u,i}(\boldsymbol{\theta}) - y_{u,i})^2 + \sum_{i=1}^{N_f} (\hat{r}_{f,i}(\boldsymbol{\theta}))^2 \right),
\end{equation}
where the former term denotes the prediction error of physics, and the latter is a residual penalty of the prediction $\hat{r}$ (also named as ``physical loss") on collocation points.

PINNs are a novel way of constructing ML models by incorporating physical principles into ML for powerful performance instead of purely data-driven models.
PINNs lay the foundations for a new paradigm in modeling and computation that enriches deep learning with the longstanding developments in mathematical physics by naturally encoding any underlying physical laws as prior information.
However, the original PINN was set up under a fixed set of initial and boundary conditions, which makes it challenging for design optimizations whose boundary condition varies after each iteration.
In addition, PINNs had not been introduced to practical large-scale ASO problems yet.
Even so, ML researchers still express great interest in PINNs because they incorporate expert knowledge.
\citet{Mao2020pinn} showed satisfactory results on both forward and inverse problems using PINN-approximated Euler equations for high-speed aerodynamic flows.

\section{Machine Learning in Aerodynamic Shape Optimization}
\label{sec:aso}

The efficiency of aerodynamic shape design optimization is mainly affected by three aspects: the dimensionality of the geometric design space, the cost of the aerodynamic analysis, and the convergence rate of the optimization algorithm.
Recently, advanced ML models have shown the potential to efficiently parameterize the aerodynamic shape, accurately predict aerodynamic performance with a low computational cost, and innovate ASO workflows.
In this section, we review relevant applications in these fields with comments on their performance and contributions.

\subsection{Geometric Design Space}
\label{sec:asoGeo}

Fine control of the aerodynamic shape is desired in design optimization to improve the aerodynamic performance.
Conventional geometric parameterization methods usually introduce a large number of design variables to guarantee the optimal design is included within the design space.
However, such design spaces contain abnormal design candidates that cause convergence difficulties for the CFD solver and the optimization.
Thus, it is desirable to exclude abnormal shapes and reduce the dimensionality of the design space.
ML has been a powerful tool to accomplish this.

We introduce two effective approaches: modal parameterization methods to improve geometric efficiency (Sec.~\ref{sec:asoGeoMode}) and geometric filtering methods to shrink the design space (Sec.~\ref{sec:asoGeofiltering}).
 {
Modal parameterization and geometric filtering shrink the design space by coupling design variables and adding geometric constraints, respectively.
From the 2000s to the 2010s, modal parameterization has mainly been based on linear methods such as PCA, focusing on airfoil shapes.
Then, nonlinear modal parameterization is developed by applying deep learning models such as GAN, emphasizing the generation of realistic aerodynamic shapes.
The applications have been extended from two-dimensional airfoils to three-dimensional wings.
Similarly, before the 2010s, geometric filtering was based on engineering knowledge with few complexities.
With deep learning, geometric filtering models are now trained to detect complex abnormalities for detailed aerodynamic shape design optimization.
The relevant publications are listed in Table~\ref{tab_designspace}.
Both modal parameterization and geometric filtering are accomplished by either explicitly using knowledge from human designers or learning knowledge from historical designs.
Thus, a concern on whether the shrunk design space could include innovative aerodynamic shapes beyond previous experience is reasonable.
}

\begin{table}[!htbp]
\centering
\caption{Research efforts on the compact geometric design space for aerodynamic shape optimization}
\label{tab_designspace}
\begin{tabularx}{\textwidth}{
    >{\raggedright\arraybackslash}p{0.54\textwidth}
    >{\raggedright\arraybackslash}p{0.16\textwidth}
    >{\raggedleft\arraybackslash}p{0.22\textwidth} }
\hline
\hline
Description &  Application  & Reference  \\
\hline
\rowcolor{lightgray}
Modal parameterization via extracting orthogonal modes from supercritical airfoils & Airfoil & \citet{Robinson2001} \\
Modal parameterization using PCA and pre-optimized airfoils                & Airfoil & \citet{Toal2008} \\
\rowcolor{lightgray}
&  &  \citet{airfoilmode4}\\
\rowcolor{lightgray}
&  & \citet{Poole2015}\\
\rowcolor{lightgray}
Modal parameterization using PCA and UIUC airfoils & Airfoil & \citet{Masters2017}\\
\rowcolor{lightgray}
&  & \citet{Allen2018}\\
\rowcolor{lightgray}
&  & \citet{Li2019b}\\
 & & \citet{chen2019_bezier}\\
Modal parameterization using GAN and UIUC airfoils & Airfoil & \citet{Du2021a} \\
& & \citet{Wang2021}\\
\rowcolor{lightgray}
Modal parameterization using PCA and pre-optimized samples & Rotor blade & \citet{Duan2019a} \\
Modal parameterization using PCA and deep-learning-based optimal samples & Airfoil and wing & \citet{Li2021} \\
&  & \citet{Li2022a} \\
\rowcolor{lightgray}
Modal parameterization using GAN and probabilistic-grammar wing samples & Wing  & \citet{Chen2021}\\
& & \citet{Lukaczyk2014}\\
& & \citet{Namura2017}\\
Modal parameterization using ASM and random samples & Airfoil and wing  & \citet{Grey2018} \\
& & \citet{Li2019}\\
\rowcolor{lightgray}
Geometric filtering by using a low-fidelity model  &  Wing planform & \citet{Giunta1995} \\
Geometric filtering by involving engineering judgment of human designers & Nacelle & \citet{Sbester2006} \\
\rowcolor{lightgray}
Geometric filtering by engineering knowledge & 2D intake duct & \citet{Li2012}\\
Geometric filtering by adding deep-learning-based validity constraints & Airfoil and wing & \citet{Li2020a}\\
&  & \citet{Li2021d}\\
\hline
\hline
\end{tabularx}
\end{table}

\subsubsection{Modal Parameterization}
\label{sec:asoGeoMode}

Modal parameterization uses derived modes to control the aerodynamic shape.
The modes are derived from a series of samples as a global representation, which can be either linear or nonlinear.
This is different from conventional parameterization methods such as CST, Hicks--Henne bump functions, and FFD, where the design variables act as local deformations of the aerodynamic shape.
Modal parameterization has drawn increasing attention in aerodynamic shape design because of its effectiveness~\cite{Robinson2001,Toal2008,Ghoman2012,airfoilmode4,Poole2015,Li2019b,Li2021,Li2022a}.
For example, \citet{Robinson2001} derived orthogonal airfoil modes from a family of supercritical airfoils and demonstrated that only a few modes were required to approximate the aerodynamic behavior of the original airfoils.
Among different modal parameterization methods, PCA-based modal parameterization is a popular choice.
As explained in Sec.~\ref{sec:mlPCA}, PCA modes can be solved using SVD, so this parameterization is also called SVD or POD modes in some references~\cite{Toal2008,airfoilmode4,Poole2015}.

The University of Illinois at Urbana--Champaign (UIUC) Airfoil Coordinates Database~\footnote{\url{https://m-selig.ae.illinois.edu/ads/coord_database.html}} contributes a lot to the development of modal parameterization.
The database is open-accessible and contains approximately 1,600 airfoils (from low Reynolds number airfoils to transonic supercritical airfoils), covering a wide range of applications for unmanned aerial vehicles (UAVs), wind turbines, and jet transports.
These airfoils contain valuable historical design knowledge for ML.
Many thanks to the UIUC applied aerodynamics group led by Michael Selig and all contributors to the database.

PCA-based airfoil mode shapes have been widely used in airfoil shape design optimization~\cite{Poole2015,Li2019b} and wing shape design optimization~\cite{Allen2018} (where airfoil modes control local wing sectional shapes).
Most airfoil mode shapes were directly extracted from airfoil coordinates (full-airfoil modes), and each mode contains both camber and thickness information~\cite{Poole2015}.
\citet{Li2019b} proposed to extract modes of airfoil camber and thickness lines (camber-thickness modes).
Camber-thickness modes and full-airfoil modes have similar geometry representation efficiency.
However,  camber-thickness modes are more intuitive and practical for airfoil shape optimization.
\citet{Masters2017} showed the outstanding efficiency of PCA-based airfoil modes by comparing seven different airfoil parameterization schemes (CST, Hicks--Henne bumps, B-splines, radial basis function domain elements, B\'ezier surfaces, and the parameterized sections method) in geometric shape recovery of over 2000 airfoils.

\citet{Berguin2014,Berguin2015} applied PCA to aerodynamic derivatives solved by the adjoint method  {to find low-dimensional modes that have significant influences on the aerodynamic coefficients of interest}.
This approach is equivalent to the active subspace method (ASM)~\citep{Constantine2014}.
Unlike PCA modes, ASM modes are characterized by maximal output variation, which finds the most influential directions in the design space with respect to the objective functions such as lift and drag.
This provides an approach to dimensionality reduction.

ASM has been applied in the aerodynamic shape optimization of aircraft wings \citep{Lukaczyk2014,Grey2018,Li2019} and the aerodynamic shape analysis for a vehicle design \citep{Othmer2016}.
\citet{Lukaczyk2014} evaluated the ASM modes of the ONERA M6 wing for lift and drag.
They showed that the lift and drag were mainly dependent on one and two ASM modes, respectively.
\citet{Li2019} coupled ASM modes with a surrogate-based optimization framework and showed that a more practical wing design with a lower drag can be obtained with a reduced computational cost compared to surrogate-based optimization with FFD control points.
To avoid solving aerodynamic derivatives, \citet{Namura2017} proposed a kriging-based method to estimate the ASM modes, which is especially useful in black-box problems.
Nevertheless, its application in the transonic flow regime with hundreds of design variables may be problematic because the aerodynamic functions are much more nonlinear, making it difficult to accurately estimate the gradient using surrogate models.
\citet{Tripathy2016} proposed a gradient-free way to find the active subspace basis by coupling the dimensionality reduction procedure with the Gaussian process (kriging), where the active subspace is found in a Stiefel manifold using a two-step maximum likelihood optimization procedure.
Similarly, \citet{Rajaram2020b} constrained the basis to be on the Grassmann manifold, which finds the active subspace rather than the specific bases.

\begin{figure}[h]
\centering
\includegraphics[width=\linewidth]{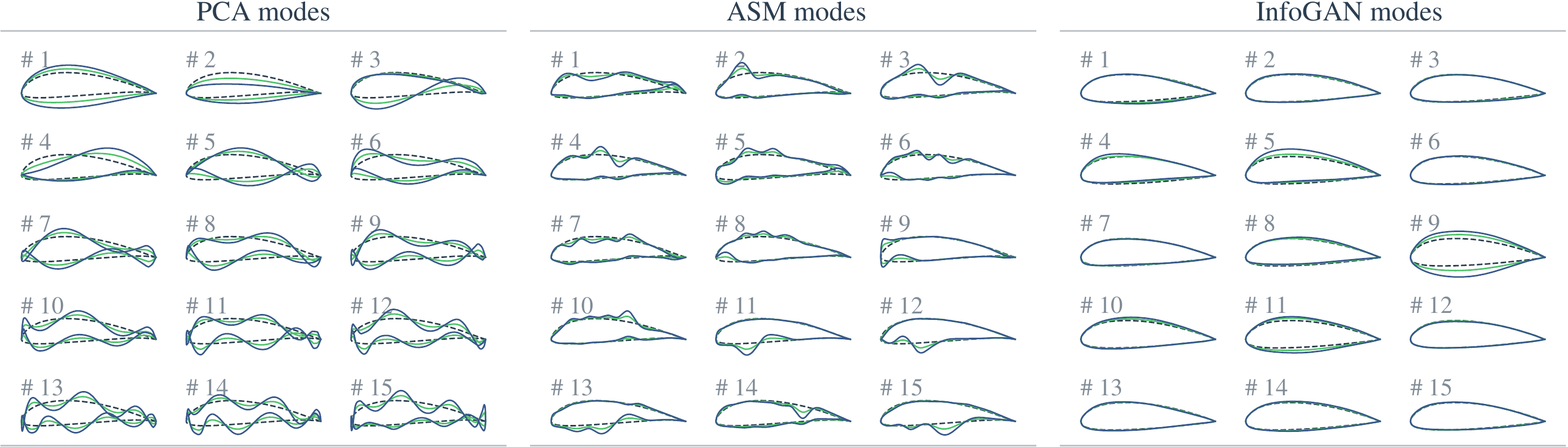}
\caption{Airfoil shape deformation by different types of modes.
The dash lines are the baseline airfoil.
The green and blue lines are deformed airfoil shapes by the mode with coefficients of one and two units, respectively.}
\label{fg_modeshapes}
\end{figure}

In addition to the linear modes, nonlinear modes have also been introduced.
The latent-variables-governed generator of InfoGAN (Sec.~\ref{sec:mlGAN}) provides a realistic approach to nonlinear modal parameterization.
GAN~\cite{Goodfellow2014} is an efficient way to produce realistic high-dimensional data by learning the distributions of low-dimensional latent variables.
\citet{Li2020a} showed that using CNN and a normalization process in GAN can generate smooth airfoil shapes without pre-parameterization and that by using the Wasserstein distance in GAN (WGAN) (see Sec.~\ref{sec:mlGAN}), the underlying distribution of the training airfoils can be captured accurately, and the possibility of mode collapse is reduced~\cite{Li2021d}.
InfoGAN~\cite{chen2016_info} extends GAN to represent data patterns (such as airfoil thickness) via latent variables by maximizing the mutual information between latent variables and the generator distribution.
\citet{chen2019_bezier} introduced InfoGAN parameterization into airfoil design and showed the shape control advantage of B\'ezier curve-based GAN model over GAN and original InfoGAN models in terms of smoothness.
\citet{Du2021a} generalized the work of \citet{chen2019_bezier} by incorporating a B-spline-curve layer into the GAN model for better shape control.
They completed rigorous parametric studies to guarantee the fitting accuracy of the existing UIUC airfoil database to include potential optimal design shapes.

Other methods, such as VAE (Sec.~\ref{sec:mlAE}), GTM (Sec.~\ref{sec:mlSOM}), and nonlinear manifolds (Sec.~\ref{sec:mlNml}), can also extract low-dimensional latent spaces for nonlinear modal parameterization.
\citet{Wang2021} used VAE and VAEGAN to extract airfoil modes and compared their performance with PCA modes.
\citet{Viswanath2011} introduced GTM in surrogate-based optimization of airfoil and wing shapes~\cite{Viswanath2014}, {and \citet{Doronina2022a} used the Grassmannian manifold to parameterize the sectional shape of wind turbine blades}.
 {In the industry, a Rolls-Royce funded study~\cite{Pongetti2021a} showed that using an autoencoder could reduce the geometric parameters of a turbomachinery blade from 100 to 15.} 
Among these methods, the GAN series is recommended because it can generate realistic shapes benefiting from the discriminator network (Sec.~\ref{sec:mlGAN}).
 {Nonlinear modes potentially produce a more efficient parameterization than linear modes, requiring fewer design variables for ASO. 
Future work is recommended to demonstrate this potential in benchmark ASO problems~\cite{Lyu2015b,Yu2018a}.
}

We show three types of modes (PCA, ASM, InfoGAN) obtained from the UIUC airfoil database in Fig.~\ref{fg_modeshapes}.
The PCA modes are derived using the airfoil coordinates, as explained by \citet{Li2019b};
The ASM modes are evaluated for the lift coefficient at $M=0.75$ based on the FFD parametrization described by \citet{Li2019}.
The InfoGAN modes are defined as the latent variables of the B-spline-based InfoGAN model, as explained by \citet{Du2021a}.
PCA and ASM modes are linear, so the shape deformation shows a linear superposition effect when the mode coefficient is doubled.
InfoGAN modes are nonlinear since nonlinear activation functions are used in the network.
Changing the latent variables of InfoGAN  {exerts} a global deformation effect of the geometric characteristics, such as thickness and camber,  {while the noisy variables complement the parameterization by adding minor thickness changes}~\cite{Du2021a}.

In evolutionary optimization or surrogate-based optimization, dimensionality reduction approaches such as PCA can be directly used to improve efficiency.
For example, \citet{Asouti2016} used PCA to guide the evolutionary optimization in the preliminary design of a supersonic business jet, aeroelastic design of a wind turbine blade, and aerodynamic design of airfoils.
\citet{Tao2020} constructed and integrated a PCA-DBN-based surrogate model into an improved particle swarm optimization framework to realize robust aerodynamic design optimizations.
\citet{Kapsouli2016} applied KPCA in the evolutionary optimization of an isolated airfoil and the DrivAer car and demonstrated higher efficiency than using the original PCA algorithm.

A database of realistic aerodynamic shapes is desired for modal parameterization, no matter which model parameterization method is used.
Mode derivation without a realistic database may show non-smooth mode shapes.
Also, these modes may lead to design space that is not restrictive enough and contains regions corresponding to abnormal airfoil shapes.
Nevertheless, unlike the situation in airfoils with the UIUC database,  publicly available information on the geometry of three-dimensional components like wings is nearly absent.
Thus, most modal parameterization  {studies are limited to} two-dimensional airfoil shape design.

\begin{figure}[H]
\centering
\includegraphics[width=\linewidth]{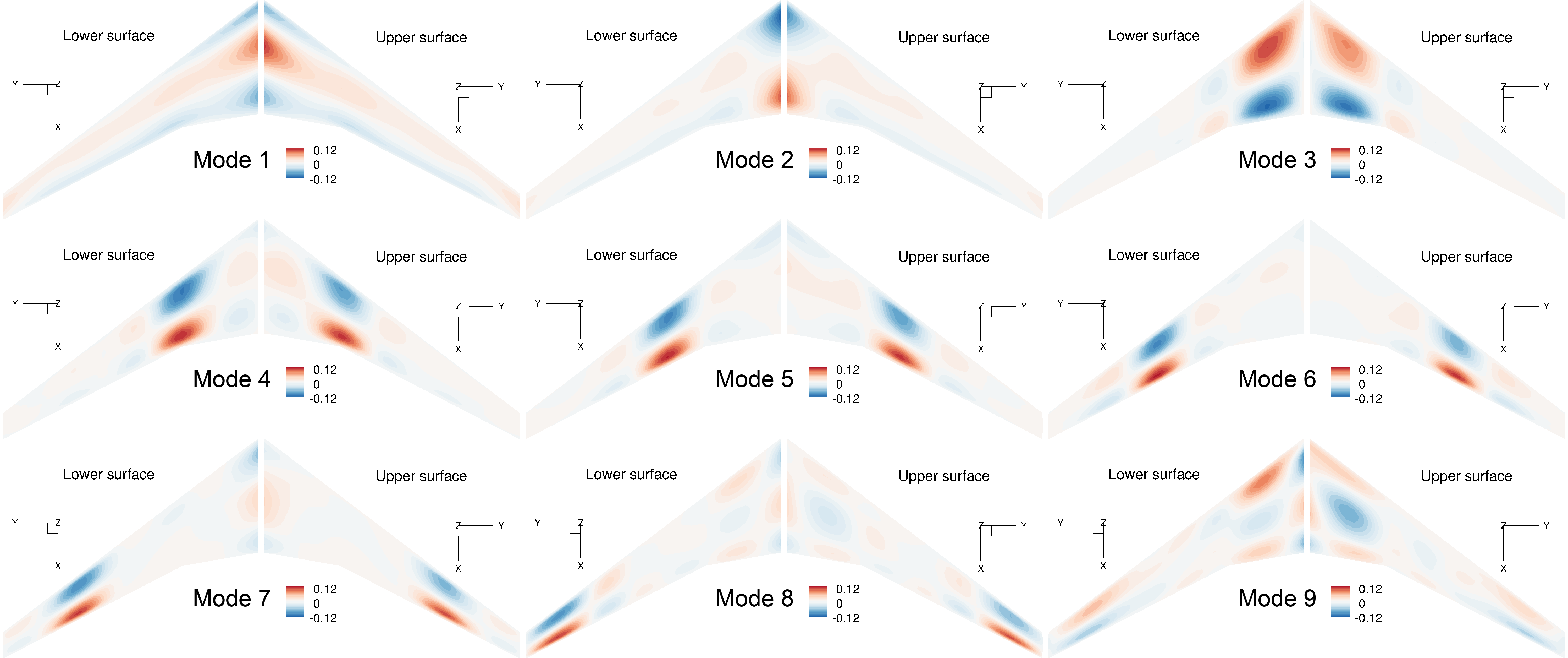}
\caption{PCA mode shapes of the Common Research Model (CRM) wing extracted from wing samples generated by the deep-learning-based optimal sampling method~\cite{Li2021,Li2021b}.}
\label{fg_wingmodes}
\end{figure}

One can perform a pre-optimization to collect suitable three-dimensional shapes with the desirable performance as did by~\citet{Toal2008} in airfoil design.
For example, \citet{Duan2019a} used this strategy to extract eight PCA modes of the NASA Rotor 37 blade from 1,734 samples searched in the pre-optimization, and then another 0.25\% gain on the adiabatic efficiency was obtained by performing optimization in the 8-D modal design space.
 {Despite its feasibility}, it is time-consuming to use pre-optimization to build the shape database.
Also, the database may not include enough shape diversity to derive effective modal parameterization for the optimal shape because the good designs obtained in pre-optimization may be far from the optimum.

To obtain effective modal parameterization of three-dimensional shapes, \citet{Chen2021} presented a \emph{probabilistic grammar} approach to creating a database for GAN to learn from.
The probabilistic grammar introduces a set of rules to ensure valid wings are sampled.
For example, the variation of the chord from root to tip has to follow a monotonic variation.
The smoothness of generated shapes is guaranteed by incorporating an FFD layer into the model.
Results showed that the FFD-GAN leads to faster convergence in wing shape optimization problems than direct FFD and B-spline parameterization.
Nevertheless, due to the arbitrariness in generating the wing data set, the geometric design space governed by these FFD-GAN modes can still be too large for specific aerodynamic design problems, which is usually the case in industrial applications.

\begin{figure}[h]
\centering
\includegraphics[width=\linewidth]{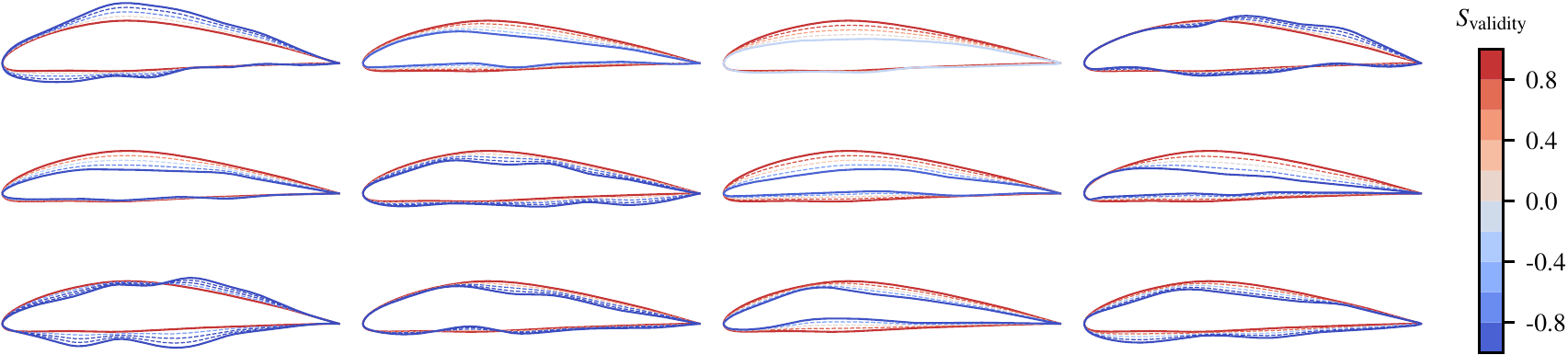}
\caption{Geometric validity scores of different airfoil shapes evaluated by the CNN discriminative model.
    A smooth monotonic decaying of the scores from realistic shapes to abnormal shapes implies that the validity constraint is suitable for gradient-based optimization.}
\label{fg_score_decay}
\end{figure}

To address the issue, \citet{Li2021} proposed a deep-learning-based optimal sampling method to generate realistic samples of three-dimensional wing-like shapes considering specified geometric constraints.
The method combines an optimization process with deep-learning-based geometric validity models~\cite{Li2020a,Li2021d} to find realistic wing shapes that satisfy the specified geometric constraints.
First, an airfoil GAN model is used to generate sectional shapes for the wings as the starting points of the optimization.
Then, the optimization is performed by minimizing the displacement to maintain the sparsity of wing samples.
Geometric validity constraints based on a CNN-based discriminative model~\cite{Li2020a,Li2021d} are enforced to ensure realistic sample shapes  {(Fig.~\ref{fg_score_decay})}.
The details of the geometric validity constraint are explained in Sec.~\ref{sec:asoGeofiltering}.
These wing samples fill a specific small region of the high-dimensional geometric design space where the aerodynamic shapes are geometrically feasible to the optimization problem.
Mode shapes derived from these samples produce an efficient wing shape parameterization~\cite{Li2021,Li2021b} that reduces dimensionality without excluding the potential optimum.
The global mode shapes of the CRM wing extracted using this method are shown in Fig.~\ref{fg_wingmodes}.
These modes lead to a compact design space, and therefore a wing shape design problem can be efficiently solved without relying on the adjoint solver or gradient-based optimization~\cite{Li2021}.
This feature is effective in solving low-Reynolds-number aerodynamic shape optimization~\cite{Li2022a} where laminar-to-turbulent transition dominates the aerodynamic performance and leads to discontinuous aerodynamic functions.

\subsubsection{Geometric Filtering}
\label{sec:asoGeofiltering}

Geometric design space with constant design variable bounds may involve regions with abnormal aerodynamic shapes because design variables  {could} have unquantifiable inner relationships~\cite{Sbester2006,Li2019b}.
These regions correspond to abnormal aerodynamic shapes that do not contribute to the design but may seriously affect optimization efficiency.
For a low-dimensional interpretable space (for example the flight conditions), the physical region could be defined by flight mechanics specialists~\cite{Zhang2020}, but it is intractable to manually determine the desired regions for high-dimensional geometric design space.
Design space filtering has been proposed in ASO as an efficient way to exclude the abnormal regions, and similar attempts have been applied to other design problems~\cite{Garriga2019,Xiong2019}.
This approach does not reduce the number of design variables; instead, it shrinks the design space by defining a constraint function to evaluate the abnormality of samples.

Geometric filtering constraints differ from geometric constraints on distance~\cite{Brelje2020a}, thickness, volume, or curvature~\cite{Kedward2020,Bons2020b}.
Geometric filtering constraints are not governed by known equations, and therefore data-based models are usually used to formulate them.
For example, \citet{Giunta1995} reduced the computational cost in an HSCT aircraft wing design optimization by first performing a design space exploration using a low-fidelity model and then excluding ``nonsense'' regions to have a much-reduced domain for high-fidelity optimization.
\citet{Li2012} constructed an engineering-knowledge-based filtering model to improve the optimization efficiency in a two-dimensional intake duct design.
The filtering model was trained using support vector regression based on penalties on the intake position, the intake duct curvature, interference area, and standard deviation of the total pressure distribution.
\citet{Sbester2006} used an RBF network as a means of capturing the geometric and engineering judgment of human designers to remove abnormal CAD models (such as those with excessive snaking and sharp transitions in the spline forward of the fan face) in nacelle design.
To exclude abnormal shapes in modal parameterization of airfoils, \citet{Li2019b} used the inverse distance weighting method to interpolate functions of higher-order airfoil modes with respect to the dominant modes, which capture the distribution feature of higher-order modes in existing airfoils.
Geometric filtering constraints were defined by adding and subtracting a margin to the interpolated functions.
With these constraints, abnormal regions were excluded from the design space, and \citet{Li2019b,Bouhlel2020} consequently trained accurate data-based airfoil analysis models.

Using deep-learning models, \citet{Li2020a} developed a generic validity model to detect the geometric abnormality of airfoils and wing sections.
The model was trained by a large number (over 10,000) of labeled realistic airfoils and abnormal airfoils.
GAN was used to generate realistic synthetic airfoils after learning the underlying distribution of existing airfoils.
Abnormal training airfoils were generated by using Latin hypercube sampling (LHS) to randomly perturb the FFD control points of realistic airfoils.
The realistic and abnormal airfoils were labeled as $+1$ and $-1$.
Then, a discriminative CNN was trained to quickly evaluate the geometric validity of arbitrary airfoils or wing sectional shapes.
The validity scores exhibit a monotonic and smooth decay from a realistic shape to an abnormal shape, which indicates that it is suitable for gradient-based optimization.
Using the validity model in geometric filtering shrinks the design space in airfoil and wing shape optimization~\cite{Li2020a,Li2021}.
This contributes to the efficiency of surrogate-based optimization as it improves the accuracy of surrogate models by preventing infilling abnormal shapes in training data.
It also helps to develop accurate and generic data-based aerodynamic models for interactive design optimization (Sec.~\ref{sec:ASOInterOpt}).
\citet{Li2021d} performed 216 airfoil design optimization and several wing design optimization of a conventional wing-body-tail configuration and a blended-wing-body configuration.
They found that using the geometric filtering with a lower bound of $\sim 0.7$ does not exclude innovative aerodynamic shapes that maximize cruise efficiency  {in the investigated transonic regime}.
The results support the usefulness of applying generic deep-learning-based geometric filtering in aerodynamic shape optimization.

 {
Current research on geometric filtering is mostly based on empirical rules or past designs, which is fundamentally limited.
Future work on forming generic geometric filtering with no prior knowledge (for example, using adaptive high-fidelity simulation samples) is of great interest.
}

\subsection{Aerodynamic Evaluation}
\label{sec:ASOAeroEvalu}

The evaluation of aerodynamic objective and constraint functions (and their derivatives with respect to the design variables) takes most of the computational cost in CFD-based aerodynamic shape design optimization.
Thus, there is always a demand to reduce the cost of aerodynamic evaluations.
ML  {could provide} an effective way to accomplish this.

 {Early research (before the 2010s) on modeling aerodynamic coefficients is based on traditional surrogate models such as the polynomial response surface method and kriging, which have moderate tunable parameters and are usually trained by small volumes of data in a relatively small space.
Then, the development of surrogate models (such as ME and ANN) makes it possible to predict aerodynamic functions in larger spaces by using large volumes of training data, which is helpful to realize interactive ASO.
Powerful CNN models show potential in modeling aerodynamic coefficients with respect to the geometry coordinates without using any shape parameterization, but its benefits over conventional approaches are to be demonstrated.
Due to the inherent interpolation feature, these models may not ensure accuracy in extrapolations.
The possibility of extrapolations increases with the dimensionality of the design space.
Thus, the curse of dimensionality is still a fundamental difficulty in modeling aerodynamic coefficients for ASO.
The construction of an accurate aerodynamic prediction model is accomplished by shrinking the design space and adding more training data to reduce the possibility of extrapolatory predictions inside the design space. 
Future research emphasizing the extrapolation ability is of great value for high-dimensional ASO problems.}

 {
Because of the large number of flow field variables, flow field modeling usually starts with a dimensionality reduction.
Early research (in the 2000s) was based on linear methods such as PCA. Later studies (in the 2010s) used nonlinear manifolds and reported better prediction accuracy.
Recently, deep learning has contributed to new nonlinear dimensionality reduction models such as CNN.
The training of deep-learning-based dimensionality reduction models can be coupled with the training of the prediction network, which is an additional advantage compared with previous methods.
Besides, physics-informed approaches such as PINNs enable the prediction models to obey the underlying physics and improve prediction accuracy.
PINNs have extrapolation ability, which could address the curse of dimensionality in ASO problems.
Clear demonstrations of extrapolation in benchmark ASO cases are recommended in future work.
}

 {
Numerical simulation acceleration has not drawn as much attention as flow prediction in the interdisciplinary community of fluid dynamics and ML.
However, it is of great value in solving high-fidelity ASO problems.
It has been shown that using advanced flow prediction methods such as CNN could accelerate the simulation.
Other contributions such as data-driven spatial discretizations and closures, despite not directly accelerating the original simulation problem, help achieve a higher fidelity with the same computational cost.
These efforts can be directly used in high-fidelity CFD-based optimization. 
}

 {
Imposing off-design aerodynamic constraints is critical in ASO, the evaluation of which is time-consuming due to the high computational cost of unsteady flow simulations.
Direct modeling of the constraints with respect to geometric parameters suffers from the curse of dimensionality as in modeling aerodynamic coefficients at the cruise condition, and herein it is more computationally costly.
This calls for advanced ML solutions.
A promising approach is constructing physics-based models as an alternative to the constraints.
Early efforts were on constructing empirical formulas using simple ML models such as linear regression.
A typical application is Korn's equation to estimate the drag divergence Mach number of airfoils.
Recently, data-driven models with higher complexities have been used to handle high-dimensional physical information and achieved higher accuracy by using large volumes of training data.
Awareness of physics in these models brings in good generalization.
More research is recommended in this field to better balance the training cost and prediction accuracy.
}

\subsubsection{Aerodynamic Coefficient Modeling}
\label{sec:ASOAeroCoef}

\begin{table}[!htbp]
\centering
\caption{Approximate models of aerodynamic coefficients used in ASO.}
\label{tab_coeffmodel}
\begin{tabularx}{\textwidth}{
    >{\raggedright\arraybackslash}p{0.35\textwidth}
    >{\raggedright\arraybackslash}p{0.27\textwidth}
    >{\raggedleft\arraybackslash}p{0.3\textwidth}}
\hline
\hline
Model  & Aerodynamic shape   & References  \\
\hline
\rowcolor{lightgray}
  & Wing planform &  \citet{Giunta1995} \\
\rowcolor{lightgray}
Polynomial response surface  & 2D diffuser   &  \citet{Madsen2000} \\
\rowcolor{lightgray}
  & Airfoil &  \citet{Ahn2001} \\
\rowcolor{lightgray}
  & Car &  \citet{Sun2014}\\
Support vector regression   & Airfoil      &  \citet{Perez2016} \\
                            & Wing      &  \citet{Andre2018} \\
\rowcolor{lightgray}
  & Nacelle & \citet{Song2007a} \\
\rowcolor{lightgray}
  & Airfoil & \citet{Han2013} \\
\rowcolor{lightgray}
  & Wing & \citet{Han2017} \\
\rowcolor{lightgray}
Kriging  & Wing & \citet{han2018aerodynamic} \\
\rowcolor{lightgray}
  & Wing & \citet{Han2018a} \\
\rowcolor{lightgray}
  & Wind-turbine blade & \citet{Xu2020a} \\
\rowcolor{lightgray}
  & Propeller & \citet{Mourousias2021} \\
  &         Wing  & \citet{Liem2015}  \\
ME&         Airfoil  & \citet{Li2019b}  \\
  &         Aircraft  & \citet{Zhang2020}  \\
\rowcolor{lightgray}
  &         Airfoil  &  \citet{Du2019a} \\
\rowcolor{lightgray}
  &         Wing  &   \citet{Palar2017pck} \\
\rowcolor{lightgray}
PCE &         Blended-Wing-Body  & \citet{Zuhal2021}  \\
\rowcolor{lightgray}
  &         Airfoil  & \citet{Lin2020pck}  \\
\rowcolor{lightgray}
  &         Airfoil  & \citet{Nagawkar2020}  \\
  & Aircraft  & \citet{Secco2017} \\
  & Airfoil  & \citet{Bouhlel2020} \\
  & Airfoil  & \citet{Du2020} \\
  ANN  & Wing  & \citet{Li2021b} \\
  & Wing  & \citet{Barnhart2021} \\
  & UAV  & \citet{Karali2021} \\
  & Nacelle  & \citet{Yao2021} \\
  & Train & \citet{Zhang2021c} \\
\rowcolor{lightgray}
CNN &  Airfoil & \citet{Zhang2018b}  \\
\rowcolor{lightgray}
  & Airfoil & \citet{Yu2020idcnn} \\
\hline
\hline
\end{tabularx}
\end{table}

Although CFD analysis provides rich information, such as all flow variables in the computational domain, ASO is typically based on a few aerodynamic performance metrics, such as lift and drag coefficients.
Thus, modeling the aerodynamic coefficients is of great interest in aerodynamic design.
Such modeling is usually accomplished by fitting a simple prediction function between the shape design variables and the aerodynamic coefficients using training data generated by high-fidelity aerodynamic analyses.
This prediction model is called the surrogate model or metamodel  {(see Sec.~\ref{sec:mlSurro})}.
As shown in Table~\ref{tab_coeffmodel}, commonly-used surrogate models include the polynomial response surface method, kriging, and neural networks~\cite{Yondo2018,AndrsPrez2021a,Barnhart2021}.
As reviewed by \citet{Viana2014a}, the polynomial response surface method attracted the most attention at the beginning of computer-aided aerodynamic design in the 1990s.
Then, kriging~\cite{Krige1951} became popular in aerodynamic shape design due to its strong fitting ability~\cite{Simpson2001}, and the capability of providing the predictive confidence interval, which enabled the development of the efficient global optimization method (EGO)~\cite{Jones1998a}.
Variations of kriging, such as co-kriging and hierarchical kriging~\cite{han2012hierarchical}, were developed to take advantage of multi-fidelity simulations.

One issue in modeling high-dimensional problems using traditional surrogates is that training the model is time-consuming.
To reduce the cost, \citet{Bouhlel2016a,Bouhlel2016b} proposed to train the hyperparameters of kriging in several directions evaluated by the partial least squares method.
They achieved significant computational gains in numerical problems with up to 100 dimensions.
Nevertheless, the accuracy still decreases with increased dimensionality (the curse of dimensionality).

An alternative approach for addressing the time-consuming surrogate training process is using the sparse basis-based PCE surrogate (Sec.~\ref{sec:mlSurro}).
 {The training of PCE is a linear regression process, which can be solved by OLS or LARS.}
 {We can also select lower-order PCE through hyperbolic truncation while maintaining sufficient predictive accuracy.}
 {Moreover, an early stop strategy based on LARS makes the training process even more efficient}~\cite{Sudret2008}.
\citet{Du2019a} completed robust airfoil design in the transonic regime using LARS-based PCE and utility theory and revealed an outstanding robust design performance compared with the commonly used weighted sum method.
\citet{Palar2017pck} developed the PC-Kriging-based EGO and demonstrated that the proposed PC-Kriging-based EGO converged faster than standard EGO in a real-world subsonic wing design problem and those cases exhibiting a polynomial-trend landscape.
\citet{Zuhal2021} extended the original PC-Kriging to gradient-enhanced PC-Kriging.
They followed the same principle as GEK and stated that gradient-enhanced PC-Kriging improved performance and robustness compared with GEK and gradient-enhanced PCE for airfoil shape, blended-wing-body shape, and wing aerostructural design.
\citet{Lin2020pck} derived an analytical statistical moment estimation method for PC-Kriging and conducted a robust airfoil design, showing that the improved PC-Kriging was more efficient than Monte Carlo-based statistical estimation.
This work further merges the advantages of the predictive confidence interval by kriging and analytical observation statistics by PCE, which makes PC-Kriging more competitive for uncertainty quantification applications.
\citet{Nagawkar2020} showed an overall better aerodynamic coefficient prediction advantage of the PC-Cokriging algorithm developed by \citet{Du2020pcco} over PCE, kriging, PC-Kring, and cokriging in airfoil robust design.
The PC-Cokriging algorithm showed better performance on a borehole case~\cite{Harper1983} and a non-destructive testing case.
However, the accuracy difference in aerodynamic coefficient prediction is narrowed down, especially when more high-fidelity data are available.
The correlation between low-fidelity and high-fidelity simulation models can be a factor, and we should leave more discoveries for further investigations.

Another issue of traditional surrogates is that it is not suitable for handling a large volume of training data.
Although the training data in aerodynamic design is typically from expensive experiments and hence is typically limited in their size,
there is such a demand to deal with a large number of training data in the construction of a generic and accurate aerodynamic analysis model~\cite{Li2019b,Li2021b} for interactive design (see Sec.~\ref{sec:ASOInterOpt}).

An intuitive way to address this shortcoming is to use a mixture of experts (ME)~\cite{Masoudnia2012a,Bettebghor2011}.
ME is based on a ``divide-and-conquer'' approach, in which the design space is divided into local domains, and each domain is handled by one expert.
The construction of a ME consists of three steps:
1) Divide the design space into local domains (a process can be done using an unsupervised clustering algorithm).
2) Compute cluster posterior probability to evaluate the mixing proportions.
3) Train local experts and combine them using the mixing proportions.
\citet{Liem2015} derived a new ME approach based on ``divide-and-conquer'' principle using GMM  {(see Sec.~\ref{sec:mlGMM})} and showed that the proposed algorithm was advantageous in surrogate-based mission analysis of conventional and unconventional aircraft configurations.
\citet{Zhang2020} used ME to construct multi-fidelity surrogate modeling for handling qualities analysis.
\citet{Li2019b} used $k$-means-based ME to handle more than 100,000 training data in the construction of a generic airfoil aerodynamic analysis model.
One potential drawback of ME is that the model is likely not smooth at the boundaries of different experts.
\citet{Hwang2018b} developed new surrogate modeling techniques, regularized minimal-energy tensor-product splines (RMTS), and gradient-enhanced RMTS, which scale well with the number of training points (up to the order of $10^5$) thanks to the use of tensor-product splines.
 {A shortcoming of RMTS is that it cannot handle high-dimensional problems efficiently.} 

A promising approach to handling a large number of training data is using a neural network as the surrogate model, and it contributes much to realizing interactive design optimization (see details in Sec.~\ref{sec:ASOInterOpt}).
\citet{Secco2017} used an ANN to model the lift and drag coefficients of a wing-fuselage aircraft configuration with various wing planforms, airfoil shapes, and flight conditions.
Using about 100,000 training data generated by a full-potential code with computation of viscous effects, the ANN exhibited average absolute errors of 0.004 and 0.0005 for lift and drag coefficient predictions, respectively.
\citet{Bouhlel2020} adopted ANN to model the lift, drag, and moment coefficients of arbitrary airfoils and found that the ANN model outperformed the mixture of kriging models~\cite{Li2019b}.
Using the ANN model in the optimization yielded similar airfoil shapes to the designs obtained by high-fidelity CFD-based optimization.
The differences were 0.01 and 0.12 drag counts in the subsonic and transonic regimes, respectively.
\citet{Du2020, Du2021a} developed generic airfoil analysis models based on ANN for fast predictions on aerodynamic drag, lift, and pitching moment coefficients, and good optimal designs were achieved compared to CFD-based optimization (see  Sec.~\ref{sec:ASOInterOpt}).
\citet{Du2020} also used RNN models to predict the pressure coefficient distributions of baseline and optimal designs for richer design insights.
\citet{Li2021b} constructed an ANN-based aerodynamic analysis model (trained using 135,108 wing samples with different aerodynamic shapes, flight speeds, and flight altitudes) for wing shape design to achieve the interactive design.
In the verification on 47,967 unseen wing samples, the mean relative errors of drag, lift, and pitching moment compared with the high-fidelity CFD are all within 0.4\%.
\citet{Karali2021} trained an ANN model to predict lift, drag, and pitching moment coefficients of small UAVs with respect to 22 input parameters (21 geometric parameters and the angle of attack).
They used a training data set composed of 94,500 UAVs evaluated by a nonlinear lifting line method; most predictions matched the lifting line method well (the mean absolute error is 12.4 drag counts in the validation dataset).

 {
Random forests can also manage a large number of training data and are less complex than ANN (see Sec.~\ref{sec:mlDTRF}).
For example, \citet{Kumar2019} used a random forest to estimate $C_L$, $C_D$, and $C_M$ in the quasi-stall regime of the DLR-ATTAS aircraft and showed better accuracy than an empirical formula with several tunable parameters.
In their work, the random forest took a four-dimensional input composed of the angle of attack, the elevator deflection, the pitch rate, and the relative airspeed.
Despite the success, it is a low-dimensional case, and the performance of random forests in high-dimensional ASO problems needs further studies.
Even for low-dimensional cases, its advantage in prediction accuracy compared with ANN has not been demonstrated~\cite{Dube2020}.}

 {
Because of a large number of tunable parameters, training ANN models can be sample-inefficient and get stuck in local minima owing to poor initial parameter selection.
DBN (Sec.~\ref{sec:mlSSL}) addresses these issues with a semi-supervised learning feature.
DBN is first trained in an unsupervised learning manner for prediction tasks using unlabeled data (pre-training) and further tuned in a supervised manner (fine-tuning).
This manner enables DBNs to be trained using small volumes of training data.}
 {\citet{Tao2019} showed the capability of DBN in the approximation of aerodynamic coefficients with respect to geometric parameters in ASO of airfoils and wings.
The numbers of geometric design variables in the two cases are 12 (airfoil) and 39 (wing), and the high-fidelity training data used are 1600 and 750 for airfoil optimization and wing optimization, respectively.
Nevertheless, no comparison studies with ANN or conventional surrogate models such as kriging were performed to show the benefits of the semi-supervised learning feature in DBN.
Further studies are of interest to demonstrate the potential advantage of DBN in aerodynamic coefficient modeling with small datasets.
}

Typically, a parameterization method is used to define the shape design variables, which are the inputs of aerodynamic surrogate models.
It is possible to construct aerodynamic models with respect to the geometry coordinates by exploiting the ability of neural networks (especially CNN) to extract geometric features.
\citet{Zhang2018b} used a CNN to predict the lift coefficients of airfoils with a variety of shapes in multiple free-stream Mach numbers, Reynolds numbers, and diverse angles of attack, where the airfoil shape coordinates were directly fed to CNN without using a shape parameterization.
\citet{Yu2020idcnn} developed an improved deep CNN approach by proposing the ``feature-enhance-image" airfoil image preprocessing strategy.
The ``feature-enhance-image" strategy consists of broadening the airfoil thickness by 10 times to make small changes on control variables more noticeable and augmenting one airfoil image into three with different brightness to enlarge the number of training and testing data samples.
Data augmentation~\cite{Mikolajczyk2018,Shorten2019} through image transformations, such as flipping, color modification, and rotation, reduces overfitting and attracts more and more attention in ML field.
Nevertheless, further studies are required to demonstrate the advantages of this direct modeling way over the more popular ``parameterization and modeling'' approach.

Although ANN has been applied to aerodynamic shape design optimization by directly coupling with an optimizer~\cite{Azabi2019,Zhang2021a}, Bayesian optimization or EGO could be more efficient, so the capability of providing the predictive confidence interval is desired.
The deep Gaussian process was developed to enable this function in ANN-based surrogate models~\cite{Damianou2013}.
\citet{Rajaram2020a} showed that the deep Gaussian process model was more accurate than the traditional Gaussian process model, albeit with a higher computational cost in the training procedure.
In the aerodynamic prediction of airfoils with 42 shape design variables, the deep Gaussian process model was reported to reduce the root mean square error by $2\% \sim 45\%$ and $2\% \sim 19\%$ for $C_l$ and $C_d$, respectively.
The training time of the deep Gaussian process model was approximately 400 seconds, which was a significant increase compared to that in training a Gaussian process model ($0.1\sim 1$ second).
In addition, a hybrid ANN-Gaussian-process model can also be used~\cite{Renganathan2021}.
However, most of the published work on deep Gaussian processes did not compare predictive performance directly with DNN models~\cite{Damianou2013,Rajaram2020a,Cutajar2018} and the training process is slowed down compared with the traditional Gaussian process model.
The concept of combining DNN's predictive power and the Gaussian process's predictive confidence is worth pursuing in future work.

In addition to modeling the forward functions between the aerodynamic shape and aerodynamic quantities, neural networks can also model the inverse relationships for inverse design~\cite{Sharma2021inverse}.
For example, \citet{Sun2015} used ANN to model the functions between the aerodynamic coefficients and aerodynamic shape parameters.
They performed optimization to obtain airfoil and wing shapes satisfying the desired aerodynamic performance.
\citet{OLeary-Roseberry2021} proposed to solve ASO problems by training a neural network surrogate model of the optimal wing shape with respect to design requirements (flight conditions and design constraints).
To alleviate the training cost (one training sample is one CFD-based wing design optimization), they proposed to project the high-dimensional output space to low-dimension bases and construct residual-based neural networks.
The adaptively trained projection-based networks model (layer-by-layer training) achieved $95\%$ relative accuracy using about $200$ training samples and outperformed the end-to-end training version and full-space model.

It is appealing to generate designs of interest by specifying the desired performances, such as the cruise lift-to-drag ratio and the maximal lift coefficient.
 {One potential issue of} inverse design  {is that it may lead} to ill-posed optimization problems since one performance may correspond to zero or multiple shapes.
Recently, conditional generative models have drawn vast attention in inverse design as a data-driven realization of Bayesian inversion~\cite{Arridge2019} that addresses the ill-posed issue~\cite{Chen2021a}.
We review the relevant papers in Sec.~\ref{sec:ASOGenOpt} with comments on the pros and cons.

\subsubsection{Flow Field Modeling}
\label{sec:ASOAeroFlow}

Modeling the flow field is useful in many scenarios such as inverse design using pressure distribution, multidisciplinary design optimization, and optimization result interpretation.
Flow field variables have a high dimension (proportional to the number of CFD mesh points).
The first step in modeling the flow field is usually performing dimensionality reduction~\cite{Lucia2004}; several commonly-used techniques are listed in Table~\ref{tab_flowmodel}.

Proper orthogonal decomposition (POD)~\cite{Sirovich1987,Willcox:2002:A} is a commonly used method (equivalent to PCA  {introduced in Sec.~\ref{sec:mlPCA}}) that is effective in reducing the dimension of aerodynamic flow fields of airfoils~\cite{Alonso2010,Bourguet2011}, wings~\cite{Thomas2003,Bryant2013,Li2018}, complete aircraft configurations~\cite{Lieu2006,Amsallem2008,Fossati2015} among others.
After reducing the dimensionality of the flow field variables, the flow field can be modeled by constructing a surrogate model of the reduced-order flow variables (such as the POD coefficients) with respect to the design variables.
This approach is called the non-intrusive reduced-order model.
The surrogate model can be kriging~\cite{Qiu2015}, RBF~\cite{Gunot2013}, or ANN~\cite{Wang2019a,Yu2019,Sun2020,Li2021b,Renganathan2020}.
Another approach, called the intrusive reduced-order model, uses projection-based reduced model~\cite{Benner2015,Collins2020a} to solve reduced-order flow variables; this approach usually requires more computational time and code implementation (such as using automatic differentiation to solve the partial derivatives~\cite{He2021b}).

\begin{table}[!htbp]
\centering
\caption{Commonly-used dimensionality reduction techniques for flow field modeling}
\label{tab_flowmodel}
\begin{tabularx}{\textwidth}{
    >{\raggedright\arraybackslash}p{0.25\textwidth}
    >{\raggedright\arraybackslash}p{0.35\textwidth}
    >{\raggedleft\arraybackslash}p{0.32\textwidth}}
\hline
\hline
Method  & Aerodynamic shape   & References  \\
\hline
\rowcolor{lightgray}
        & Airfoil     &  \citet{LeGresley2000} \\
\rowcolor{lightgray}
        & Airfoil     &  \citet{Alonso2010} \\
\rowcolor{lightgray}
        & Airfoil     &  \citet{Bourguet2011} \\
\rowcolor{lightgray}
        & Airfoil     &  \citet{Wang2019a} \\
\rowcolor{lightgray}
       & Airfoil     &  \citet{Renganathan2020} \\
\rowcolor{lightgray}
POD     & Wing        &  \citet{Thomas2003} \\
\rowcolor{lightgray}
        & Airfoil     &  \citet{Bryant2013} \\
\rowcolor{lightgray}
        & Aircraft     &  \citet{Lieu2006} \\
\rowcolor{lightgray}
        & Aircraft and helicopter rotor     &  \citet{Fossati2015} \\
\rowcolor{lightgray}
        & Car     &  \citet{Bertram2018} \\
        & Aircraft     &  \citet{Amsallem2008} \\
Manifold learning & Airfoil and wing     &  \citet{Franz2014} \\
        & Aircraft     &  \citet{Ripepi2018} \\
        & Airfoil     &  \citet{Decker2020} \\
\rowcolor{lightgray}
        & Cylinder, 2-D car, and sphere &  \citet{Guo2016} \\
\rowcolor{lightgray}
        & Cylinder &  \citet{Jin2018a} \\
\rowcolor{lightgray}
        & Airfoil &  \citet{bhatnagar2019} \\
\rowcolor{lightgray}
        Autoencoder  & Airfoil &  \citet{Sekar2019} \\
\rowcolor{lightgray}
        & Airfoil &  \citet{Thuerey2020} \\
\rowcolor{lightgray}
        & Cylinder and airfoil &  \citet{Eivazi2020} \\
\rowcolor{lightgray}
        & Airfoil &  \citet{Duru2020} \\
\rowcolor{lightgray}
        &  Airfoil & \citet{An2021} \\
        & Airfoil  & \citet{Chen2020a} \\
    GAN & Airfoil  & \citet{Wu2020b} \\
        & Airfoil  & \citet{Wang2021c} \\
\hline
\hline
\end{tabularx}
\end{table}

The underlying assumption in POD-based reduced-order models (ROM) is that the flow solution lies in a low-dimensional linear subspace.
This makes it challenging for these models to predict nonlinear flow phenomena such as shock waves.
For flows with slowly decaying Kolmogorov $n$-width~\cite{Pinkus2012} such as advection-dominated problems, the linear subspace dimensionality should be significantly increased to reach acceptable accuracy~\cite{Ohlberger2016}.
Efforts such as using multiple linear subspaces~\cite{Amsallem2012} and shifting the POD basis~\cite{Reiss2018} have been made to refine the linear basis.
\citet{Wang2021d} proposed an adaptive sampling strategy to automatically infill more training points close to the strongly nonlinear region of the flow regime and demonstrated the effectiveness of flow field prediction for an airfoil in the transonic regime.
\citet{Dupuis2018}  {used GMM to divide the transonic regime into subdomains and trained multiple local POD-based ROMs.
This strategy was shown to have an improvement in terms of prediction accuracy of the turbulent flow around the RAE2822 airfoil.}
These methods may require a substantial understanding of physics.
 {
The Lawrence Livermore National Laboratory has active research in ROM and developed an open-source library to facilitate POD-based ROMs, called libROM~\footnote{\url{https://www.librom.net/}}.
The library provides multiple SVD versions (to improve the data collection efficiency) and supports both global ROM and local ROMs.
One unique feature of libROM is a physics-informed sampling strategy to control the prediction error with a limited number of snapshots}~\cite{Choi2020a}.

Using nonlinear dimensionality reduction techniques in ROM can be advantageous.
Early research in this field was based on nonlinear manifold learning such as Isomap and LLE (introduced in Sec.~\ref{sec:mlNml}).
\citet{Franz2014} applied Isomap to improve shock prediction behavior in the transonic flow regime with the constant aerodynamic shape, and the Isomap-based model led to a rather sharp pressure gradient across the shock, which had better agreement with CFD than the POD-based model.
\citet{Ripepi2018} reported that the Isomap method was employed to predict surface pressure distributions of a transonic wing-body transport aircraft configuration in the DLR Digital-X project~\cite{Kroll2016}, and that the prediction was more accurate than POD-based methods.
\citet{Decker2020} also showed that using Isomap and LLE could lead to smaller prediction errors in regions near shock waves than POD-based models.

ANN, such as autoencoder~\cite{Lee2020,Eivazi2020} and GAN~\cite{Wu2020b}, provides an efficient way to find the underlying nonlinear manifold of flow filed variables for nonlinear dimensionality reduction.
\citet{Guo2016} used a convolutional autoencoder to provide a fast prediction of non-uniform steady laminar flow for two-dimensional or three-dimensional shapes.
\citet{bhatnagar2019} applied a convolutional autoencoder to predict the velocity and pressure field in unseen flow conditions and airfoil shapes. 
The model achieved a relative error of less than 10\% over the entire flow field.
\citet{Eivazi2020} showed that a nonlinear autoencoder network is more effective than POD in extracting low-dimensional flow features for the accurate prediction of unsteady velocity fields around a cylinder and an oscillating airfoil.
\citet{Thuerey2020} used a conventional U-Net (a particular case of autoencoder) to predict the velocity and pressure fields of different airfoils with various Reynolds numbers.
The mean relative pressure and velocity errors were less than 3\% in tests on unseen airfoils.
 {\citet{Tangsali2020} investigated the generalizability of a convolutional autoencoder in predicting aerodynamic flow field across various flow regimes and airfoil geometric variation.
They concluded that the autoencoder architecture allows the extraction of relevant geometric features from largely different geometries (geometric generalization), providing a better out-of-sample prediction accuracy than physics-based generalization (variation in flow regimes).
The geometric generalization feature is beneficial for aerodynamic shape design optimization.}
\citet{Duru2020} applied a conventional autoencoder to model the pressure field around airfoils at a relatively high subsonic Mach number ($M=0.7$).
Their model achieved 88\% accuracy for unseen airfoil shapes and showed promise in capturing shock waves accurately.

One can formulate the nonlinear functions between shape parameters and high-dimensional flow fields using conditional generative models.
For example, \citet{Chen2020a} used a conditional GAN to train a prediction model of the velocity field with respect to different shapes and various conditions ($\alpha \in [-22.5\degree, 22.5\degree]$ and $Re \in [5\times 10^5, 5\times 10^6]$).
In the prediction of 75 unseen UIUC airfoils, the mean relative error of $x$- and $y$-components of velocities in regions of interest were 8.13\% and 20.83\%, which were smaller than those in CCN-based autoencoder~\cite{bhatnagar2019} and U-net~\cite{Thuerey2020}.
\citet{Wu2020b} applied a conditional GAN with a CNN generator to predict transonic flow fields of supercritical airfoils whose shapes are parameterized using 14 Hicks--Henne bump functions.
The conditional generative model was trained by 500 airfoils sampled around the RAE2822 airfoil and then was shown to roughly predict the flow fields of unseen airfoils.
\citet{Wang2021c} used GAN to predict the pressure coefficient distribution of different airfoil shapes at various $M$, $Re$, and $\alpha$.

The nonlinear manifold for dimensionality reduction can be obtained together with the training of the prediction model in ANN.
This differs from the two-step modeling approach, which first performs dimensionality reduction and then models the reduced-order flow variables in conventional reduced-order models. 
The feature that fuses dimensionality reduction and prediction may help to improve the accuracy of ANN-based reduced-order models.
\citet{Kashefi2021}  {applied a PointNet architecture to learn the end-to-end mapping between arbitrary spatial positions and CFD quantities at these positions.
The network is designed to handle irregular geometries and unstructured grids.
The generalizability is demonstrated in the prediction of the velocity and pressure fields around multiple sectional shapes, including an airfoil.
}
One potential issue of directly predicting the flow field using ANN is that the smoothness of contour lines is not guaranteed.
\citet{An2021} introduced the structural similarity and multiscale structural similarity into aerodynamic flow-field prediction.
They investigated the effects of various weight pairs of multiscale structural similarity on the predictive accuracy and perceived visual quality.
The proposed method outperformed the root mean squared error-based loss function by more than 40\% in terms of visual quality (such as smoothness) with a cost of less than 6\% on root mean squared error.

Besides, the physics-informed loss function~\cite{Raissi2019} enables ANN to obey the known physical governing equations such as RANS, which has been shown to improve the model accuracy and generalization~\cite{Karniadakis2021a}.
 {\citet{Cai2021a} showed the ability of PINN to infer the full continuous three-dimensional velocity and pressure fields from experimental flow snapshots over an espresso cup.
\citet{Kissas2020} applied PINN models to de-noise and reconstruct clinical magnetic resonance imaging data of blood velocity.
\citet{Tay2020} investigated both a conventional ANN and a PINN in the prediction of flow past a converging-diverging nozzle and concluded that the PINN reduced the prediction error.
Many applications in the literature have shown that PINNs provide accurate and physically consistent predictions by using prior knowledge or constraints, even when extrapolating~\cite{Karniadakis2021a,Cai2021,Viana2021,Yang2021a}.
Nevertheless, there is still a lack of research demonstrating the advantage of PINNs over conventional approaches such as POD with kriging in benchmark aerodynamic shape optimization cases.
Embedding physical laws enables PINNs to have the potential of extrapolatory predictions.
This feature is of great value to ASO and may lead to an evolution of surrogate modeling architectures used in aerodynamic design.
More research in PINN-based models focusing on aerodynamic predictions considering the change of aerodynamic shapes is recommended to provide a clear demonstration of the potential.
}

Similar to the curse of dimensionality in modeling aerodynamic coefficients, it is challenging to model flow fields of various aerodynamic shapes in high-dimensional geometric design space.
\citet{Xiao2015} concluded that the computational cost of non-intrusive methods increases exponentially with the number of design variables.
Thus, most applications are restricted to the change of flow condition parameters or geometric parameters of two-dimensional shapes such as airfoils~\cite{Sekar2019,Thuerey2020,Eivazi2020,Duru2020} and car profiles~\cite{Guo2016}.
To effectively model the flow field of three-dimensional aerodynamic shapes,  {it would be helpful to shrink the geometric design space} by using modal parameterization and geometric filtering introduced in~\ref{sec:asoGeo}.

 {The truncated prediction models using a low-dimensional basis may not preserve global integral quantities, which would introduce additional errors when evaluating aerodynamic functions.
Thus, it should be careful to use flow field models in ASO, even though they can model the objective aerodynamic function.}
For example, in the shape design of an intake port to maximize the mass flow and the tumble (characterizing the mixture capacity of the air-fuel), \citet{Xiao2009} found that modeling the velocity field using POD and kriging was worse than directly modeling the integral quantities (mass flow and tumble).
 {This is a reason why most ASO research focuses on directly predicting aerodynamic coefficients rather than the flow field.
However, PINNs may help because of the inclusion of physics and the ability to extrapolate.
This motivates deeper research in this area.}

In addition to building surrogate models of the flow field with respect to design variables, constructing models of the adjoint variables with respect to state variables is of interest in aerodynamic shape design optimization, because with this kind of model, we can directly evaluate the aerodynamic derivatives after CFD simulations without the need to solve the adjoint equation.
\citet{Xu2020} presented a proof-of-concept of this idea by training an ANN-based surrogate model between the flow variables and the adjoint vector, and the model was verified in the drag minimization starting from the NACA~0012 airfoil.
Further research is recommended to show if this approach is generalizable to a broader range of ASO problems.

\subsubsection{Numerical Simulation Acceleration}
\label{sec:ASOAeroSimu}

In high-fidelity aerodynamic shape optimization, surrogate models (even those considered to have high accuracy) may not meet the required fidelity, so high-fidelity CFD simulations are still in demand.
ML techniques are also useful to accelerate high-fidelity CFD simulations.

A straightforward approach is performing CFD refinement starting with a guess given by the surrogate model of flow fields~\cite{TromeurDervout2006,Grinberg2011}.
For example, \citet{ObiolsSales2020} used a CNN to predict the primary fluid properties, including velocity, pressure, and eddy viscosity.
They accelerated the simulations by up to a factor of 1.9 to 7.4 in both steady laminar and turbulent flows on various geometries.
Instead of the initial guess, \citet{Andersson2016} adopted DMD in intermediate iterations to correct the solution closer to a steady-state condition and thus accelerated the convergence of steady flow simulations.
\citet{Liu2019a} proposed a mode multigrid method using DMD to truncate all the high-frequency parts in steady flow simulations and achieved speedup ratios of $3$ to $6$ while ensuring the computational accuracy, and \citet{Chen2020} used a similar strategy to accelerate the convergence of the steady adjoint equations.
\citet{Liu2018a} improved the efficiency of the implicit dual time scheme for unsteady flow simulations by providing proper initial solutions using DMD at each physical time step.

The computational cost of fluid simulations increases rapidly with grid resolution, and efforts have been made to reduce the grid size without losing simulation resolutions.
\citet{BarSinai2019a} proposed a data-driven discretization method to systematically derive discretizations based on spatial derivatives for continuous physical systems.
This allowed accurate time integration of nonlinear equations in one spatial dimension at resolutions 4 to 8 $\times$ coarser than standard finite-difference methods.
\citet{Zhuang2021a} applied the data-driven discretization in a two-dimensional turbulent flow and showed that using a grid with 4$\times$ coarser resolution can still maintain the simulation accuracy.
\citet{Kochkov2021a} further showed that the data-driven discretization can result in 40 to 80$\times$ computational speedups for both direct numerical simulation and large-eddy simulation of two-dimensional turbulent flows.

Turbulence modeling using ML techniques has attracted much interest~\cite{Duraisamy2019}, where ML is generally used to fit data-driven closures to classical turbulence models using high-fidelity simulation data.
These data-driven closures potentially make the simulation achieve higher fidelity than traditional turbulence models.
Although they are not designed to accelerate simulations, the feature may reduce the computational expense of solving ASO problems that require higher fidelity simulations.
\citet{Tracey2015a} demonstrated the ability of ML in turbulence modeling by training ANN to reproduce the behavior of the Spalart--Allmaras turbulence model in simulations of flat plate boundary layers and 3-D transonic wings.
\citet{Ling2016} used ANN to learn the Reynolds stress tensor from high fidelity simulation data (DNS and well-resolved LES) and presented a novel network architecture to embed Galilean invariance into the predicted anisotropy tensor, which was shown to have higher accuracy than a conventional network.
The tensor model improved the accuracy of RANS in simulations of duct flow corner vortices and the flow separation in a wavy wall.
\citet{Wang2017a} showed that ML using DNS data can predict the Reynolds stresses in the flow over periodic hills and then improve the accuracy of RANS.
\citet{Duraisamy2019} provide a review of data-driven turbulence modeling.
The applications of these methods are rare in ASO.
\citet{Zhang2021b} performed two-fidelity RANS-LES aerodynamic shape optimization of the periodic-hill geometry (one shape design variable) using a data-driven turbulence closure model.
Further research is expected to improve the simulation fidelity in the design optimization of industrial shapes such as aircraft and wind turbines.

Some parameters in CFD solvers such as the relaxation parameter are adjustable, and it has been shown that tuning them using ML yields faster convergence while retaining stability~\cite{Runchal2020}.
\citet{Discacciati2020} designed an ANN to estimate artificial viscosity in discontinuous Galerkin schemes and integrated it into a Runge--Kutta solver.
Their ANN-based artificial viscosity model guaranteed optimal convergence rates for smooth problems and accurately detected discontinuities, independently of the selected problem and parameters.

\subsubsection{Practical Constraint Formulation}
\label{sec:ASOAeroCon}

Practical constraints, such as the maximum lift coefficient $C_L^\text{max}$ and buffet onset, restrict the aircraft flight envelope and thus are essential in aerodynamic shape optimization.
Objective functions are usually computed at steady cruise flight conditions.
However, the constraints might involve unsteady simulations and therefore be much more time-consuming.

\begin{figure}[h]
\centering
\includegraphics[width=0.75\linewidth]{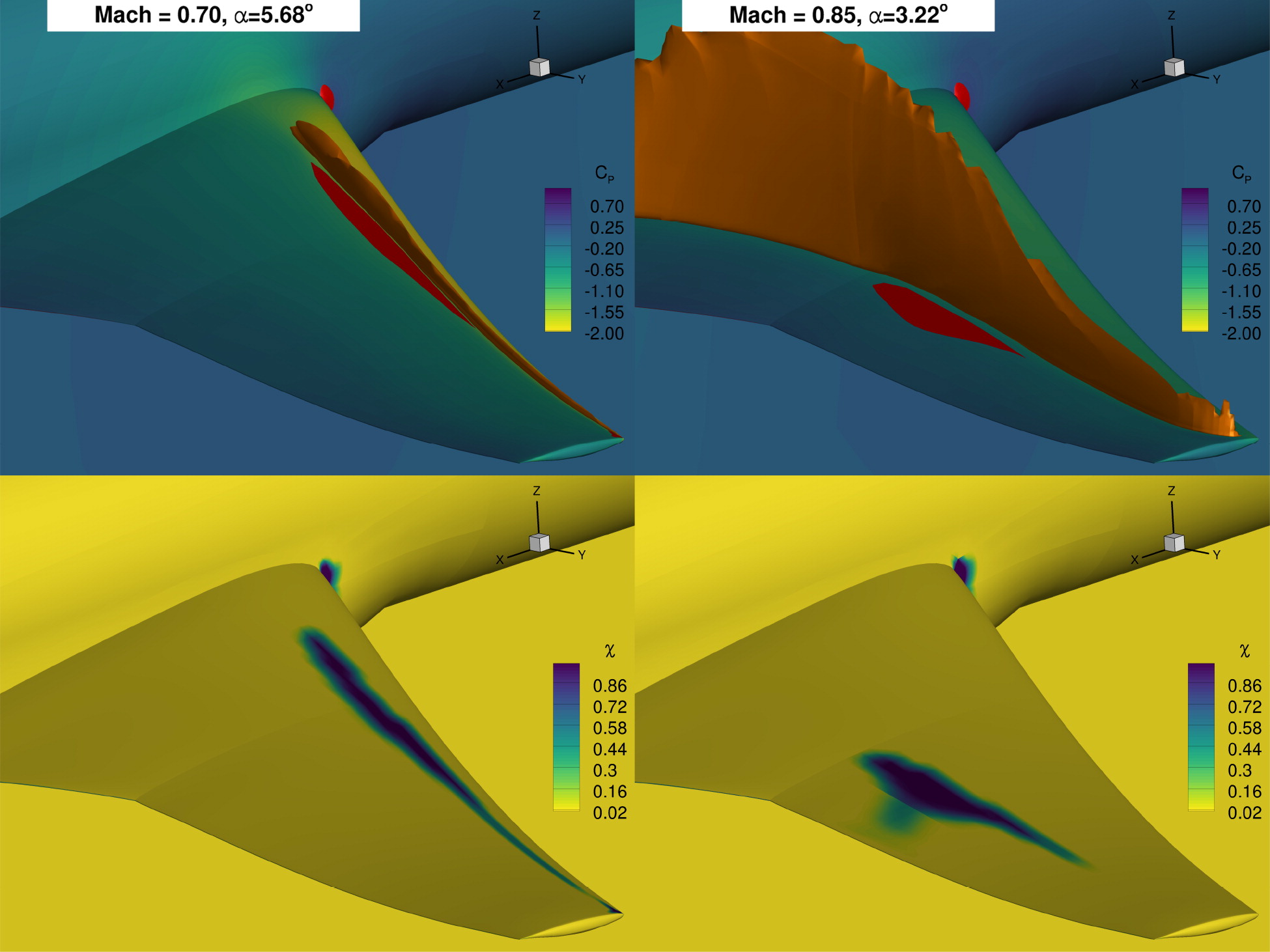}
\caption{Smoothed separation sensor surface distribution for two experimental buffet-onset conditions~\cite{Kenway2017b}.
The separation sensor adequately predicts buffet onset at both high subsonic and transonic flow conditions, implying a suitable generalization of the proposed method.
}
\label{fg_buffet}
\end{figure}

Suppose we want to constraint $C_L^\text{max}$ to match or exceed a target value.
The $C_L^\text{max}$ is achieved at a high angle of attack before the aerodynamic stall when the flow becomes unsteady due to massive flow separation.
To deal with this constraint, \citet{Buckley2010} treated the angle of attack as an independent design variable and ensured the optimized airfoil had sufficient lift by adding a constraint on the difference from the target $C_L^\text{max}$.
The optimized airfoil achieved the desired $C_L^\text{max}$, and the shape looked more practical than those optimized merely under cruise conditions.
However, they evaluated $C_L^\text{max}$ using a steady RANS solver, which is likely not accurate at $C_L^\text{max}$ conditions.

Other researchers have sought to approximate unsteady aerodynamics using surrogate models or data-driven models~\cite{Kou2021}.
Such surrogate models are useful in the flight simulation and control law design of a fixed aircraft shape~\cite{Ghoreyshi2014,Wang2015d}.
Surrogate models of unsteady aerodynamics can also be useful for aerodynamic shape design but are limited to problems with few design variables.
\citet{Wang2018f} used detached eddy simulation (a hybrid method coupling RANS and LES) to train surrogate models in the inverse design of the leading edge location (three geometric parameters) of a three-element airfoil.
\citet{Raul2021} used surrogate models trained by URANS in airfoil shape optimization with six geometric parameters to delay the dynamic stall.
 {
However, it would be too costly to use surrogate models for high-dimensional ASO problems because of the curse of dimensionality.
Thus, off-design aerodynamic constraints call for more advanced modeling methods than those used in predicting cruise aerodynamic coefficients.
}

A promising approach to addressing this issue is to construct alternative constraint formulations, which can be formed by theoretical or experimental studies (such as the criteria on the pressure distribution~\cite{Zhao2016,Li2018e}) or a data-driven approach.
For example, the pressure gradient at the aft of the upper airfoil surface can be used as a constraint to restrict the trailing edge flow separation~\cite{Zhao2016,Li2018e}.
\citet{Li2021f} proposed a modified version of Korn's equation to improve the prediction accuracy of the drag divergence Mach number in supercritical airfoil design.
\citet{Li2019a} presented a data-driven approach to improving the low-speed $C_L^\text{max}$ in transonic wing shape design by adding an implicit constraint on the first and third thickness modes.
Transonic buffet is an undesirable phenomenon caused by shock-induced flow separations, and as shown in Fig.~\ref{fg_buffet}, \citet{Kenway2017b} developed a buffet onset constraint formulation based on a separation sensor function with an upper bound of 4\% determined by experiment-based comparison.
Although the bound of 4\% was defined using a limited number of observations, the formulated constraint is effective and has been used in aircraft aerodynamic and aerostructural design optimization~\cite{Burdette2015a,Mader2017a,Brooks2018a,Bons2018a}.
In addition to being implemented in ADflow~\cite{Mader2020a}, the Kenway--Martins buffet-onset sensor has been implemented in the open-source CFD solver SU2~\cite{Mungua2019a}.
\citet{Garg2017a} developed a similar approach to constrain cavitation in hydrofoil design optimization.

These alternative formulations efficiently model potentially costly constraints using knowledge about the flow mechanisms.
With a good understanding of physics, the data-based model does not necessarily need a large volume of data.
This approach could be used to formulate many other practical design optimization constraints to replace the original constraints at a manageable computational cost, albeit with potentially lower accuracy.

 {
More complex data-driven models are recommended to achieve higher accuracy.
For example, \citet{Li2022c} proposed a physics-based data-driven buffet analysis model generalizable for airfoil and wing shapes.
The analysis model was constructed using a mixture of CNNs and took the pressure and friction distributions as inputs to discover the key physics (shock waves and flow separation) of transonic buffeting.
The model was trained with more than one hundred thousand data points and had a mean absolute error of $0.05\degree$ in buffet factor prediction as measured by 14,886 unseen testing data.
This was shown to be more reliable than the Kenway--Martins sensor.
The training cost of this approach is relatively high, as more than $10^5$ CFD simulations were used.
Future work on physics-based data-driven modeling of other off-design constraints is recommended. 
Research on ensuring high prediction accuracy and sample efficiency will be of great interest.
}

\subsection{Optimization Architecture}
\label{sec:ASOOptArch}

The optimization architecture determines the workflow of ASO.
For example, it determines which optimization algorithm to use and how to utilize high-fidelity aerodynamic analyses.
We have introduced a conventional CFD-based optimization architecture and its limitations in Sec.~\ref{sec:cfd-aso}.
ML enables new optimization architectures that can address some of the limitations in conventional ASO.
We review these developments in this section.

 {
Surrogate-based aerodynamic shape design optimization such as EGO (proposed in 1998), based on iterative infilling refinement guided by the predictive confidence interval, has become increasingly efficient with ML.
The contributions are summarized in three parts:
1) more accurate surrogate modeling such as ME and ANN;
2) more compact geometric design space such as the deep-learning-based modal parameterization;
3) more efficient infill criteria to refine surrogate models.
Applications of surrogate-based optimization have been gradually extended from two-dimensional geometries such as airfoils to three-dimensional wings and aircraft.
Surrogate-based optimization has the potential to tackle multiobjective design and robust design of three-dimensional aerodynamic shapes.
}

 {
Advanced ML techniques in surrogate modeling and compact geometric design space have led to a new ASO architecture---interactive aerodynamic shape design optimization, where accurate and generic data-based aerodynamic models are constructed in advance and then coupled with a suitable optimization algorithm for fast interactive design.
A representative application is Webfoil (deployed in 2018).~\footnote{\url{http://webfoil.engin.umich.edu}}.
The approach uses a large volume of training data ($10^5$) to reduce extrapolatory predictions.
The high computational cost in generating the training data is a significant obstacle to widely applying interactive aerodynamic shape design optimization.
This difficulty calls for subsequent research on more advanced methods in the future.
}

 {
Reinforcement learning has been applied to aerodynamic shape design optimization, promising to improve the efficiency of ASO.
A well-trained RL optimizer is reported to achieve gradient-based optimization efficiency without solving/using aerodynamic derivatives, although the training process of RL could be time-consuming.
Using physical features rather than geometric parameters as the state helps improve the generalizability of RL.
This is similar to the situation in training data-driven models of off-design aerodynamic constraints.
Concerns of RL-based ASO include the high training cost (especially for high-dimensional problems) and difficulties in handling adjustable constraints.
These could be critical drawbacks of RL in practical ASO problems.
}

 {
Conditional generative models enable a quick inverse aerodynamic design architecture called generative inverse design.
These models provide a probabilistic approach to solving ill-condition inverse design problems.
Conditional GAN and conditional VAE are two typical models that have drawn much attention from academia and the industry.
After being trained by labeled data, the model can provide fast inverse designs with a desired aerodynamic performance.
However, due to inaccuracy and exploration capability, solutions of generative inverse design could be distinct from the true optimum.
Thus, they may be more suitable for conceptional design tasks with a low fidelity requirement.
}

\subsubsection{Surrogate-based Optimization}
\label{sec:ASOSBO}

Most aerodynamic design optimization methods using surrogate models rely on an iterative model refinement~\cite{Forrester2009a}.
Such surrogate-based optimization strategies have shown to be useful in various aerodynamic shape optimization applications including single-point design~\cite{Bartoli2017b,Bartoli2019a}, multipoint design~\cite{han2018aerodynamic,Nagawkar2021}, massively multipoint design~\cite{Li2020e}, multi-objective design~\cite{Koziel2016,Leifsson2016,Jim2021a}, inverse design~\cite{Leifsson2011,DU2019Inverse}, and robust design~\cite{Keane2012,Bartoli2018a,Keane2020,Tao2020,Zhang2013pce}.
 {The optimization efficiency of these methods has been improved with the application of ML models.}

Surrogate-based optimization is generally composed of two phases.
The first phase is the design of the experiment (DoE) process, where the initial samples are selected and evaluated to train an initial surrogate model.
The second phase is a refining process where new training samples are iteratively added to refine the surrogate model.
In each iteration, the infill samples are determined by solving sub-optimization problems based on the infill criteria~\cite{Queipo2005a,Forrester2009a}.
 {The performance of surrogate-based optimization is closely related to surrogate modeling efficiency, the design space size, and the infilling effectiveness.}

 {
ML has contributed to surrogate models with higher prediction accuracy and data efficiency (see discussions in Sec.~\ref{sec:ASOAeroCoef}), which helps to improve the efficiency of surrogate-based optimization in aerodynamic design.}
\citet{Bartoli2019a} presented the super-efficient global optimization coupled with a mixture of experts (SEGOMOE), which combined the SuperEGO~\cite{Sasena2002a}, kriging model with PLS~\cite{Bouhlel2016a,Bouhlel2016b,Bouhlel2018c}, and ME~\cite{Bettebghor2011} to handle constrained aerodynamic optimization problems~\cite{Bartoli2016a,Bartoli2019a,Lefebvre2020}.
SEGOMOE has been demonstrated in ASO applications such as wing and nacelle~\cite{Bartoli2016a,Bartoli2018a,Bartoli2019a} and applied to the AGILE project for MDO of the next-generation aircraft~\cite{Lefebvre2017a}.
\citet{Han2016surroopt} developed a general surrogate-based optimization toolbox called SurroOpt with multi-fidelity surrogate modeling~\cite{han2013improving,Han2012b} such as the hierarchical kriging~\cite{han2012hierarchical}, which has been used in multiple aerodynamic and multidisciplinary design optimization problems~\cite{han2018aerodynamic,Zhang2018c,Han2020}.
In addition, deep Gaussian process approximations of the aerodynamic performance have been used to make surrogate-based optimization more efficient~\cite{Renganathan2021,Rajaram2020a}.
The research efforts on surrogate-based aerodynamic shape design optimization are summarized in Table~\ref{tab_sbo}.

\begin{table}[!htbp]
\centering
\caption{Research on surrogate-based aerodynamic shape optimization}
\label{tab_sbo}
\begin{tabularx}{\textwidth}{
    >{\raggedleft\arraybackslash}p{0.1\textwidth}
    >{\raggedright\arraybackslash}p{0.12\textwidth}
    >{\raggedright\arraybackslash}p{0.17\textwidth}
    >{\raggedright\arraybackslash}p{0.23\textwidth}
    >{\raggedleft\arraybackslash}p{0.25\textwidth}}
\hline
\hline
Design variables  & Surrogate model  &  Aerodynamic shape & Description & Reference  \\
\hline
\rowcolor{lightgray}
 2 & Kriging & Building cross-section & $\min C_d$ and $C_l$ variation  & \citet{Bernardini2015}  \\
 4 & Space mapping & Airfoil & $\min C_d$ and $\max C_l$ & \citet{Leifsson2016} \\
\rowcolor{lightgray}
 5 & Kriging & Wing & $\min C_D$ & \citet{Li2019} \\
 6 & Kriging & Airfoil & Delaying dynamic stall & \citet{Raul2021} \\
\rowcolor{lightgray}
 8 & Manifold mapping & Airfoil and wing & $\min ||C_p - C_p^*||_2$ & \citet{DU2019Inverse} \\
 8 & Manifold mapping & Airfoil & $\min C_d$ & \citet{Nagawkar2021} \\
\rowcolor{lightgray}
 8 & ANN & Airfoil & $\min C_d$ & \citet{Renganathan2021}  \\
 10 & Kriging & Airfoil & $\max C_l/C_d$ & \citet{jeong2005efficient}  \\
\rowcolor{lightgray}
 12 & Kriging & Airfoil & $\min C_d$ and $\max C_m$ & \citet{Koziel2016} \\
 17 & Kriging & Wing & $\min C_D$ & \citet{Bartoli2017b,Bartoli2019a} \\
\rowcolor{lightgray}
 30 & ANN & Wing & $\min C_D$ & \citet{Zhang2021a}  \\
40 & ANN & Propeller & $\min$ aeroacoustic noise & \citet{Sun2020}  \\
\rowcolor{lightgray}
 42 & Kriging & Wing & $\min C_D$ & \citet{han2018aerodynamic}\\
 48 & Kriging & Wing & $\min C_D$ & \citet{Li2021} \\
\rowcolor{lightgray}
 19/75 & Kriging & Wing & $\min C_D$ & \citet{Li2019f}  \\
 80 & Kriging & Wing &  $\min C_D$ & \citet{Han2018a,Han2020}\\
\rowcolor{lightgray}
36/72/108 & Kriging & Wing & $\min ||C_p - C_p^*||_2$ & \citet{Han2017}\\
\hline
\hline
\end{tabularx}
\end{table}

Although surrogate-based optimization could be applied to high-dimensional problems such as wing shape optimization with more than 100 design variables, optimization convergence in these cases may not be guaranteed.
\citet{Li2019f} compared wing shape optimization using two numbers of design variables (19 and 75).
More design variables provide more geometric deformation freedom, so the larger number of design variables should result in a better design.
However, they found that the optimization with fewer design variables led to a larger drag reduction using the same computational budget (300 CFD analyses), which means the surrogate-based optimization with 75 design variables was far from convergence.
\citet{Han2018a,Han2017} showed similar issues in high-dimensional wing shape design optimization.
Although using multi-fidelity surrogate modeling could reduce the computational cost~\cite{Han2020}, surrogate-based optimization does not scale well with the dimensionality, and this makes it unsuitable for high-dimensional design problems.

Another contribution of ML to surrogate-based optimization is dimensionality reduction of shape design variables via modal shape parameterization methods.
The benefit of adopting modal parameterization has been demonstrated in the shape design optimization of the CRM wing~\cite{Li2021}.
In that work, surrogate-based optimization was first used in the CRM wing design with 192 FFD design variables.
This approach was shown to be ineffective because there was no improvement to the baseline wing after 1000 CFD simulations.
However, by extracting 40 global wing modes using PCA from deep-learning-based optimal samples, the surrogate-based optimization using wing modes was shown to be almost as efficient and effective as the CFD-based optimization using 192 FFD design variables.
With each adjoint analysis being counted as 0.5 CFD evaluation, the total computational cost of CFD-based optimization was about 500 CFD evaluations in this case, and surrogate-based optimization using 40 modes found similar results (with a difference of 0.3 drag count) after 600-800 CFD evaluations including the training cost in DoE.
This indicates an impressive convergence efficiency of using modal parameterization in surrogate-based optimization.

ML also contributes to making the infill samples more suitable, thus improving the optimization efficiency.
An efficient infill strategy should balance exploitation and exploration.
The expected improvement criterion is one of the most successful infill strategies by providing a suitable trade-off between exploration and exploitation.
With this criterion, EGO gets popular in engineering optimization with black-box solvers~\cite{Queipo2005a,Forrester2009a,Shahriari2016a,Bartoli2019a}.

Since the initial development of EGO, multiple infill criteria have been developed, such as maximizing the probability of improvement and minimizing the lower confidence bounding~\cite{Parr2012,Liu2016b,Shi2019c,Long2021}.
\citet{Liu2016b} showed that using these criteria in parallel could improve the optimization efficiency.
Nevertheless, these criteria always suffer from inaccurate surrogate models and therefore add new samples with abnormal aerodynamic shapes that would not contribute to a better design as these shapes generally lead to inadequate aerodynamic performances.
To address this issue, ML has been used to constrain the searching for infill samples inside reasonable domains, for example, by constructing geometric validity functions~\cite{Li2020a,Li2021d} (described in Sec.~\ref{sec:asoGeofiltering}).
Also, the reasonable domains can be identified by using approximate models trained by aerodynamic data.
For example, \citet{Owoyele2021} proposed a machine-learning-driven active optimizer (ActivO) method using two different machine-learning-based surrogate models called ``weak'' and ``strong'' learners, where the weak learner identifies promising regions in the design space to explore, while the strong learner determines the exact location of the optimum in identified promising regions.
Preliminary proof-of-concept studies in an engine design with nine variables demonstrated superior convergence rates (speedups of 4 to 5$\times$) compared to state-of-the-art optimization methods such as PSO.
\citet{Shi2017} proposed to identify the interesting sampling region by using a feature space fuzzy $c$-means clustering method and SVM, which improved surrogate-based optimization efficiency in an Earth observation satellite MDO problem.
Using a similar concept, \citet{Shi2019d} used SVM to construct a filter-based region of interest for sequentially bias sampling in surrogate-based optimization, and they showed that a better feasible satellite design was obtained with lower computational cost than \citet{Shi2017}.

\subsubsection{Interactive Design Optimization}
\label{sec:ASOInterOpt}

Interactive aerodynamic shape design optimization where ASO can be done near instantaneously is of great interest to aircraft designers.
This demand used to be intractable because of the challenges of training a fast, accurate, and general aerodynamic analysis model~\cite{Laurenceau2008}.
The issue has been addressed to some extent with the advanced modal shape parameterization and data-based modeling techniques, which have been applied to airfoil shape design optimization\cite{Li2018a,Li2019b,Bouhlel2020,Du2020,Du2021a}, wing shape design optimization~\cite{Li2021b}, and conceptual design optimization of wing-fuselage transports and UAVs~\cite{Secco2017,Karali2021}.
The research efforts in this area are listed in Table~\ref{tab_realtimeopt}.

\begin{table}[!htbp]
\centering
\caption{Efforts toward interactive aerodynamic shape optimization with a universal data-based model. (The mean absolute error of the drag coefficients is provided. The two error values (\citet{Li2019b,Du2021a}) are for subsonic and transonic regimes, respectively. )}
\label{tab_realtimeopt}
\begin{tabularx}{\textwidth}{
    >{\raggedright\arraybackslash}p{0.12\textwidth}
    >{\raggedright\arraybackslash}p{0.17\textwidth}
    >{\raggedleft\arraybackslash}p{0.07\textwidth}
    >{\raggedleft\arraybackslash}p{0.09\textwidth}
    >{\raggedleft\arraybackslash}p{0.12\textwidth}
    >{\raggedleft\arraybackslash}p{0.27\textwidth}}
\hline
\hline
 Aerodynamic shape & Model fidelity & Training data & Design variables  & Error (counts) & Reference \\
\hline
\rowcolor{lightgray}
 Transport airplane & Full-potential code & 100,000 & 40 & 5.0 & \citet{Secco2017} \\
 UAV &  Nonlinear lifting line method & 94,500 & 22 & 12.4 & \citet{Karali2021} \\
\rowcolor{lightgray}
 Airfoil & RANS & 113,400 & 16/10  & 0.1 and 0.8 & \citet{Li2019b}  \\
 Airfoil & RANS & 85,201 & 28  & 1.4 and 5.0 & \citet{Du2021a} \\
\rowcolor{lightgray}
 Wing    & RANS & 135,108 & 60 & 0.9 & \citet{Li2021b} \\
\hline
\hline
\end{tabularx}
\end{table}

It is relatively easy to realize interactive optimization in conceptual design because the low-fidelity aerodynamic models are fast and can thus generate large volumes of training data.
For example, \citet{Secco2017} trained ANN models to predict aerodynamic coefficients of transport airplanes with 40 inputs, including wing aspect ratio, wing taper ratio, and maximum airfoil thickness and camber of airfoils.
The ANN model was trained with a database consisting of approximately 100,000 samples evaluated by a full-potential code with computation of viscous effects.
The model has an average error of five drag counts and enabled a 4,000$\times$ speedup compared with the time required by the full-potential code.
\citet{Karali2021} trained an ANN to model the aerodynamics of small UAVs with respect to design variables such as wing span, root and tip chord lengths, and airfoil thickness and camber (22 variables in total).
Optimization using the trained surrogate, which was trained by 94,500 UAVs evaluated by a nonlinear lifting line method, achieved a 5\% increase in aerodynamic efficiency with a cost of a few seconds.

In conceptual design, the geometric variables describe the wing planform (span, taper, sweep, and twist) and overall airfoil parameters such as maximum thickness without detailed airfoil shape parametrization.
These variables have fewer inner relationships than the airfoil shape variables.
Thus, the geometric validity of samples can be easily satisfied, and the accuracy of data-based models can be improved by increasing the number of training samples.
For detailed aerodynamic shape design optimization, a practical sampling method that ensures geometric validity (Sec.~\ref{sec:asoGeofiltering}) is vital to the success of these works since abnormal shapes would bring too many difficulties to the training of accurate surrogate models~\cite{Li2018a}.
For airfoil shape design, one may merely use real-world airfoils such as the UIUC airfoils to bypass the issue~\cite{Zhang2018b,Sekar2019,Thuerey2020}.
For example:
\citet{Zhang2018b} trained a $C_l$ model using UIUC airfoils, and prediction errors on several validation airfoils were mostly smaller than 0.1.
\citet{Sekar2019} trained a CNN model using 1343 UIUC airfoils to recover the airfoil shape using the pressure distribution at a fixed Reynolds number and angle of attack, achieving relative errors below 3\%.
\citet{Thuerey2020} achieved a relative error of 3\% for the pressure and velocity fields using 1505 UIUC airfoils.
Although these results are promising, they may not meet the demand for high-fidelity aerodynamic shape design, where higher accuracy and hence a large number of samples are required.

\begin{figure}[]
\centering
\subfigure[Webfoil, an online data-based airfoil design tool~\cite{Li2019b,Bouhlel2020,Du2021a}]
{\includegraphics[width=0.7\linewidth]{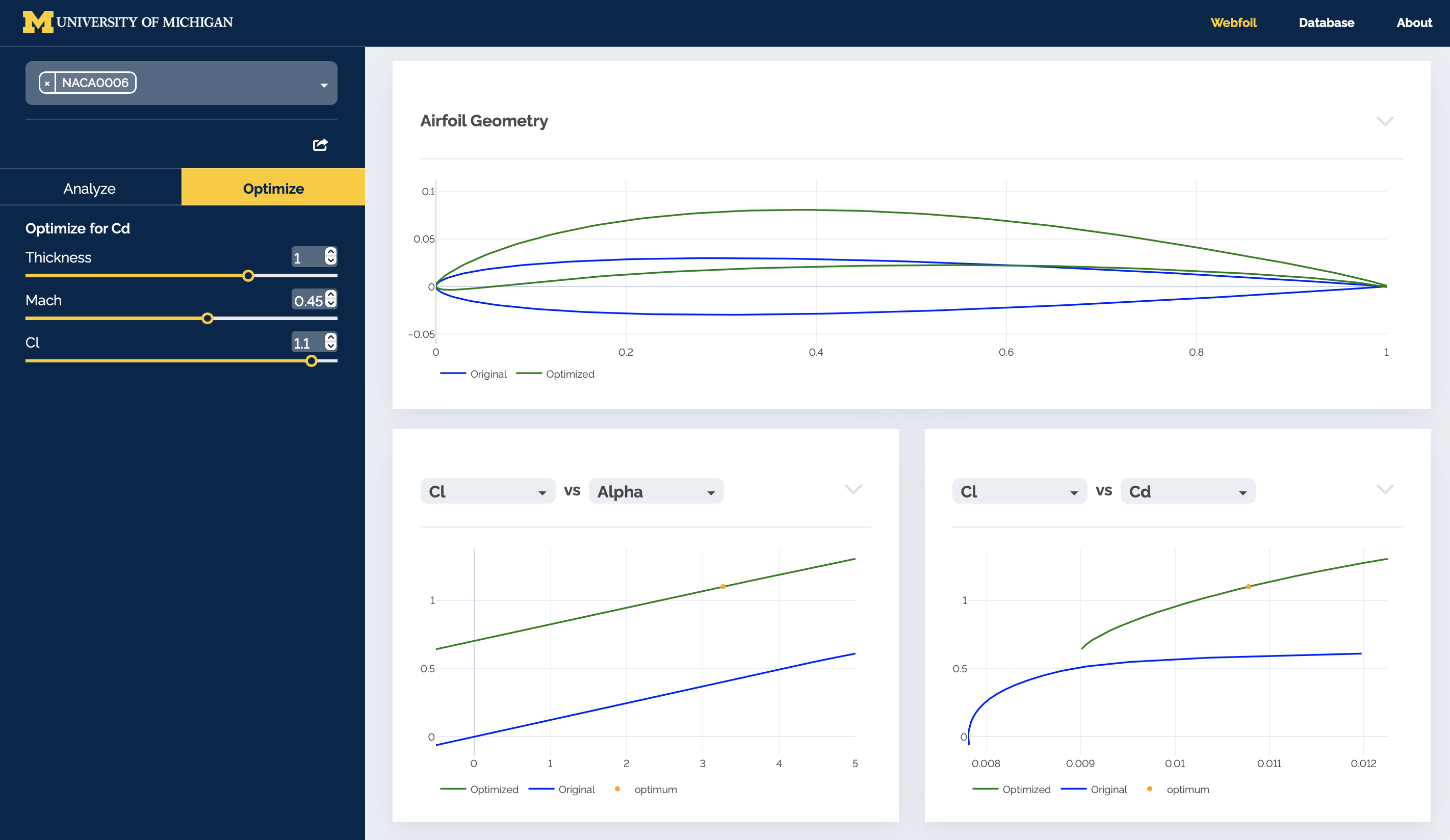}}
\subfigure[Wing designed by multipoint optimization with many flight conditions using the data-based wing aerodynamic analysis model~\cite{Li2021b}.
The colors show the reduced drag forces at different cruise altitudes and Mach numbers via the optimization compared with the CRM wing.]
{\includegraphics[width=0.7\linewidth]{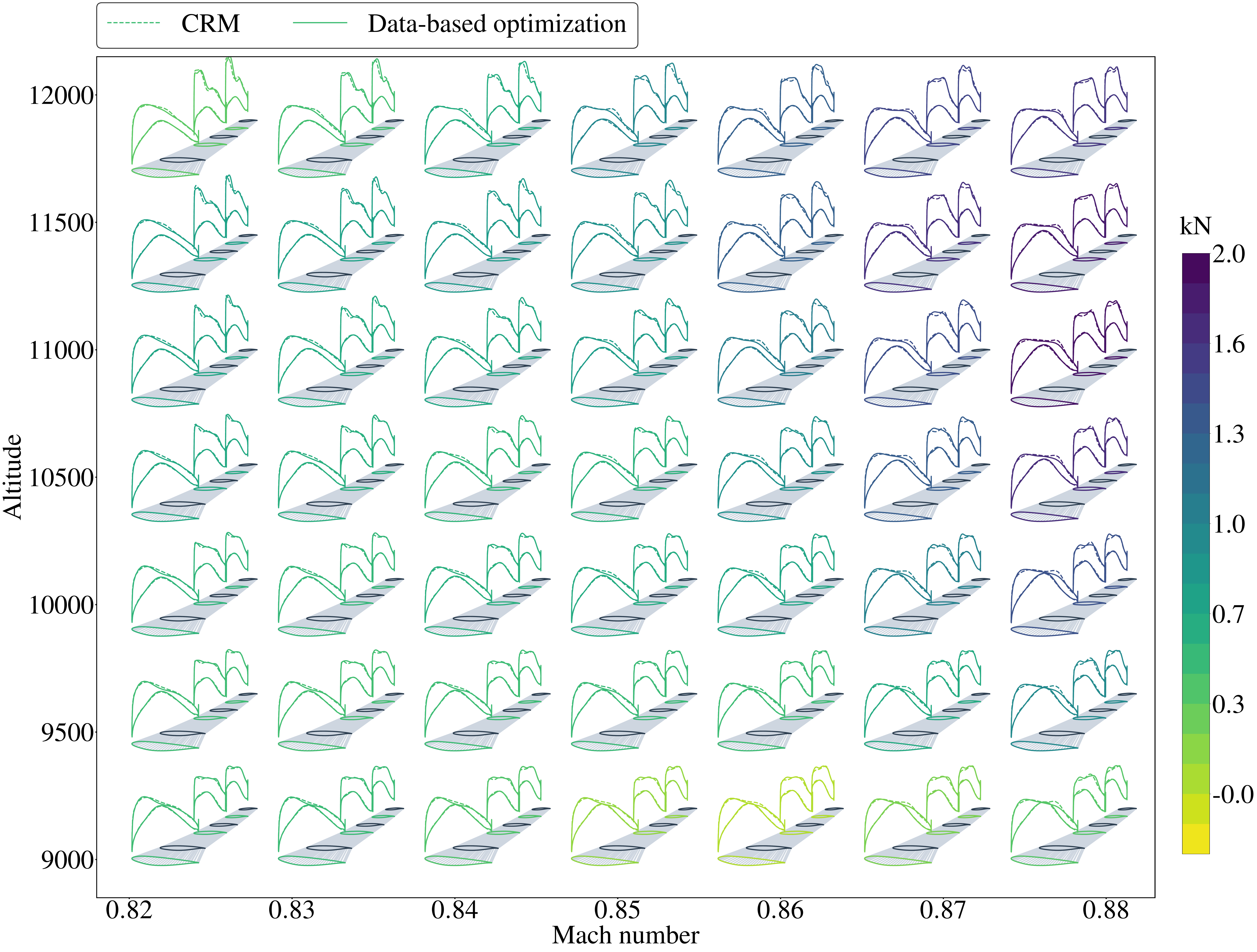}}
\caption{Interactive aerodynamic shape design optimization.}
\label{fg_dataopt}
\end{figure}

Webfoil~\footnote{\url{http://webfoil.engin.umich.edu}} is a web-based tool for fast, interactive airfoil analysis and design optimization via surrogate models on any computer or mobile device with a web browser.
With geometric filtering (introduced in Sec.~\ref{sec:asoGeofiltering}), \citet{Li2019b} built a database with aerodynamic analysis of more than 100,000 airfoils by solving the RANS equations.
The mean relative errors of the data-based airfoil analysis models (for $C_l$, $C_d$, and $C_m$) are 0.145\%, 0.255\%, and 0.160\% in the subsonic regime, respectively, and 0.404\%, 0.826\%, and 0.966\% in the transonic regime, respectively.
When comparing the data-based optimization with RANS-based optimization, the largest differences in minimum $C_d$ are 0.04 counts for subsonic cases and 2.5 counts for transonic cases.
\citet{Du2020, Du2021a} continued this research, and unlike \citet{Li2019b}, a B-spline-based GAN model (with 26 independent shape design variables) was used to generate realistic airfoil shapes without sacrificing thickness accuracy compared with the PCA modal parameterization.
Broader ranges in Reynolds and Mach numbers were considered via a dependence sampling strategy.
In addition to modeling scalar aerodynamic coefficients using feed-forward networks, vector aerodynamic quantities (pressure distribution) were predicted using RNN.
 {Research related to Webfoil opens the door for high-fidelity interactive airfoil analysis and design optimization (Fig.~\ref{fg_dataopt}).}

As mentioned in Sec.~\ref{sec:asoGeoMode}, \citet{Li2021} proposed a deep-learning-based optimal sampling method to generate realistic wing shapes subject to both deep-learning-based validity~\cite{Li2020a,Li2021d} and geometric feasibility constraints~\cite{Kenway2010b}.
Based on this method, \citet{Li2021b} realized fast, high-fidelity shape optimization of the CRM wing by training data-based aerodynamic analysis models of wings.
The models were trained by 135,108 RANS samples encompassing different wing shapes, Mach numbers, and flight altitudes.
The verification on 47,967 unseen wing shapes showed that mean relative errors of $C_L$, $C_D$, and $C_M$ compared to RANS simulations are all within 0.4\%.
The models were further verified in multiple single-point, multipoint, and multiobjective wing design optimization problems.
The optimized wings have similar shapes to those obtained by CFD-based optimization, with differences in $C_D$ of one to two counts.
 {This work showcases the possibility of high-fidelity interactive aerodynamic analysis and design optimization of high-dimensional wing shapes using data-based models.}

 {
Prediction accuracy is essential in high-fidelity interactive design, and this is because the optimum design can be sensitive to the inaccuracy of either objective or constraint functions.
To produce similar optimized shapes to CFD-based optimization, the relative errors of data-based models in evaluating aerodynamic coefficients should be less than 1\%.
It is crucial to control the upper bound of the errors.}
In the efforts mentioned previously, a large aerodynamic database with tens or even hundreds of thousands of samples is essential to train general and accurate aerodynamic models.
The volume of training data makes it unsuitable to use conventional surrogate models like kriging (except when using ME, see Sec.~\ref{sec:ASOAeroCoef}), and NN is a better choice.
\citet{Bouhlel2020} showed that the ANN-based model outperformed the mixture of gradient-enhanced kriging surrogate models in the aerodynamic analysis of airfoils.
ANN can provide accurate derivatives using the built-in AD implementation, which helps perform fast design optimization.
 {Using ANN and RNN,} \citet{Du2021a} enabled airfoil aerodynamic predictions at a wide range of Mach numbers (0.3 to 0.7) and Reynolds numbers ($10^4$ to $10^{10}$).

Optimality and convergence speed are the two key factors that determine the selection of optimization algorithms for interactive aerodynamic design.
Gradient-based optimization algorithms perform a local search and can get trapped in local minima, while some gradient-free algorithms perform a global search and avoid this issue.
However, many investigations have consistently shown that gradient-based algorithms are preferable in practical aerodynamic shape optimization problems, especially for high-dimensional fast design~\cite{Lyu2015b,Yu2018a,Li2019b,Li2021b}.
The reasons are two-fold.

First, the issue of multiple local minima (multi-modality) is not that prevalent.
It has been shown that gradient-based airfoil design optimization starting from different shapes converged to the same results~\cite{He2019c,Li2019a} and gradient-based wing shape design optimization also converged to very similar shapes~\cite{Lyu2015b,Yu2018a,Li2021b}.
Multi-modality issues reported in the literature are usually associated with nonphysical flows such as wing shape design with the inviscid assumption~\cite{Bons2019a}.
Practical aerodynamic shape optimization problems may have several local minima, but the number of minima is limited~\cite{Buckley2010,Chernukhin2013,Bons2019a}.
Still, gradient-based algorithms can find the global minimum by using a multistart strategy~\cite[Tip 4.8]{Martins2021}~\cite{Streuber2018a}.

Second, gradient-free algorithms are inefficient and may not ensure convergence.
The computational cost of gradient-free algorithms increases dramatically with the number of design variables~\cite{Lyu2014f}.
\citet{Li2021b} showed that the performance of two popular evolutionary algorithms (GA and PSO) was much less ineffective than SLSQP (a gradient-based algorithm) in the wing shape design optimization with 57 design variables.
In the single-point wing design optimization, SLSQP converged after one hundred iterations, while PSO required half a million evaluations to find a similar solution.
In the two-objective wing design, SLSQP (with 15 two-point optimizations) solved the Pareto frontier using about 2000 objective and derivatives analyses, while NSGA-II cannot get comparable results even after one million evaluations of the objective functions.
Although data-based aerodynamic analyses are cheap, the low efficiency of gradient-free optimization algorithms makes them unsuitable for interactive aerodynamic shape design, especially for design problems subject to geometric constraints where the surface mesh is needed to be deformed in each evaluation.

 {
For research in interactive design reviewed above, there is a lack of methods for modeling dynamics and multidisciplinary effects, which are essential to evaluate the maneuver performance and control stability.
Despite the challenges of enabling this capability, future work in this direction is recommended.
PINNs and other physics-based data-driven models are worth investigating to improve data efficiency and prediction accuracy for high-fidelity interactive design.}
These studies would contribute to realizing digital twins in aircraft design~\cite{Kapteyn2021}.

\subsubsection{Reinforcement-learning-based Optimization}
\label{sec:ASORLOpt}

Applications of reinforcement learning in aerodynamic shape design have focused mostly on the training of an efficient optimizer.
One of the popular choices is using the current design variables and desired design variables (or increments) as the states and actions, respectively.
\citet{Yan2019} used reinforcement learning to train an optimizer for aerodynamic shape design of missile control surfaces with eight design variables.
\citet{Viquerat2021} applied reinforcement learning to learn how to maximize the lift-to-drag ratio at $Re=200$ by modifying the cylinder shape using $3 \sim 12$ design variables.
\citet{Li2021c} adopted RL to train an optimizer for the section shape design of a tall building with two design variables.
\citet{Qin2020} applied RL in the cascade blade profile design, where ten Hicks--Henne bump functions are used as the design variables to reduce the total pressure loss and increase the laminar flow region.
In these applications, the training is likely a reinforcement process for the agent to memorize the best path to the optimal design.
Although the trained RL policy in this way can provide an optimized shape without the need to call for additional CFD analyses, the policy may not apply to other conditions such as $M$ and $Re$.

\begin{figure}[h]
\centering
\includegraphics[width=\linewidth]{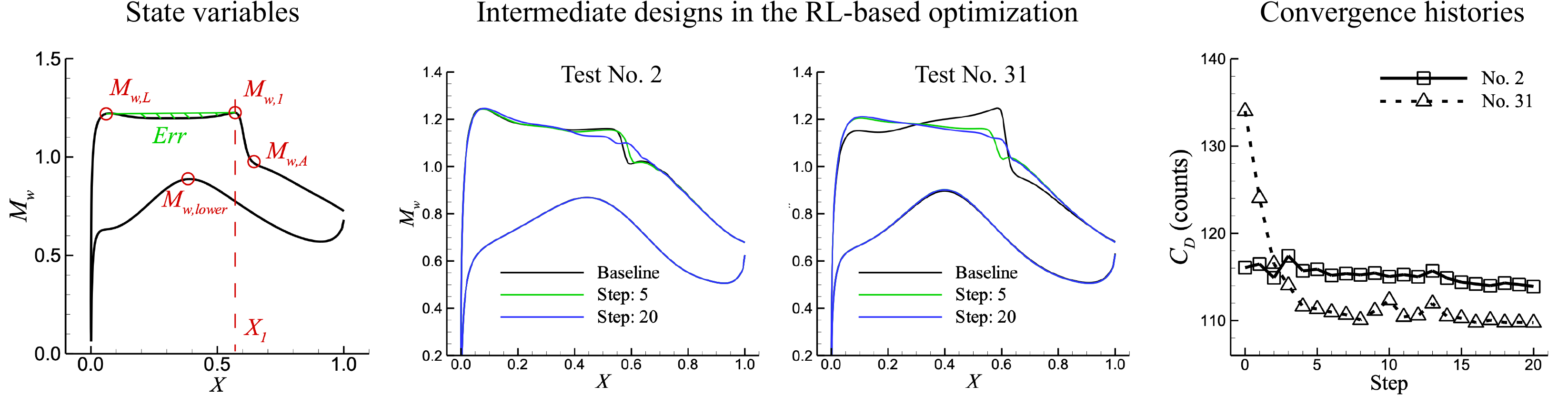}
\caption{RL-based optimizer trained with state variables on the wall Mach number distribution shows a good generalization in supercritical airfoil shape design~\cite{Li2020d}.
Although no aerodynamic derivatives are required, the convergence rate of the RL-based optimizer is comparable to gradient-based optimization.
}
\label{fg_RLOpt}
\end{figure}

To generalize the optimizer trained by RL, \citet{Li2020d} used physical features such as the shock wave location and the suction peak Mach number as the state variables to learn the drag minimization policy for supercritical airfoil design.
(Wall Mach number is defined as the equivalent Mach number calculated based on an isentropic relation using the local pressure coefficient and the free-stream Mach number.)
The shape optimization was implemented by adding a Hicks--Henne bump function controlled by three design variables.
As shown in Fig.~\ref{fg_RLOpt}, most of the drag reduction is achieved in the first five steps using the RL policy as the optimizer, which is as fast as gradient-based optimization.
Surprisingly, the RL policy trained at one flow condition ($M=0.76$ and $Re=5. \times 10^6$) can be used for drag reduction at other flow conditions ($M \in [0.72,0.76]$ and $Re \in [5. \times 10^6, 1. \times 10^7]$).
It would be interesting to investigate whether this generalization could be maintained in airfoil design with more design variables.

One drawback of RL-based optimization is that it does not scale well with the number of design variables, and the overall computational cost may turn out to be similar to performing evolutionary optimization~\cite{Thiele2020a}.
Computational cost reductions have been achieved by using low-fidelity models~\cite{Yan2019,Li2021c} and surrogate models~\cite{Li2020d,Qin2020}.
These approaches trained an initial policy using low-cost models and then used it as a starting point in training the RL-based optimizer with high-fidelity models.
Without using low-fidelity models or surrogate models, training RL-based optimizer is shown to have no advantages over using a conventional gradient-based or gradient-free optimizer\cite{Li2021c,Yan2019}.
Thus, it is more suitable to use RL-based optimization in many-query design than in a single-query design task.
This can be an issue in cases without a suitable low-fidelity model (enough accuracy with negligible computational costs) or high-dimensional problems where it is difficult to ensure the accuracy of surrogate models.

Another drawback of RL-based optimizers is the inability to handle adjustable constraints.
Training an RL policy requires a clear definition of the objective and constraint functions to evaluate the reward.
Most applications either explicitly treated the constraint as a penalty in the reward function~\cite{Yan2019,Viquerat2021} or implicitly imposed it into the aerodynamic analysis (for example, the lift constraint~\cite{Li2020d}).
Thus, the corresponding RL policy may not be directly applied to ASO design tasks with different constraints.
 {
\citet{Kim2022a} used the weighted Chebyshev method to enable RL to handle multiple objective functions and verified the method in airfoil shape design optimization with three geometric parameters. 
Despite the feasibility of using RL in multi-objective airfoil design, the training cost of RL is relatively high ($10^5 \sim 10^6$ episodes in this simple case).
}
This issue may make it impractical for real aerodynamic shape design problems, where high-dimensional geometric parameters and multiple constraints are generally required.

\subsubsection{Generative Inverse Design}
\label{sec:ASOGenOpt}

For a forward problem $\boldsymbol{y} = F( \boldsymbol{x})$, where $\boldsymbol{x}$ and $\boldsymbol{y}$ are the design variables and concerned performance meterics such as drag and lift, inverse design is to solve the corresponding $\boldsymbol{x}$ with the given $\boldsymbol{y}$.
Most inverse design problems in ASO are ill-posed since different shapes ($\boldsymbol{x}$) could have the same performance ($\boldsymbol{y}$), and thus constructing a deterministic function of $\boldsymbol{x}$ with respect to $\boldsymbol{y}$ may fail.
Bayesian inversion~\cite{Arridge2019} provides a probabilistic approach to solving ill-posed inverse problems, where all parameters are treated as realizations of certain random variables.
Then, the key to solving the inverse problem is to evlaute the posterior probability $p(\boldsymbol{x} | \boldsymbol{y})$.

Conditional generative models (introduced in Sec.~\ref{sec:mlAE} and Sec.~\ref{sec:mlGAN}) have drawn vast attention in the inverse design because of their capability of learning the posterior $p(\boldsymbol{x} | \boldsymbol{y})$ from data.
 {
In the industry, generative inverse design is usually shorted as generative design. 
Many artificial intelligence companies, such as Monolith AI and Altair Engineering, take AI-based generative design as a showcase. 
Reported study cases include structural design subject to manufacturing constraints, vehicle outer shape design, and turbomachinery blades, where the number of physical tests is reduced in the design cycle.
}

\begin{table}[!htbp]
\centering
\caption{Airfoil inverse design using deep-learning-based conditional generative models. The performance metrics are continuous conditions except explicitly noted as discrete.}
\label{tab_inversedesign}
\begin{tabularx}{\textwidth}{
    >{\raggedright\arraybackslash}p{0.09\textwidth}
    >{\raggedright\arraybackslash}p{0.16\textwidth}
    >{\raggedright\arraybackslash}p{0.14\textwidth}
    >{\raggedright\arraybackslash}p{0.11\textwidth}
    >{\raggedright\arraybackslash}p{0.14\textwidth}
    >{\raggedleft\arraybackslash}p{0.20\textwidth} }
\hline
\hline
Generative model & Conditions ($\boldsymbol{y}$) & Design variables ($\boldsymbol{x}$) & Noise/latent dimension ($\boldsymbol{z}$) & Training data & Reference  \\
\hline
\rowcolor{lightgray}
CGAN & $C_l/C_d$ and airfoil area (4 categories) & 40 coordinates & 100 & 800 UIUC airfoils & \citet{Achour2020} \\
CGAN & Stall lift or angle  & 40 coordinates & -- & 1550 UIUC airfoils & \citet{Yilmaz2020} \\
\rowcolor{lightgray}
CGAN & Drag polar data       & 40 coordinates & -- & 1550 UIUC airfoils & \citet{Yilmaz2020} \\
CEGAN & $M$, $Re$, and $C_l$ & 192 B\'ezier points and $\alpha$  & 13 & 995 optimized airfoils & \citet{Chen2021a} \\
\rowcolor{lightgray}
PCDGAN & $C_l/C_d$ & 192 coordinates  & -- & 38,802 airfoils & \citet{Nobari2021a} \\
CWGAN-GP & $C_l$ & 248 coordinates  & 3/6 & 3,709 NACA 4-digit airfoils & \citet{Yonekura2021b} \\
\rowcolor{lightgray}
CVAE    & $C_l$ & 248 coordinates  & 2--86 & 3,646 NACA 4-digit airfoils & \citet{Yonekura2021a} \\
CVAE    & Flow outlet angle and aerodynamic loss coefficient & 248 coordinates  & -- & 50,621 airfoils & \citet{Yonekura2021a} \\
\rowcolor{lightgray}
CVAE-GAN & 10 Mach distribution features & 255 wall Mach numbers & 10 & 1500 sample airfoils & \citet{Wang2021b} \\
\hline
\hline
\end{tabularx}
\end{table}

Generally, the trained generative model takes inputs of $\boldsymbol{y}$ (conditions) and $\boldsymbol{z}$ (noises or latent variables) to generate $\boldsymbol{x}$ after learning their distribution in the training dataset.
Then, one can impose certain performance conditions (such as the lift-to-drag ratio at cruise and the maximal lift coefficient) to generate designs of interest.
\citet{Achour2020} used conditional GAN (CGAN) in airfoil inverse design based on conditions of the lift-to-drag ratio and the airfoil area.
\citet{Yilmaz2020} applied CGAN based on a vector of conditions indicating desired aerodynamic performance metrics (such as the stall condition and airfoil drag polar) in the inverse design of airfoil shapes.
\citet{Chen2021a} proposed to use conditional entropic GAN (CEGAN) in airfoil inverse design and showed the benefits over conditional GAN in maximizing the lift-to-drag ratio.
Many efforts of \citet{Chen2021a} were spent in generating training airfoils by performing optimization using the SU2 framework in different $M$, $Re$, and $C_l$.
\citet{Yonekura2021b} showed that conditional Wasserstein GAN with gradient penalty (CWGAN-GP) was advantageous over CGAN in learning the distribution bewteen NACA four-digit airfoils and $C_l$ at $\alpha=5.0\degree$ (evaluated using Xfoil with $Re=3.0 \times 10^6$).
The CWGAN-GP model managed to generate smooth airfoil shapes based on the given $C_l$ conditions, overcoming the nonsmooth airfoil issue in CGAN.
More shape variations were held in airfoils generated by CWGAN-GP as the model avoided mode collapse.

In addition to GANs, conditional VAE (CVAE) has also been investigated in the inverse design of airfoils~\cite{Wang2021b}.
\citet{Yonekura2021a} used CVAE in the inverse design of NACA four-digit airfoils with the condition of $C_l$ at $\alpha=5.0\degree$ (same with that in~\cite{Yonekura2021b}).
They concluded that using a moderate latent dimension (no more than 16) was preferable to compromise influences of the error of matching the condition ($C_l$ at $\alpha=5.0\degree$) and the airfoil shape variation.
\citet{Yonekura2021a} also applied CVAE to the inverse design of turbine blade airfoils, where the flow outlet angle and aerodynamic loss coefficient were two performance conditions of interest and 50,621 sample airfoils were generated based on the Pak-B turbine blade shape to train the model.
With such a large volume of training data, the mean absolute error of the flow outlet angle in airfoils generated by the CVAE model was $0.609\degree$, where the absolute angle was approximately $62.9\degree$.
This error is acceptable in the rough design of turbine blades.
\citet{Yonekura2021c} investigated the performance of CVAE with the normal distribution ($\mathcal{N}$-CVAE) and the von Mises-Fischer distribution ($\mathcal{S}$-CVAE) in inverse airfoil design.
They concluded that $\mathcal{S}$-CVAE was superior to $\mathcal{N}$-CVAE when applied to a single type of airfoils because it could separate the data in the latent space, and for multiple types of airfoils, $\mathcal{N}$-CVAE was more capable of generating novel shapes by combining different features among them.

As shown in Table~\ref{tab_inversedesign}, most conditional models are used to design the airfoil shape.
Once trained, these models can provide fast designs with a reasonable performance.
Nevertheless, these designs usually do not precisely meet the condition requirement due to errors in the conditional generative model.
\citet{Chen2021a} showed that conditional generative models can provide fast reasonable airfoil designs but they were still distinct from the optimal airfoil shapes.
There are also other constraints (both geometric and aerodynamic) that need to be imposed in practical applications.
Thus, it is essential for the models to hold a good exploration capability, that is, to generate diverse designs with the same condition.
The noise or latent term ($\boldsymbol{z}$) in the input of conditional generative models plays such a role to make the model maintain the exploration capability.
\citet{Yilmaz2020} showed that the exploration can also be achieved by the inclusion of dropout layers to the generative models, and other efforts such as the development of performance conditioned diverse GAN (PCDGAN)~\cite{Nobari2021a} have been made to enhance this capability.
However, diversity leads to another intractable problem for design optimization because it is not easy to choose the optimal one from the diverse designs.
In the work of \citet{Achour2020}, it costs 4,000 evaluations to explore the noisy space of CGAN to obtain an airfoil with a lift-to-drag ratio of 140 and an area of 0.10.
It is computationally expensive to analyze these designs using a high-fidelity aerodynamic analysis model.
Thus in detailed aerodynamic shape design, generative inverse design may not show advantages over conventional ASO methods.
\citet{Chen2021a} suggested using the generative inverse design as a start point for conventional CFD-based optimization, and with this strategy, it was shown that half iterations can be reduced in multiple airfoil optimization tests.

Another application of conditional generative models is to design the distribution of pressure coefficients or wall Mach numbers rather than the aerodynamic shape.
These distributions contain more information than drag and lift coefficients and can reveal off-design performance, such as buffeting and transition.
Thus,  {they are useful} in the design of supercritical airfoils~\cite{Zhao2016} and natural laminar flow airfoils~\cite{Zhang2015c}.
In inverse design, it is not easy for designers to provide an optimal targeted distribution.
Nevertheless, it is easy to describe the features of interest (such as the suction peak and the shock wave location).
\citet{Wang2021b} used an integrated generative network of conditional VAE and GAN (CVAE-GAN) to generate target wall Mach distributions using the conditions of several essential features, including locations of the suction peak, shock, and aft loading.
\citet{Lei2021a} provided another alternative that used WGAN to learn physical pressure distributions of transonic airfoils and selected a desirable distribution using GA based on features of interest such as the suction peak and pressure gradient of the suction platform.
Then, multiple approaches can be used to find the aerodynamic shape corresponding to the distribution, such as constructing surrogate models~\cite{Wang2021b,Lei2021a} and using adjoint-based optimization~\cite{Zhang2020a}.

\section{Conclusions and Outlook}
\label{sec:conclusion}

 {
In this paper, we reviewed recent applications of ML in aerodynamic shape design optimization.
The technical concepts of commonly-used ML models were introduced for convenience.
Recently developed deep learning models have strong learning abilities.
This makes it possible to learn the underlying features from high-dimensional sparse data sets and to train accurate prediction models with large volumes of data.
The application of these models has shown promise in addressing challenging aerodynamic shape design optimization problems that require high dimensionality, high computational costs, or both.}

 {
We review ML applications in ASO, addressing three aspects: geometric design space, aerodynamic evaluations, and optimization architectures.
The conclusions regarding ML applications are as follows:}
\begin{itemize}
\item  {ML helps parameterize the geometric design space using fewer design variables or more restricted bounds by learning from the empirical knowledge of human designers, historical designs, or low-fidelity optimization results. 
Well-trained ML-based parameterization can exclude abnormal aerodynamic shapes with a low probability of excluding innovative shapes.
Such parametrizations reduce the number of design space dimensions, significantly reducing the computational cost of optimization. 
} 
\item  {Advances in ML have made it possible to handle larger volumes of training data and thus achieve higher prediction accuracy in the aerodynamic coefficients.} 
 {Nonlinear dimensionality reduction methods, such as manifold learning, VAE, and GAN, help predict the high-dimensional flow solutions, which is useful to accelerate high-fidelity simulations, inverse design, and design optimization.}
 {Most ML prediction models are interpolative, so they may lose accuracy in extrapolatory predictions.
PINNs obey the underlying physical lows and thus potentially have the extrapolation ability.
Applications of PINNs in ASO warrant further research.
Coupling physical mechanisms and ML has shown effectiveness in modeling off-design aerodynamic constraints such as buffeting onset.}
 {Physics-based data-driven modeling provides an accurate alternative to imposing off-design constraints than empirical formulas, and more studies of them would contribute to obtaining practical designs with ASO.}
\item  {ML contributes new optimization architectures to address the limitations of CFD-based optimization.
Surrogate-based optimization is becoming more efficient with advanced ML models, which help solve ASO difficulties with discontinuous aerodynamic functions, multi-objective optimization, and design optimization considering uncertainties. 
Interactive design has been achieved in the ASO of airfoils, wings, and aircraft with ANN-based aerodynamic coefficient prediction models pre-trained by a large number of simulation data.}
 {Reinforcement learning can perform ASO without using derivatives, but this has only been demonstrated for problems with a few design variables.}
 {More research is required to overcome the difficulty of RL-based optimization in training efficiency and handling adjustable constraints. 
Conditional generative models (such as conditional GAN and VAE) provide a fast inverse design approach and show promise in conceptional design.}
\end{itemize}

 {The applications show that ML can address some challenging issues in CFD-based optimization (discussed in Sec.~\ref{sec:cfd-asoChallenges}).
Nevertheless, we find that many ML models are applied to less challenging ASO problems without comparing to conventional ASO.}
We think that the core strength of ML is that it can solve problems that are intractable using conventional methods rather than providing an alternative to solving the more manageable problems.
Thus, it is of greater interest to study ML models with practical ASO problems focusing on solving intractable problems when using conventional CFD-based optimization.
Future ML studies in ASO should address an unsolved issue and end with a comparison with state-of-the-art methods.

To address the industrial demands in ASO, we recommend further research on the following topics:
\begin{itemize}
\item  {Extrapolatory prediction models.
Most prediction models used in ASO inherently interpolate the data and are thus generally incapable of extrapolation. 
This limitation leads to the need for a large volume of training data to avoid extrapolations. 
PINNs are a potential solution to extrapolation beyond the efforts that shrink the design space.
Further studies applying PINNs in benchmark ASO cases are expected to validate its potential.}
\item  {Efficient sampling methods. 
It is expensive to obtain high-fidelity simulation data, and the interpolation feature increases the requirement for large volumes of training data. 
For example, the data levels in the interactive airfoil and wing shape design reach $10^5$, which may bring in a too high computational cost for industrial cases. 
Thus, in addition to improving the extrapolatory ability, advanced sampling methods are expected to improve the data efficiency.}
\item  {Handling multi-sources of data. 
Most studies in ASO with ML lack a coupling of different data sources, such as simulation and experimental data. 
Aerodynamic design with no support from experiments may lack practicality from an industrial view, but experimental data depends on specialized facilities and are much inaccessible for most academic researchers. 
Thus, deeper collaborations between industry and academia are recommended to accomplish such research.}
\end{itemize}

 {
To achieve faster progress, it is essential to keep students and engineers in the ASO field familiar with the development of ML models.
Open accessible workshops and short online courses have played an essential role in the past.
In the future, the introduction of ML models and related applications in engineering problems are recommended for undergraduate and graduate curricula.
Textbooks covering both fundamental optimization and ML basics with engineering examples are helpful.}~\cite[Sec.~10.5]{Martins2021}~\cite{Brunton2019}

 {
ML should not be viewed as a replacement for CFD-based optimization with gradient-based algorithms and gradients computed using an adjoint method, which effectively solves large-scale high-dimensional ASO problems.
If the ASO problem is deterministic and does exhibit much multimodality, conventional CFD-based optimization is a suitable choice.
Thus, while following up on the progress of ML techniques, it is also essential to get familiar with conventional methods and their limitations.
We recommend that young researchers build a solid background in specific research fields before turning to ML solutions.
}

 {
We recommend open-source models and open-access data in future research on ML for aerodynamic shape design optimization.}
The rapid development of ML is partly due to the community's open-source tradition.
With more data, models, and software accessible for ASO, exciting new developments can be expected in the near future.

\section{Bibliography}
\bibliographystyle{elsarticle-num-names}
\bibliography{mlaso}
\end{document}